\documentclass{article}

\PassOptionsToPackage{numbers, compress}{natbib}

\usepackage[preprint]{neurips_2021}

\usepackage[utf8]{inputenc} 
\usepackage[T1]{fontenc}    
\usepackage{hyperref}       
\usepackage{url}            
\usepackage{booktabs}       
\usepackage{amsfonts}       
\usepackage{nicefrac}       
\usepackage{microtype}      
\usepackage{xcolor}         

\usepackage[toc,page,header]{appendix}

\usepackage{url}
\usepackage{microtype}
\usepackage{graphicx}
\usepackage{subfigure}
\usepackage{booktabs} 
\usepackage{xcolor}
\usepackage{amsthm}
\usepackage{multirow}
\usepackage{diagbox}
\usepackage{arydshln}
\usepackage{mathtools}
\usepackage{soul}
\soulregister\cite7
\soulregister\gls7
\soulregister\etal7

\usepackage{booktabs}
\usepackage{float}
\usepackage{hhline}
\usepackage{tabularx}
\newcolumntype{Y}{>{\centering\arraybackslash}X}
\newcolumntype{s}{>{\hsize=.8\hsize}Y}
\newcolumntype{t}{>{\hsize=.6\hsize}Y}

\newcolumntype{?}{!{\vrule width 1pt}}
\usepackage{multirow}
\usepackage{makecell}
\usepackage{stfloats}
\usepackage{wrapfig}
\usepackage{dsfont}

\usepackage{commath} %
\usepackage{authblk}

\title{JOKR: Joint Keypoint Representation for Unsupervised Cross-Domain Motion Retargeting}

\author{Ron Mokady}
\author{Rotem Tzaban}
\author{Sagie Benaim}
\author{Amit H. Bermano}
\author{Daniel Cohen-Or}
\affil{The Blavatnik School of Computer Science, Tel Aviv University}

\begin{document}

\maketitle

\begin{abstract}
The task of unsupervised motion retargeting in videos has seen substantial advancements through the use of deep neural networks.  While early works concentrated on specific object priors such as a human face or body, recent work considered the unsupervised case. When the source and target videos, however, are of different shapes, current methods fail. To alleviate this problem, we introduce JOKR - a JOint Keypoint Representation that captures the motion common to both the source and target videos, without requiring any object prior or data collection. By employing a domain confusion term, we enforce the unsupervised keypoint representations of both videos to be indistinguishable. This encourages disentanglement between the parts of the motion that are common to the two domains, and their distinctive appearance and motion, enabling the generation of videos that capture the motion of the one while depicting the style of the other. To enable cases where the objects are of different proportions or orientations, we apply a learned affine transformation between the JOKRs. This augments the representation to be affine invariant, and in practice broadens the variety of possible retargeting pairs. This geometry-driven representation enables further intuitive control, such as temporal coherence and manual editing. Through comprehensive experimentation, we demonstrate the applicability of our method to different challenging cross-domain video pairs. We evaluate our method both qualitatively and quantitatively, and demonstrate that our method handles various cross-domain scenarios, such as different animals, different flowers, and humans. We also demonstrate superior temporal coherency and visual quality compared to state-of-the-art alternatives, through statistical metrics and a user study. Source code and videos can be found at: \url{https://rmokady.github.io/JOKR/}.
\end{abstract}

\vspace{-0.1cm}
\section{Introduction}
\label{sec:intro}
\vspace{-0.1cm}

One of the fields that has seen the greatest advancements due to the deep learning revolution is disentangled content creation. Under this paradigm, deep neural networks are leveraged to separate content from style, even when this separation is highly non-trivial. For example, in the image domain, several works have examined disentangling the scene's geometry (or \textit{content}), from its appearance (or \textit{style}). This enables exciting novel applications in image-to-image translation, such as converting day images to night ones, giving a photo-realistic image the appearance of a painting, and more \cite{Gatys_2016_CVPR,unit,munit, DRIT_plus}.  It turns out that preserving the geometry and translating the texture of objects in an image is much simpler for networks, compared to performing translations on the geometry itself (e.g., translating a horse to a giraffe) \cite{transgaga, katzir2020crossdomain}. This task, of transferring shape or pose, is even more difficult when considering videos and motion. To address shape-related translations, many works use a sparse set of 2D locations to describe the geometry. For static images, it has already been shown that these \textit{keypoints} can be learned in an unsupervised manner, enabling content generation that matches in pose (or other geometric features) across domains \cite{transgaga}. For videos, however, success is more limited. To tackle videos, many approaches employ heavy supervision. For example, prior knowledge regarding human body skeletons~\cite{chan2019everybody, shysheya2019textured} or human facial expressions ~\cite{zakharov2019few, zollhofer2018state} is often employed to drive keypoint locations. 
These approaches are limited to the specific domain of faces or human bodies, where such supervision is abundant, and do not allow transferring or retargeting motion between domains. Unsupervised approaches have also been proposed \cite{bansal2018recycle,siarohin2019first}, but they too fall short when it comes to cross-domain inputs (see Section~\ref{sec:results}).

\begin{figure*}
\begin{tabular}{lcc}

~ & ~~~ $t$ ~~~~~~~~~~~ $t+5$ ~~~~~~~~ $t+10$ ~~~~~~ $t+15$  & ~~~ $t$ ~~~~~~~~~~~ $t+5$ ~~~~~~~~ $t+10$ ~~~~~~ $t+15$ \\
\rotatebox[origin=t]{90}{Input}  &
\raisebox{-.3\totalheight}{\includegraphics[width=0.45\textwidth]{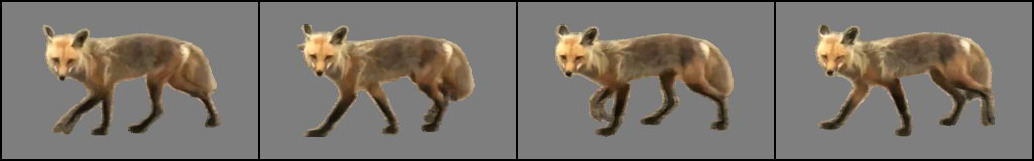}}  &
\raisebox{-.3\totalheight}{\includegraphics[width=0.45\textwidth]{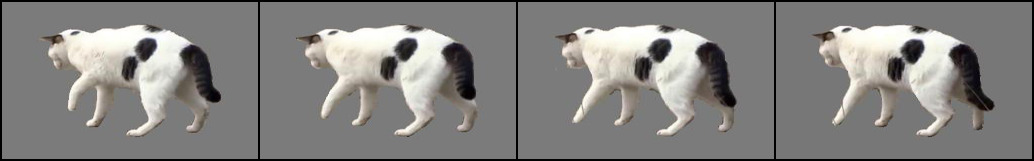}} \\
\rotatebox[origin=t]{90}{Ours}  &
\raisebox{-.3\totalheight}{\includegraphics[width=0.45\textwidth]{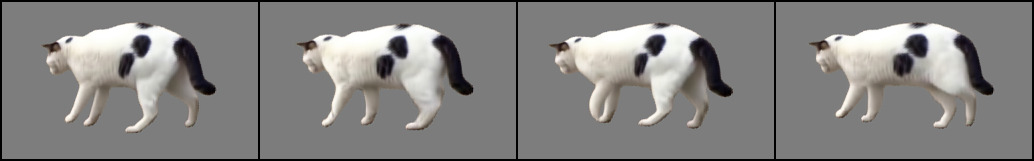}}  &
\raisebox{-.3\totalheight}{\includegraphics[width=0.45\textwidth]{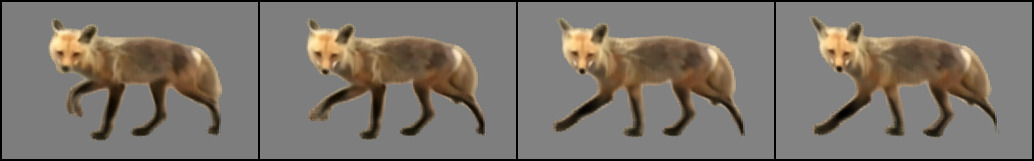}} \\
~ & ~~~~~~~~~ (a) ~~~~~~~~~ &  ~~~~~~~~~ (b)  ~~~~~~~~~ \\
\end{tabular}
\caption{Motion retargeting results from a single video pair. The input videos (top) are used to generate the retargeting (bottom). As can be seen, each generated frame corresponds to the motion portrayed by the source, while keeping the style of the target. For example, note how at frame $t$ (b), both animals simultaneously lift their front leg, however each one does it in its own style, i.e. the generated fox's leg pose is similar to the original video ($t+10$ (a)). For all examples, we denote the starting frame as $t$.
}
\label{fig:teaser}
\end{figure*}

In this paper, we demonstrate how enforcing shared geometric meaning on the latent representation induces the desired disentanglement between motion and appearance in videos. More specifically, we demonstrate how said representation can be employed to retarget motion across domains in a temporally coherent manner, without the need for any prior knowledge. In other words, we propose to use a joint keypoint representation as a bottleneck to portray pose and motion. 
Unlike skeletons, which require prior knowledge to drive video generation, we let the network describe the pose and motion by freely choosing keypoint locations, such that the same description is meaningful in both domains. As can been seen in Figure~\ref{fig:teaser}, this concept enables cross-domain motion retargeting. By employing a domain confusion term, we enforce the keypoint representations of both domains to be indistinguishable, thus achieving our introduced JOint Keypoint Representation (\textit{JOKR}). 

We evaluate the expressiveness of our representation through the setting of cross-domain motion retargeting using a single pair. Using a single video pair alleviates the need for extensive data collection and labelling, which is a costly and exhaustive task, especially for videos. Given a source and target videos, we generate a new video which exhibits the motion of the source video with the style of the target. The design of our joint keypoint representation encourages disentanglement between the motion (content) and the native style of the videos. For example, Figure~\ref{fig:teaser} portrays motion retargeting results between a cat and fox. As can be seen in $t+10$,(a) both the fox and cat lift their front leg in sync, however the amount that is lifted and the pose of the paws, remain distinctive to each animal to produce a realistic result. Through a novel affine-invariant domain-confusion loss, we prevent the keypoints from capturing information about the image-level texture, shape, or distinctive poses, thus enabling the disentanglement between the parts of the motion that are common to the two domains and their distinctive styles. Furthermore, this geometry-driven representation is also meaningful enough to enable intuitive control, such as imposing temporal coherence or even simple manual editing (see Fig.~\ref{fig:editing}). Note that we use auxiliary input in the form of the object's silhouette to avoid background related artifacts. Our method operates on both manually annotated silhouettes and silhouettes obtained using an off-the-shelf pretrained saliency segmentation network, thus our method is not limited to specific objects. From the JOKR bottleneck, we train two domain specific decoders that are used to generate realistic videos  --- one for the source video and another for the target. This results in realistic videos which portray one object performing movements of the same meaning and timing as another while keeping the original style.

We evaluate our method both qualitatively and quantitatively on a variety of video pairs from the YouTube-VOS dataset~\cite{xu2018youtube} depicting different cross-domain objects. For example, we use animals that exhibit different shapes and styles. Numerically, we demonstrate that our method is superior both in temporal coherency and visual quality. We also demonstrate the capability of our method on other object types such as flowers and dancing persons as well as in the setting of GIF synchronization. We then illustrate how our representation can be leveraged for simple and intuitive manual editing, demonstrating that our method generates semantic keypoints. Lastly, a comprehensive ablation study is carried, demonstrating the necessity of each component of our method.

\vspace{-0.1cm}
\section{Related Work}
\vspace{-0.1cm}

\noindent\textbf{Motion retargeting.}\quad
Many approaches exist for transferring motion from one video to another. 
Several works operate in the supervised video-to-video setting ~\cite{wang2018video, pan2019video, mallya2020world}, which requires supervision in the form of source and corresponding target frames. Other works consider motion specific to the human face or body~\cite{averbuch2017bringing, 9009591, nagano2018pagan, kim2018deep, nirkin2019fsgan, villegas2018neural, aberman2019learning}, still requiring a strong prior in the form of extracted landmarks or a 3D model. For instance, Zakharov et al.~\cite{9009591} use facial landmarks to transfer face movement. DeepFake methods~\cite{dfgithub, masood2021deepfakes}  were the first to swap faces between two input videos using deep neural networks. 
These methods either implicitly use 3D facial representations, or use facial landmarks to crop and align the face, and are again limited to facial data. Other works ~\cite{chan2019everybody, aberman2019deep, gafni2020single} transfer motion from one human body to the other using extracted human silhouettes or 2D skeletons. Unlike these approaches, our method does not assume any specific prior. 

Siarohin et al.~\cite{siarohin2019animating, siarohin2019first} and Wiles et al.~\cite{wiles2018x2face} assume no prior and consider the task of image animation given a source video and a target image. As the target video is not provided, the motion is borrowed completely from the source video, meaning that in the case of transferring between different objects, the resulting motion is unrealistic. Motion synchronization ~\cite{bazin2016actionsnapping} can also be used, but it cannot generate novel frames, therefore is limited to only reordering the motion. Most related to ours is the work of RecycleGAN~\cite{bansal2018recycle}, which considers unsupervised motion retargeting using a single video pair. The approach is based on cycle consistency losses in space and time, and adversarial losses. However, their approach struggles to align objects with a substantially different shape. Using our joint keypoint representation that encodes motion common to both videos, our method correctly handles a variety of cross-domain scenarios, where objects are of different shapes.

\noindent\textbf{Shared Geometric Representation.}\quad
A large body of works considers the unsupervised learning of keypoint representations ~\cite{ thewlis2017unsupervised, zhang2018unsupervised, transgaga, suwajanakorn2018discovery, jakab2018unsupervised, siarohin2019animating, siarohin2019first}. Some, by directly training an autoencoder to reconstruct the given images ~\cite{zhang2018unsupervised}, while others by solving a downstream task, such as conditional image generation ~\cite{jakab2018unsupervised}.  
Wu et al.~\cite{transgaga} translate between objects of different shapes by using keypoints as a shared geometric representation. However, unlike our method, their method does not contain any temporal constraint, intuitive editing, nor does it demonstrate any result over videos.
Jakab et al.~\cite{jakab2020self} use unpaired pose prior to train a keypoint extractor, but their work is also limited to humans.  
Similar to ours, Siarohin et al.~\cite{siarohin2019first} learn keypoint representations in an unsupervised fashion, however, as we demonstrate (see Section~\ref{sec:results}), their method cannot handle cross-domain videos well.
The use of a shared representation is also prevalent in other image generation tasks. UNIT~\cite{unit} and MUNIT~\cite{munit}, for example, use a shared representation for image-to-image translation.
Other works ~\cite{ori, domainintersection, mokady2019mask} use shared representations to disentangle the common content of two domains from the separate part. Unlike these methods, our work disentangles motion from style over videos.

\begin{figure*}
\begin{tabular}{l}

\includegraphics[width=1\textwidth]{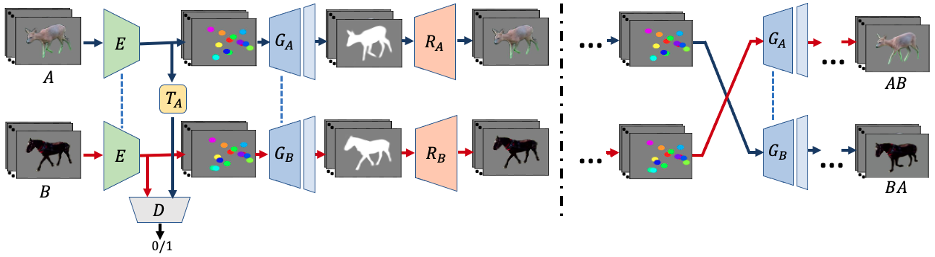} \\
~~~~~~~~~~~~~~~~~~~~~~~~~~~~~ $(A)$ Training  ~~~~~~~~~~~~~~~~~~~~~~~~~~~~~~~~~~~~~~~~~~~~~~~~~~~~~~~~~~~~~~ $(B)$ Inference \\

\end{tabular}
\caption{Method illustration. $(A)$ Training: Given Videos $A$ and $B$, we extract our Joint Keypoint Representation using network $E$. Discriminator $D$ is used to encourage the representation to be of the same distribution, up to the learned affine transformation $T_A$. $G_A$ and $G_B$, which share weights for all but the last layer, translate the given keypoint to a segmentation map that is then refined by $R_A$ and $R_B$. $(B)$ Inference: We pass the keypoints of $A$ (resp. $B$) to $G_B$ (resp. $G_A$). The result $BA$ depicts the motion of $A$, as represented by the keypoints, and the appearance and style of $B$, due to the usage of $G_B$ and $R_B$. The result $AB$ is its counterpart. }
\label{fig:method}
\end{figure*}

\vspace{-0.1cm}
\section{Method}
\vspace{-0.1cm}

\begin{figure*}
\begin{tabular}{lll}
& ~~~~~~~~ $t$ ~~~~~~~~~~ $t+1$ ~~~~~~~~ $t+2$  ~~~~~~~~ $t+3$
& ~~~~~~~~ $t$ ~~~~~~~~~~ $t+1$ ~~~~~~~~ $t+2$  ~~~~~~~~ $t+3$ \\

$(1)$ &
\raisebox{-.4\totalheight}{\includegraphics[width=0.45\textwidth]{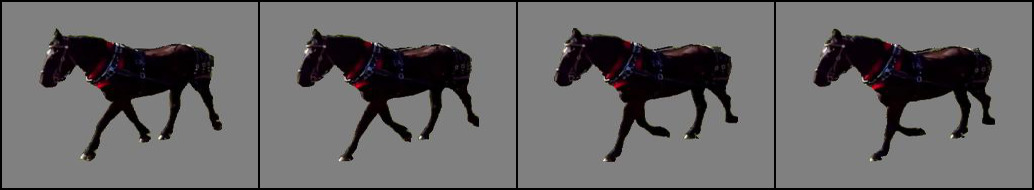}} &
\raisebox{-.4\totalheight}{\includegraphics[width=0.45\textwidth]{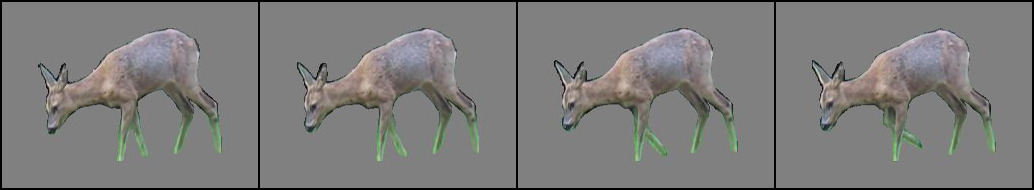}} \\
\noalign{\vskip 1mm} 
$(2)$ &
\raisebox{-.4\totalheight}{\includegraphics[width=0.45\textwidth]{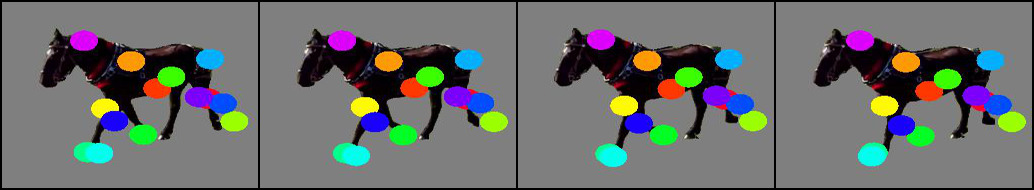}} &
\raisebox{-.4\totalheight}{\includegraphics[width=0.45\textwidth]{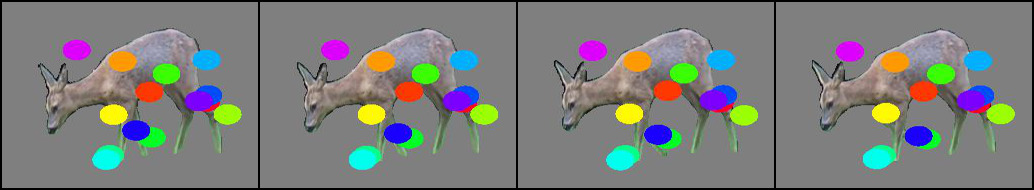}} \\
\noalign{\vskip 1mm} 
$(3)$ &
\raisebox{-.4\totalheight}{\includegraphics[width=0.45\textwidth]{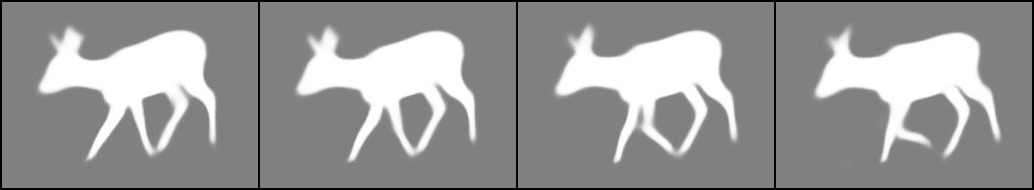}} &
\raisebox{-.4\totalheight}{\includegraphics[width=0.45\textwidth]{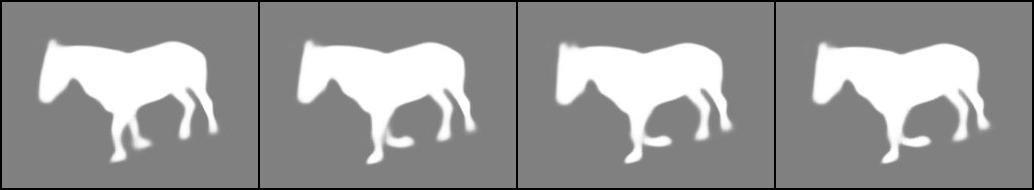}} \\
\noalign{\vskip 1mm} 
$(4)$ &
\raisebox{-.4\totalheight}{\includegraphics[width=0.45\textwidth]{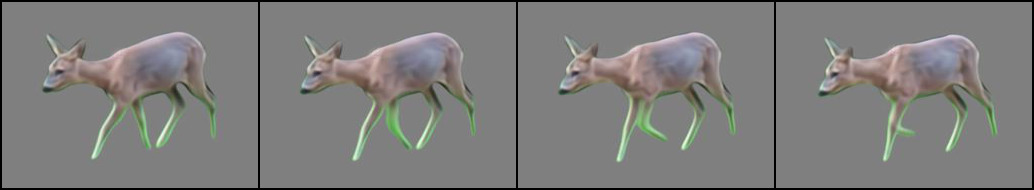}} &
\raisebox{-.4\totalheight}{\includegraphics[width=0.45\textwidth]{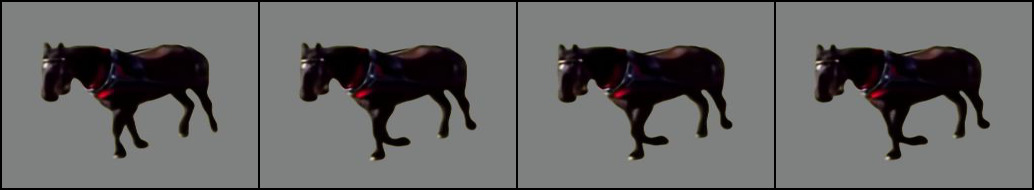}} \\
\end{tabular}
\caption{Intermediate results. Given input images $(1)$ from video $A$ (resp. $B$), we extract corresponding keypoints $(2)$, which are then translated to analogue shape of $B$ (resp. $A$) $(3)$, then the final texture is added $(4)$. Zoom-in is recommended.}
\vspace{-0.3cm}
\label{fig:inter}
\end{figure*}

We consider two input videos $A = \{a_i\}_{i=0}^{N_A-1}$ and $B = \{b_i\}_{i=0}^{N_B-1}$, where $a_i$ (resp. $b_i$) is the $i$th frame of $A$ (resp. $B$). WLOG, we wish to generate a new video $AB=\{ab_i\}_{i=0}^{N_B-1}$ portraying the analogues motion of $B$, while maintaining the appearance and style of $A$. We assume a single object is present in each video. To alleviate cases where the background rapidly changes, we mask out the background of each frame. The segmentation maps $\{s_{a,i}\}_{i=0}^{N_A-1}$,$\{s_{b,i}\}_{i=0}^{N_B-1}$ are given as part of the data, or are acquired using off-the-shelf image segmentation networks (see Sec.~\ref{sec:results}). We now describe the different components of our method. An illustration is provided in Figure~\ref{fig:method}, and a visual example of the intermediate results is shown in Figure~\ref{fig:inter}. For brevity, we describe the retargeting from $A$ (source) to $B$ (target), keeping in mind that the opposite direction is symmetrical.

\noindent\textbf{Shape-Invariant Representation.}\quad
To translate between objects of different shapes and texture, we use JOKR as a bottleneck. As manual keypoint annotation is not always available (see Section~\ref{sec:intro}), we use an unsupervised keypoint extractor $E$, similar to previous work ~\cite{suwajanakorn2018discovery, siarohin2019animating, siarohin2019first}, to extract $K$ keypoints, denoted $k_{a,i} = \{k_{a,i}^\ell\}^{K-1}_{\ell=0}$. To leverage the convolutional network's ability to utilize spatial information, we project the extracted keypoints to spatial maps by fitting a Gaussian for each keypoint, obtaining $K$ confidence maps $h_{a,i} = \{h_{a,i}^\ell\}^{K-1}_{\ell=0}$ 
(see Appendix~\ref{sec:impl} for more details).

To encourage the disentanglement between geometry and appearance, the generation process is divided into two steps. First, given $h_{a,i}$, generator $G_A$ is trained to output a silhouette that corresponds to the extracted keypoints and at the same time to the shape of the object in $A$. 
To reduce the number of parameters, $G_A$ and $G_B$ share the same weights, except for the last layer. Similarly, the same keypoint extractor $E$ is used for both videos. Formally, given frames ${a_{i}}$, ${b_{j}}$, we generate the silhouettes by minimizing the following MSE loss:
\begin{align}
\mathcal{L_{\text{seg}}} = \sum_{i=0}^{N_A - 1} \norm{G_A(E({a_{i}})) - s_{a,i}}_2 +
\sum_{j=0}^{N_B - 1} \norm{G_B(E({b_{j}})) - s_{b,j}}_2
\end{align}
We now train generators $R_A$ and $R_B$ to translate the obtained segmentation map to the original image, thereby adding texture. Specifically we consider the following reconstruction and perceptual losses:
\begin{gather}
\mathcal{L}_{L1} = \sum_{i=0}^{N_A - 1} \norm{R_A(G_A(E({a_{i}}))) - {a_{i}}}_1 + \sum_{j=0}^{N_B - 1} \norm{R_A(G_B(E({b_{j}}))) - {b_{j}}}_1 \\ 
\mathcal{L_{\text{LPIPS}}} =
\sum_{i=0}^{N_A - 1} \norm{\mathcal{F}(R_A(G_A(E({a_{i}})))) - \mathcal{F}({a_{i}})}_2 +
\sum_{j=0}^{N_B - 1} \norm{\mathcal{F}(R_A(G_B(E({b_{j}})))) - \mathcal{F}({b_{j}})}_2
\end{gather}
$\mathcal{L}_{L1}$ uses the $L1$ norm and $\mathcal{L_{\text{LPIPS}}}$ uses the LPIPS perceptual loss proposed by Zhang et al.~\cite{zhang2018unreasonable}, where the feature extractor is denoted by $\mathcal{F}$.

\noindent\textbf{Shared representation.}\quad
Videos $A$ and $B$ may depict objects from a different domain, and so the extracted keypoints for each video may have a different semantic meaning. For example, the same keypoint might represent a leg in one video and tail in the other. Therefore, we enforce that the encoded keypoints for both videos are from a shared distribution, thus encouraging the keypoints to capture motion that is common to both videos. The specific style of $A$ or $B$ is then encoded in the generator's weights. To enforce the shared distribution between the keypoints, we use a domain confusion loss~\cite{tzeng2014deep}. Specifically, a discriminator $D$ is used to distinguish between keypoints of domains $A$ and those of $B$, while the encoder is trained adversarially to fool the discriminator, thus forcing the keypoints of the two domains to statistically match:
\begin{align}
\mathcal{L}_{\text{DC}} =
\sum_{i=0}^{N_A - 1} \ell_{\text{bce}}(D(k_{a,i}),1) + \sum_{j=0}^{N_B - 1} \ell_{\text{bce}}(D(k_{b,j}),1) \label{eq:confusion_1}
\end{align}
where we use the binary cross entropy loss function $\ell_{\text{bce}}(p,q) = -(q\log(p) + (1-q)\log(1-p))$. While the keypoint extractor $E$ attempts to make the keypoints distributions indistinguishable, the discriminator is trained adversarially using the objective function:
\begin{align}
\mathcal{L}_{\text{D}} = \sum_{i=0}^{N_A - 1} \ell_{\text{bce}}(D(k_{a,i}),0) +
\sum_{j=0}^{N_B - 1} \ell_{\text{bce}}(D(k_{b,j}),1) \label{eq:confusion_2}
\end{align}
In some cases, the object appearing in video $A$ can be of different proportions or location in the images, which makes it already distinguished from $B$, just based on a change in rotation, scale or translation. 
We therefore augment Eq.~\ref{eq:confusion_1} and Eq.~\ref{eq:confusion_2} so that our domain confusion loss is invariant to affine transformations, thus enabling a broader variety of possible retargeting pairs. To do so, a learned affine transformation $T_A$ is applied to $B$'s keypoints before passing them to discriminator $D$, where $T_A$ is optimized with the keypoint extractor $E$.

\noindent\textbf{Temporal Coherence.}\quad
We would like to ensure that the generated videos are temporally coherent. That is, a smooth and non-jittery motion is generated. To this end, we apply a temporal regularization on the generated keypoints and minimize the distance between keypoints in adjacent frames:
\begin{align}
\mathcal{L}_{\text{tmp}} = 
\sum_{i=0}^{N_A - 1} \norm{k_{a,i} - k_{a,i+1}}_2 + \sum_{j=0}^{N_B - 1} \norm{k_{b,j} - k_{b,j+1}}_2
\end{align}
Since JOKR is encoded for every frame only from its respective image, sometimes flickering is introduced because a keypoint shifts in meaning between frames (e.g., a keypoint describing a back leg suddenly describes the tail). We observe this usually happens when the figures undergo large motion. Hence, similarly to Siarohin et al.,~\cite{siarohin2019first}, we ensure that the generated keypoints are equivariant under an arbitrary affine transformation; We apply a random affine transformation on the keypoints and the original frame, and compare the transformed keypoints with the keypoints extracted from the transformed image. This ensures the semantic meaning of each keypoint is consistent, and significantly improves coherency, since decoding temporally coherent keypoints results in temporally coherent frames. 
For an affine transformation $T$, transformation equivariance loss is defined as: 
\begin{align}
\mathcal{L}_{\text{eq}} = 
\sum_{i=0}^{N_A - 1} \norm{T(E({a_{i}})) - E(T({a_{i}}))}_1 + \sum_{j=0}^{N_B - 1} \norm{T(E({b_{j}})) - E(T({b_{j}}))}_1
\end{align}

\noindent\textbf{Keypoints Regularization.}\quad
Ensuring a shared representation and temporal coherence is important, but not sufficient to ensure the encoded keypoints capture meaningful information about motion. 
Specifically, the keypoints might collapse to a single point without any relation to the object itself. Therefore, we suggest additional two loss terms based on the terms used by Suwajanakorn et al.~\cite{suwajanakorn2018discovery}. First, we use a separation loss which prevents the keypoints sharing the same location, by penalizes two keypoints if they are closer than some hyperparameter threshold $\delta$: 
\begin{align}
\mathcal{L}_{\text{sep}} = \frac{1}{K^2} \sum^{K-1}_{\ell=0} \sum_{\ell\neq r} \left( \sum_{i=0}^{N_A - 1} \max (0, \delta - \norm{k_{a,i}^\ell - k_{a,i}^r}^2) + \sum_{j=0}^{N_B - 1} \max (0, \delta - \norm{k_{b,j}^\ell - k_{b,j}^r}^2) \right) \nonumber
\end{align}
where $k_{a,i} = \{k_{a,i}^\ell\}^{K-1}_{\ell=0}$ are the extracted keypoints from frame ${a_{i}}$. Second, we use silhouette loss to encourage the keypoints to lie on the object itself:
\begin{align}
\mathcal{L}_{\text{sill}} = \frac{1}{K} \sum^{K-1}_{\ell=0} \left( \sum_{i=0}^{N_A - 1} -\log \sum_{u,v} s_{a,i}(u,v) H_{a,i}^\ell(u,v) + \sum_{j=0}^{N_B - 1} -\log \sum_{u,v} s_{b,j}(u,v) H_{b,j}^\ell(u,v) \right) \nonumber
\end{align}
where the sum $\sum_{u,v}$ is over all image pixels $(u,v)$ and $H_{a,i}^\ell$ is the heatmap generated by the keypoint extractor for the $\ell$th keypoint from frame ${a_{i}}$ (see implementation details in Appendix~\ref{sec:impl}). Without this loss, the representation might focus on meaningless regions of the image, rendering some of the points irrelevant.

\noindent\textbf{Two-Steps Optimization.}\quad
Since most loss terms are related to shape and not to texture, we can optimize our objective function in two steps. First, we train the discriminator using $\mathcal{L}_{\text{D}}$, while also training the networks $E$, $G_A$, and $G_B$. In the second step, we train the refinement networks $R_A$, $R_B$ to add the texture, using the aforementioned $\mathcal{L}_{L1}$ and $\mathcal{L}_{\text{LPIPS}}$ losses. For challenging textures, such as in Fig.~\ref{fig:giraffe} (bottom), we employ additional adversarial loss over the generated frames at the second stage training (texture), using the discriminator and adversarial loss proposed by Wang et al.~\cite{pix2pixHD}.

\noindent\textbf{Augmentations.}\quad
Using adversarial loss with very limited data might cause mode collapse. Thus, we use augmentations in the form of random affine transformations during training. However, when preserving the background is necessary as in Fig.~\ref{fig:giraffe} (Bottom), these augmentations might leak to the generated frames resulting with artifacts. Hence, similar to ~\cite{zhao2020image, karras2020training}, we perform the augmentations directly over the keypoints just before passing them to the discriminator, resulting with less artifacts and stable training.

\noindent\textbf{Inference.}\quad
Since the distribution of keypoints match, we can use $G_A$, $R_A$ to translate the keypoints of $B$ to the appearance and style of $A$, while also adhering to the motion of $B$: ${ab}_j = R_A(G_A(T(E({b_{j}}))))$. Note that the learned affine transform $T$ may be omitted. Using $T$, the generated object will more faithfully preserve the target object's rotation, scale and translation. Without it, the source object's rotation, scale and translation will be better preserved (See Fig.~\ref{fig:abl_affine}).

\begin{figure*}

\centering
\begin{tabular}{ll}

~~~~~~~~~~~~~~~~~~~~~~~~~ $t$ ~~~~~~~~~~ $t+5$  ~~~~~ $t+10$  ~~~~~ $t+15$  & 
~~~~ $t$ ~~~~~~~~~~ $t+5$  ~~~~~ $t+10$  ~~~~~ $t+15$ \\

\hspace{-0.5cm} \begin{tabular}{l} \noalign{\vskip -4mm} \small{Input} \\ \noalign{\vskip 7mm}  \small{Ours}\\ \noalign{\vskip 7mm} \small{FOMM~\cite{siarohin2019first}} \\ \noalign{\vskip 7mm} Cycle~\cite{CycleGAN2017}\\  \noalign{\vskip 7mm} \small{ReCycle~\cite{bansal2018recycle}} \\ \end{tabular}   
\raisebox{-.45\totalheight}{\includegraphics[width=0.42\textwidth]{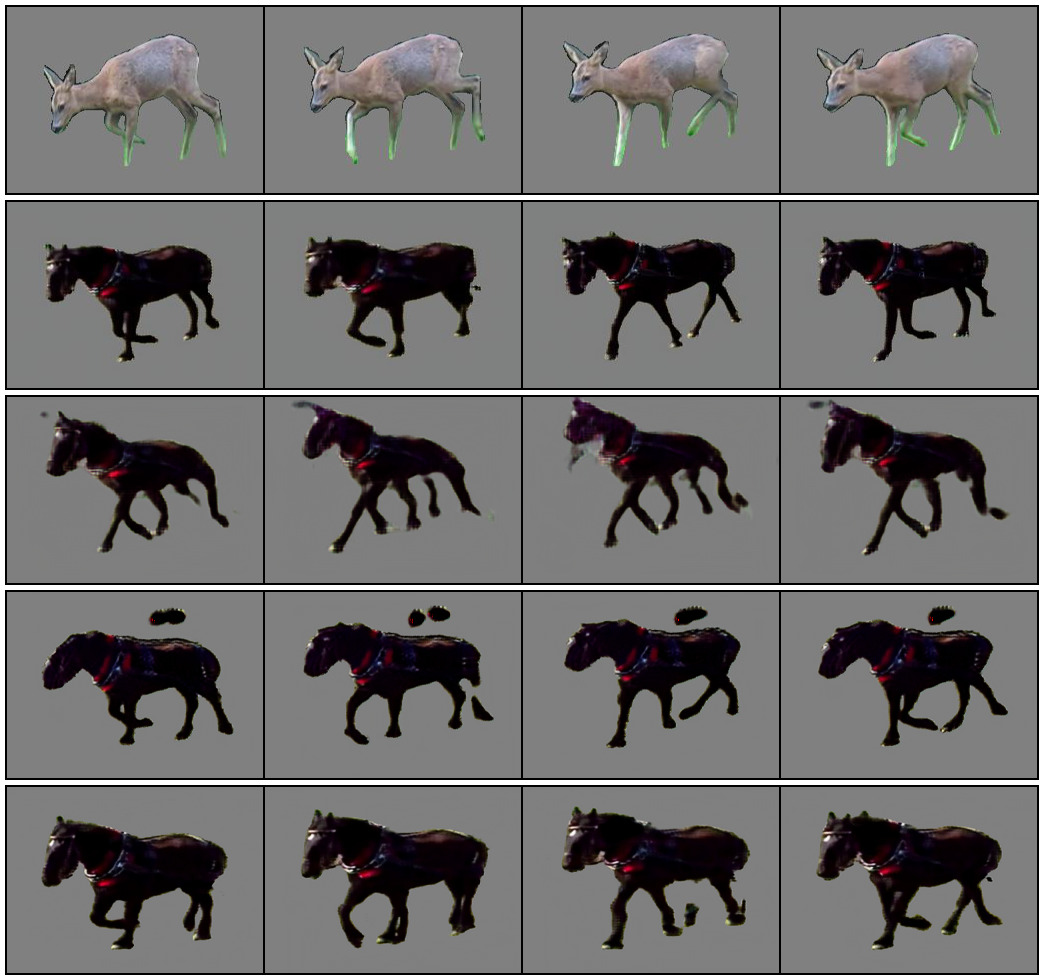}} &
\hspace{-0.3cm}
\raisebox{-.45\totalheight}{\includegraphics[width=0.42\textwidth]{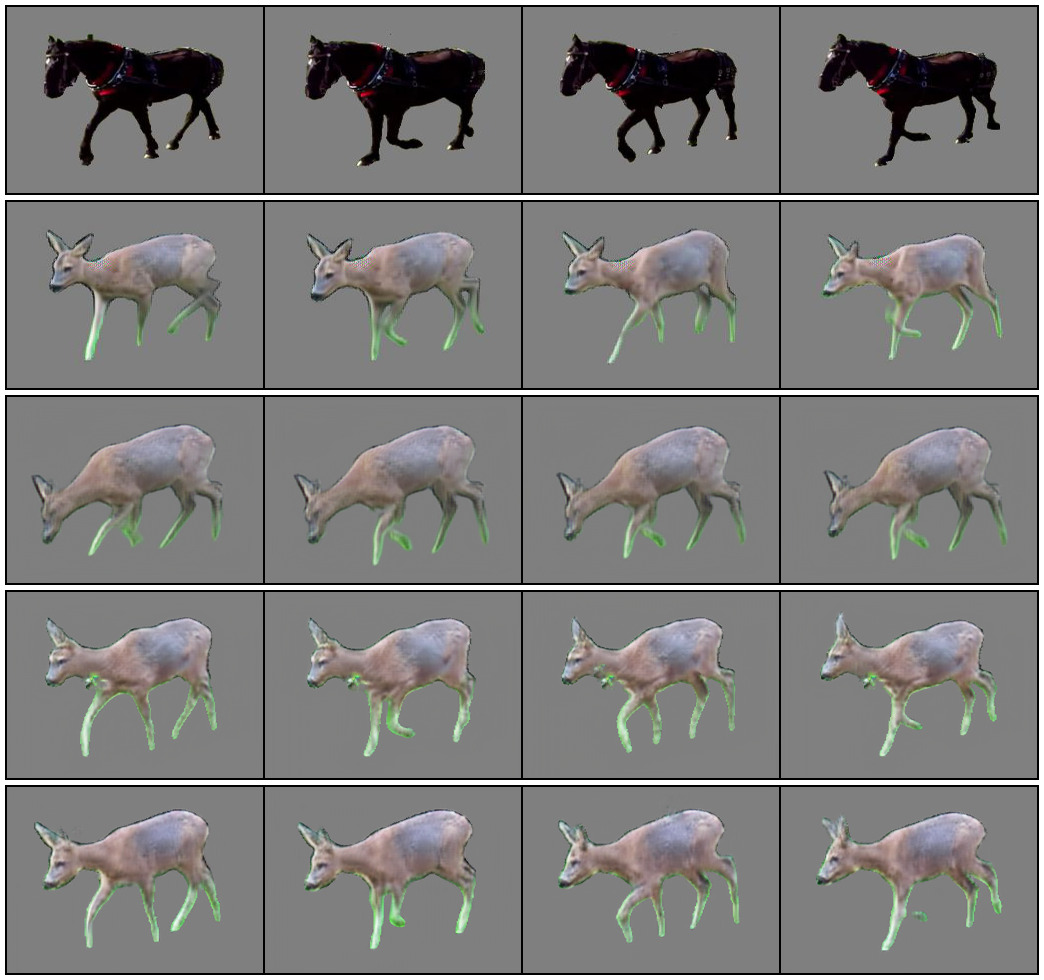}} \\
\noalign{\vskip 0.5mm} 
\hspace{-0.5cm} \begin{tabular}{l} \noalign{\vskip -3mm} \small{Input} \\ \noalign{\vskip 5mm}  \small{Ours}\\ \noalign{\vskip 5mm} \small{FOMM~\cite{siarohin2019first}} \\ \noalign{\vskip 5mm} Cycle~\cite{CycleGAN2017}\\  \noalign{\vskip 5mm} \small{ReCycle~\cite{bansal2018recycle}} \\ \end{tabular}   

\raisebox{-.45\totalheight}{\includegraphics[width=0.42\textwidth]{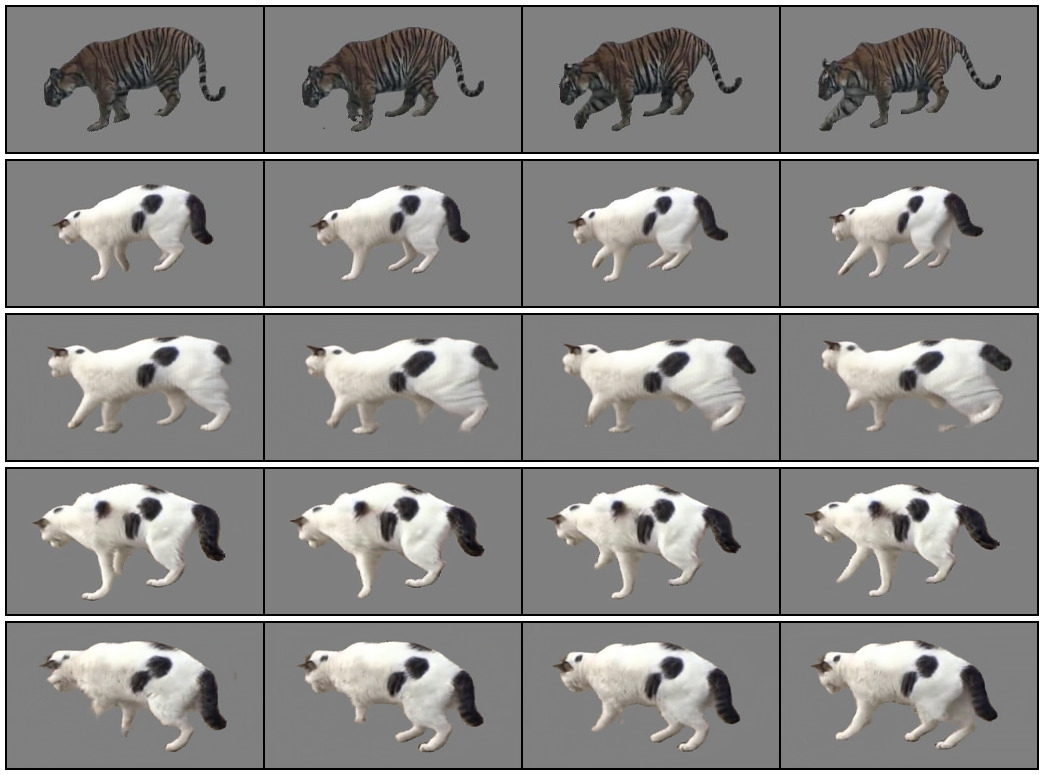}} &
\hspace{-0.3cm}
\raisebox{-.45\totalheight}{\includegraphics[width=0.42\textwidth]{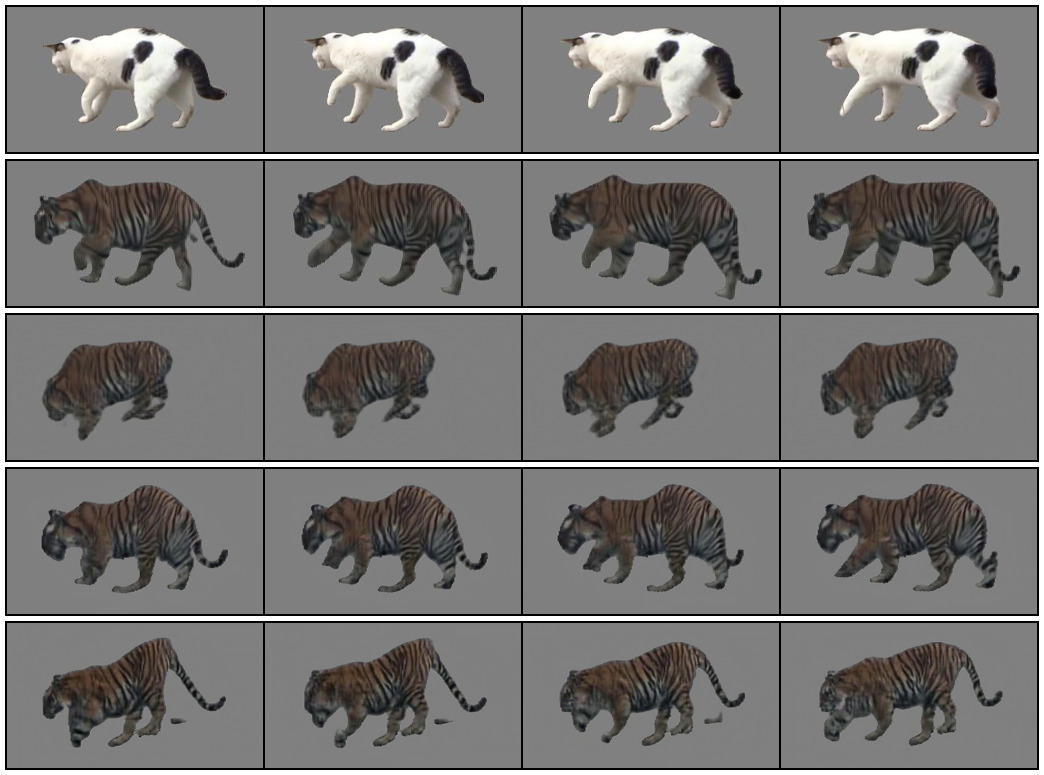}} \\

\end{tabular}

\caption{A comparison of our method to baselines. As can be seen, ours is the only one to transfer the correct pose while faithfully generating the target style. Zoom-in recommended. 
} 

\vspace{-0.5cm}
\label{fig:comp}
\end{figure*}

\begin{table*}
\begin{center}

\begin{tabular}{lcccc}
\toprule
Measure & \textbf{Ours}  & FOMM~\cite{siarohin2019first}  &  CycleGAN~\cite{CycleGAN2017} &  RecycleGAN~\cite{bansal2018recycle} \\  
\midrule
FID $\downarrow$ &  $\mathbf{39.01}$ & $74.35$ &  $86.41$ & $55.0$ \\
SVFID $\downarrow$ &  $\mathbf{294.89}$ & $345.1$ &  $349.43$ & $317.84$ \\
User Study - Appearance $\uparrow$ &   $\mathbf{3.91}$ & $2.07$ & $2.64$ & $2.27$ \\
User Study - Motion $\uparrow$ &  $3.27$ & $1.57$ & $\mathbf{3.61}$ & $2.75$ \\

\bottomrule
\end{tabular}
\caption{Numerical evaluation for our method and baselines. FID and SVFID are used to measure the temporal realism of generated videos (lower is better). A user study is used to analyze the appearance and motion consistency, measured by mean opinion score (higher is better).
}
\vspace{-0.5cm}
\label{tab:quant} 
\end{center}
\end{table*}

\vspace{-0.1cm}
\section{Results}
\label{sec:results}
\vspace{-0.1cm}

We evaluate our method both qualitatively and quantitatively on video pairs of walking animals from the YouTube-VOS dataset~\cite{xu2018youtube}. All pairs are challenging as they contain different shapes such as a cat/fox or a deer/horse. To demonstrate versatility, we also present results for humans and flowers, and for synchronization of short GIFs. For segmentation, we used ground-truth if available or extract it using a pretrained network, see more details in Appendix~\ref{sec:impl}. In addition, we show that our learned keypoints have semantic meaning by performing simple editing. Lastly, an ablation analysis is performed to illustrate the effectiveness of the different components. For all experiments, videos can be found at our webpage: \url{https://rmokady.github.io/JOKR/}.

\noindent\textbf{Qualitative evaluation.}\quad
We consider several baselines most related to ours. First, we consider FOMM~\cite{siarohin2019first}, which, similarly to ours, learn a keypoint representation in an unsupervised fashion. 
A second baseline is that of CycleGAN~\cite{CycleGAN2017}, where a cycle loss is used to perform unsupervised image-to-image translation for every frame. Since CycleGAN is normally trained on many images, we perform augmentations to avoid overfitting. Lastly, we compare to ReCycleGAN~\cite{bansal2018recycle}, which extends CycleGAN by employing additional temporal constraints. Fig.~\ref{fig:comp} gives a visual comparison of our method (additional results are in Fig.~\ref{fig:comp1} to Fig.~\ref{fig:comp6}, videos can be found at the webpage). As can be seen, for both examples in Fig.~\ref{fig:comp}, our method correctly transfers the leg movement present in the source video, while preserving the style of the target. For instance, the horse legs appear realistic and are not of the same shape as the input deer. As can be seen, FOMM is unable to produce realistic target shapes, as the warping mechanism matches the keypoint locations themselves, without considering the different semantic meanings of the two domains. CycleGAN struggles in changing the deer's shape to a realistic horse shape. ReCycleGAN manages to alter the shape correctly, but suffers from significant artifacts, such as a missing leg.

\begin{figure*}[ht]
\centering
\begin{tabular}{lll}
& ~~~~~ $t$ ~~~~~~~~$t+5$ ~~~~ $t+10$  ~~ $t+15$ ~~ $t+20$
& ~~~~~ $t$ ~~~~~~~~$t+5$ ~~~~ $t+10$  ~~ $t+15$ ~~ $t+20$\\

\rotatebox[origin=t]{90}{Input}   &
\raisebox{-.4\totalheight}{\includegraphics[width=0.45\textwidth]{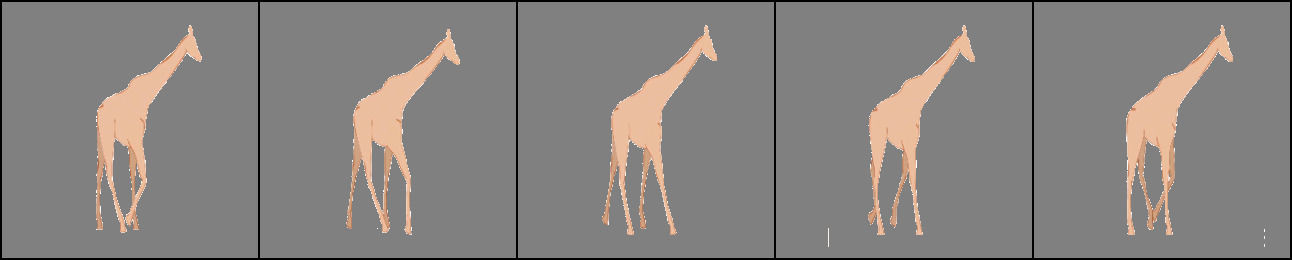}} &
\raisebox{-.4\totalheight}{\includegraphics[width=0.45\textwidth]{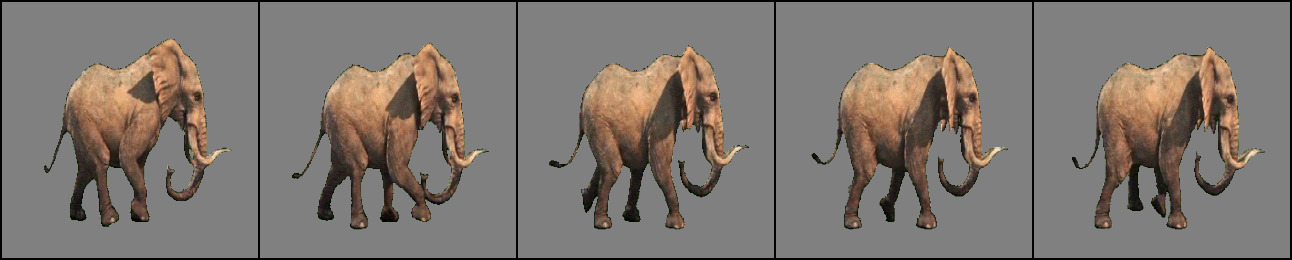}} \\
\rotatebox[origin=t]{90}{Ours}    &
\raisebox{-.4\totalheight}{\includegraphics[width=0.45\textwidth]{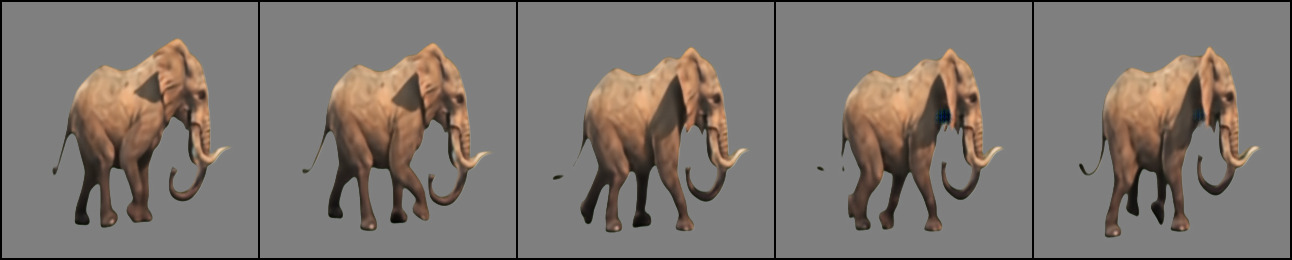}} &
\raisebox{-.4\totalheight}{\includegraphics[width=0.45\textwidth]{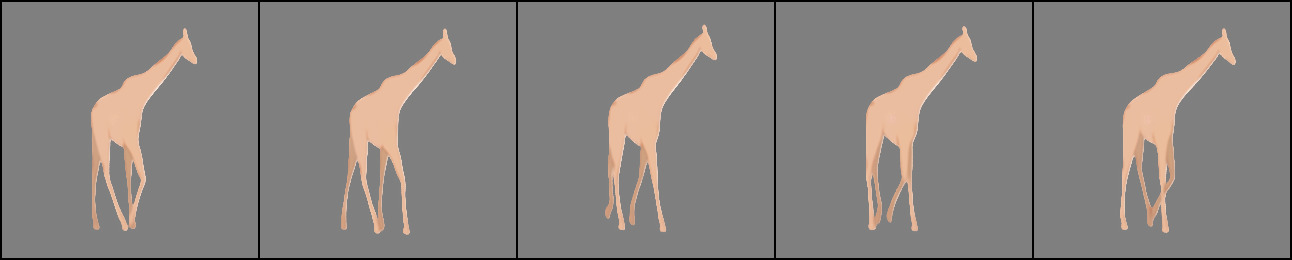}} \\

\noalign{\vskip 0.5mm} 
\rotatebox[origin=t]{90}{Input}   &
\raisebox{-.3\totalheight}{\includegraphics[width=0.45\textwidth]{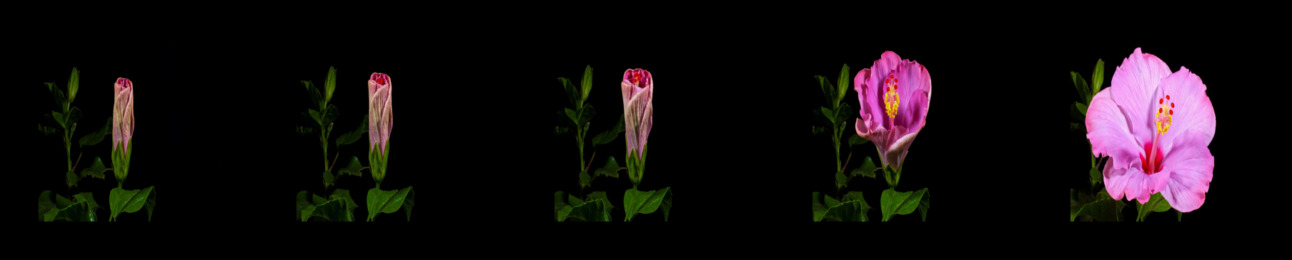}} & \raisebox{-.3\totalheight}{\includegraphics[width=0.45\textwidth]{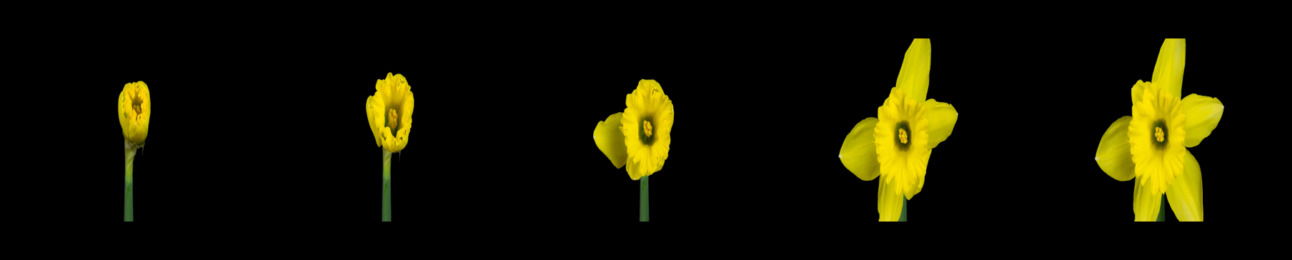}}  \\
\noalign{\vskip 0.5mm} 
\rotatebox[origin=t]{90}{Ours}   &
\raisebox{-.3\totalheight}{\includegraphics[width=0.45\textwidth]{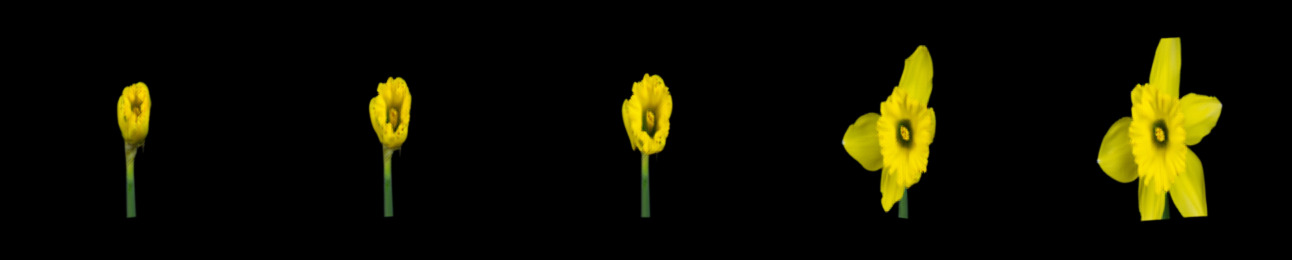}} &
\raisebox{-.3\totalheight}{\includegraphics[width=0.45\textwidth]{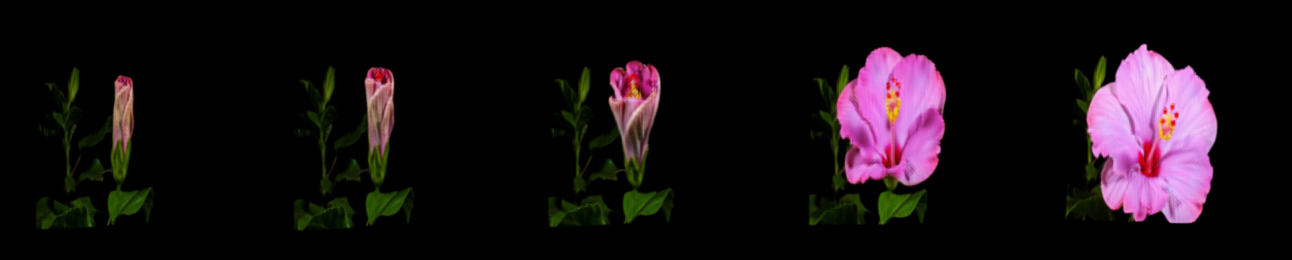}} \\

\noalign{\vskip 0.5mm} 
\rotatebox[origin=t]{90}{EDN ~~~~ Ours ~~~~ Input}   &
\raisebox{-.425\totalheight}{\includegraphics[width=0.45\textwidth]{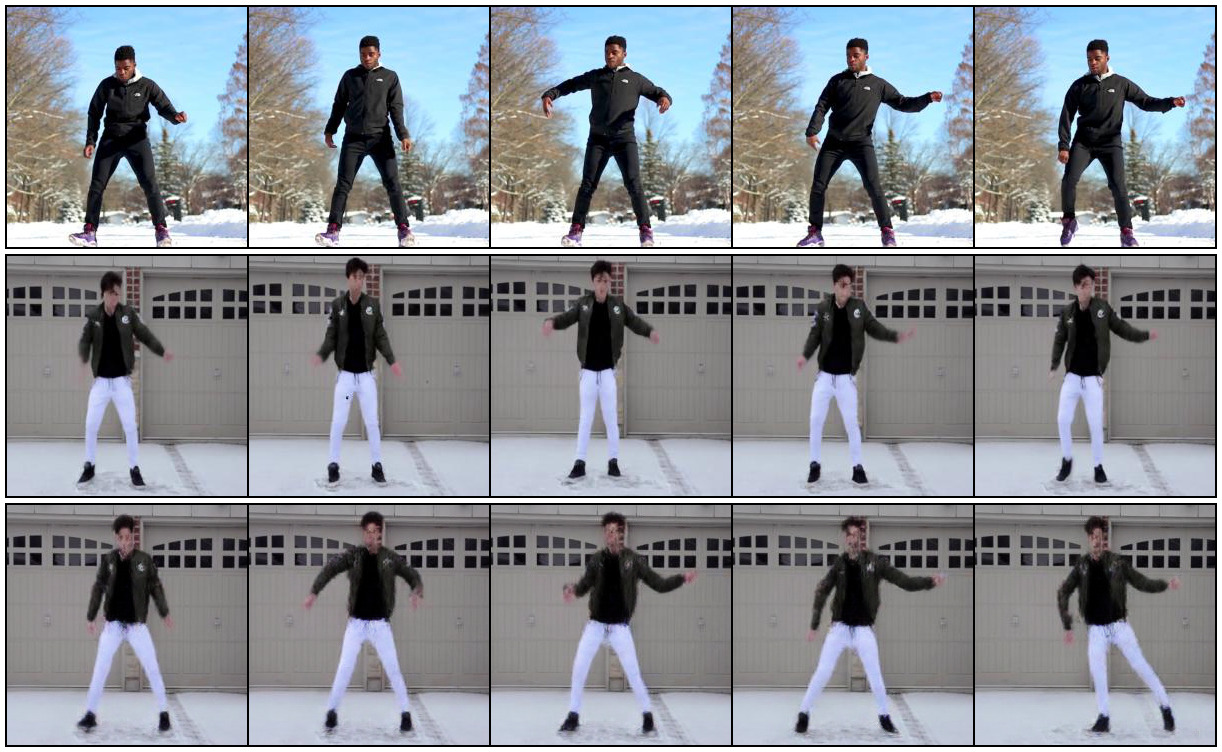}} & \raisebox{-.425\totalheight}{\includegraphics[width=0.45\textwidth]{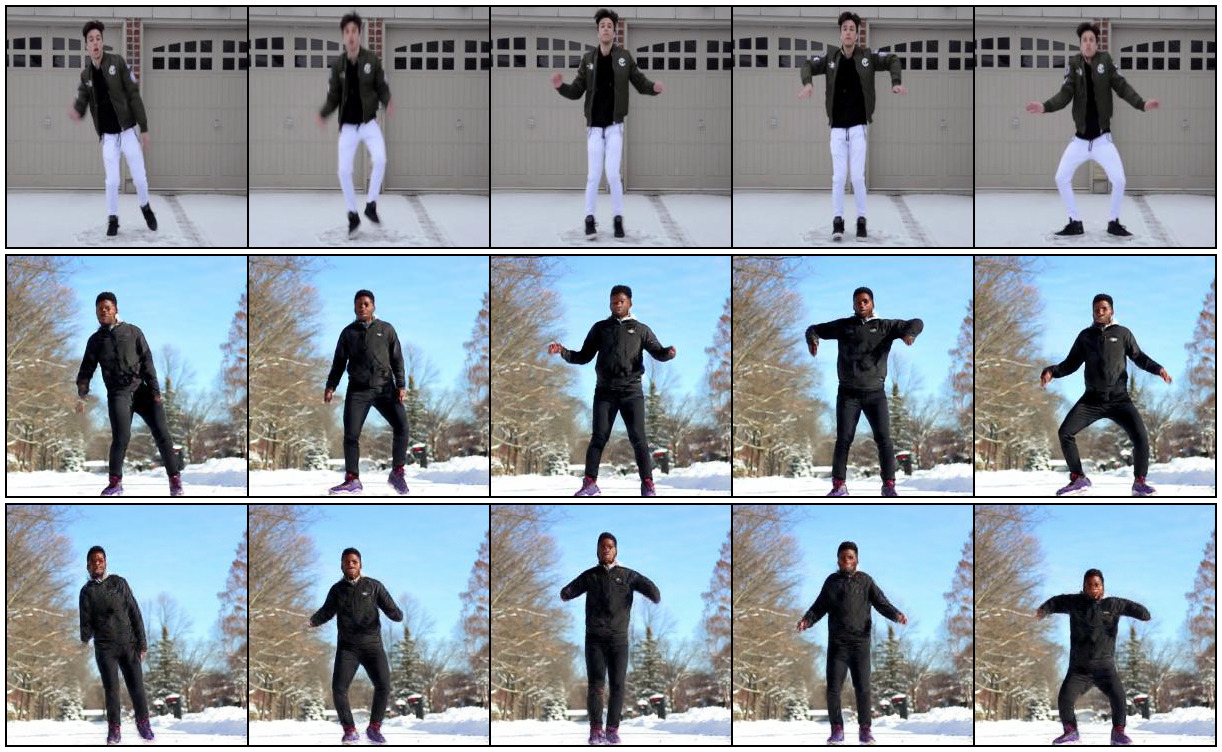}}  \\
\end{tabular}
\caption{Top: GIF Synchronization for a video pair of a zebra and a giraffe. 
Middle: Two videos of blooming flowers. Bottom: Pairs of dancing persons videos with comparison to EDN~\cite{chan2019everybody}. 
}
\vspace{-0.5cm}
\label{fig:giraffe}
\end{figure*}

\noindent\textbf{Quantitative evaluation.}\quad
To evaluate the realism of generated frames, we use the FID metric~\cite{fid} over each one. For temporal consistency, we adopt the recently proposed SVFID score introduced by Gur et al.~\cite{gur2020hierarchical}. 
SVFID is an extension of FID for a single video, evaluating how the generated samples capture the temporal statistics of a single video, by using features from a pretrained action recognition network. 
We compare video $AB$ (resp. $BA$) and video $A$ (resp. $B$), and report a superior result (Tab.~\ref{tab:quant}). 
To evaluate the quality of the motion transfer, we performed a user study, as we do not have ground truth keypoint supervision. 
For each video pair, users were asked to rank from 1 to 5: (1) How much the appearance of $AB$ looks realistic compared to $B$? (User Study - Appearance) and (2) How similar is the motion of $A$ to that of $AB$? (User Study - Motion). As can be seen, our generated videos are much more realistic than all baselines. 
While CycleGAN results are better for motion transfer, it scores significantly worse both in appearance and on FID/SVFID scores. The reason is that CycleGAN struggles with geometric changes and focuses on the texture. For example, the horse in Fig~\ref{fig:comp} is well aligned but bears very different appearance from the original horse.

\noindent\textbf{GIF synchronization.}\quad
We demonstrate our ability to handle very modest data, through the GIF synchronization setting, where the number of different frames is approximately $40$. We use video pairs, depicting different objects from the MGif dataset~\cite{siarohin2019animating}. Synchronization of an elephant video and a giraffe video is shown in Fig.~\ref{fig:giraffe} where our method successfully generates an analogue motion. In addition, we successfully keep the target object's appearance, even when the shapes and texture of source and target videos are different. For example, the same leg moves forward at the same time for both input and respective generated sequence, while the way the leg moves (in terms of length, bendiness, and stride size) is preserved well. As motion range is limited, our generated frames are similar to the input ones, hence we define this result as synchronization. Additional result in Fig.~\ref{fig:elk}.

\noindent\textbf{Other domains}\quad
Fig.~\ref{fig:giraffe} (Middle) presents blooming flowers as used in Bansal et al.~\cite{bansal2018recycle}, and Fig.~\ref{fig:giraffe} (Bottom) presents dancers. For the dancers, we compare against EDN~\cite{chan2019everybody}. The latter assumes the supervision of a semantic skeleton, and so our setting is more challenging. Further, we use very short (less than a minute) videos that consist of a wide range of different motions. As can be seen, our method is comparable to EDN. Additional results are in Fig~.\ref{fig:fl}, Fig~.\ref{fig:edn}, and Fig~.\ref{fig:edn2}

\begin{figure*}[h]
\centering
\vspace{-0.2cm}
\begin{tabular}{l}

~~~~~~ $\leftarrow$ ~~~~~~~~~~~~~~~~~~~~~~~~~~~~~~~~~~~~~~~~~~~~~~~~~~~~~ Original  ~~~~~~~~~~~~~~~~~~~~~~~~~~~~~~~~~~~~~~~~~~~~~~~~~~~~~~ $\rightarrow$ ~~~~ \\
\raisebox{-.5\totalheight}{\includegraphics[width=.9\textwidth]{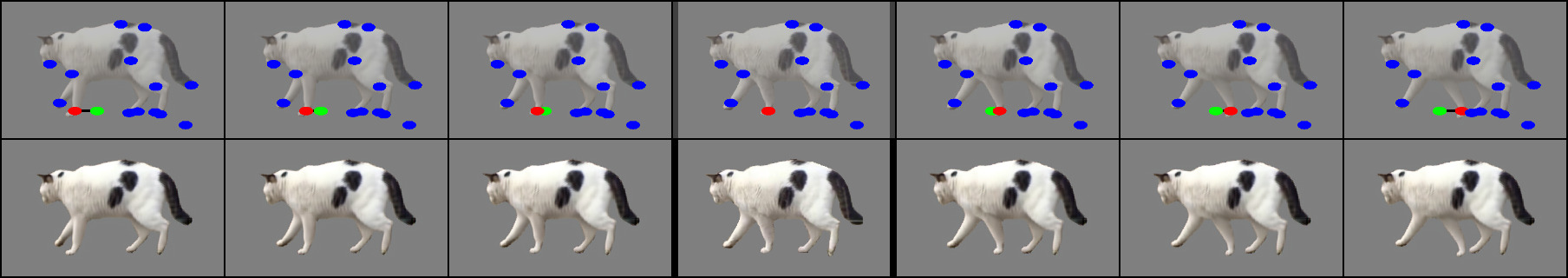}} \\

~~~~~~ $\downarrow$ ~~~~~~~~~~~~~~~~~~~~~~~~~~~~~~~~~~~~~~~~~~~~~~~~~~~~~ Original  ~~~~~~~~~~~~~~~~~~~~~~~~~~~~~~~~~~~~~~~~~~~~~~~~~~~~~~ $\uparrow$ ~~~~ \\
\raisebox{-.5\totalheight}{\includegraphics[width=.9\textwidth]{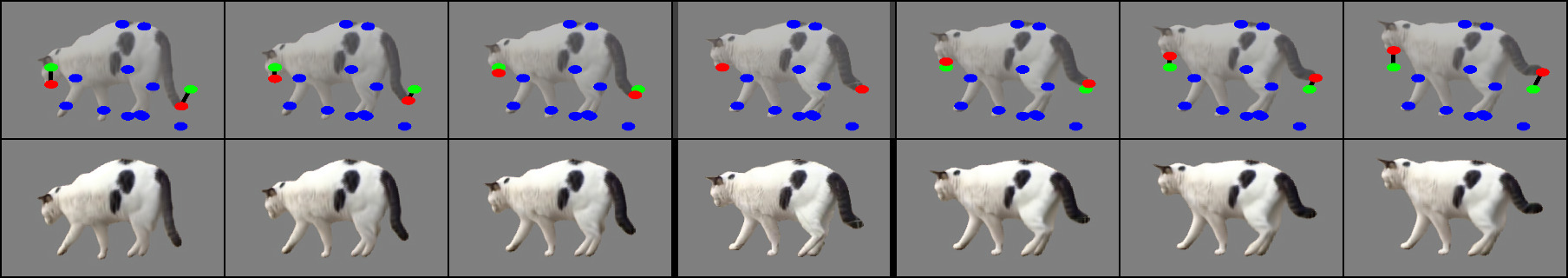}} \\

\end{tabular}
\caption{
Moving the keypoints linearly induces semantic editing. All keypoints are colored in blue, while the original location of the edited keypoints in green and their new location in red. Top: We manipulate the front leg by moving the corresponding keypoint right and left. Bottom: We now move both the head and the tail up and down, where additional small movement right and left (resp.) is applied.  As the red points move, we get a corresponding movement in the generated frame. }
\label{fig:editing}
\end{figure*}

\noindent\textbf{Keypoint Interpretability.}\quad
To demonstrate the semantic meaning of our generated keypoints, we perform a simple, yet effective, editing procedure of manual manipulation of the obtained keypoints. For a given real frame $a_i$, we select one or two of the keypoints generated using $E(a_i)$ (see Sec.~\ref{sec:results}). We then move these keypoints linearly and generate the corresponding frame by passing the new set of keypoints to the generators. Results presented in Fig.~\ref{fig:editing} demonstrate that this type of editing yields semantic manipulation of the given object (additional results in Fig.~\ref{fig:editing_supp}).

\noindent\textbf{Ablation Study.}\quad
\label{sec:ablation}
An ablation analysis is presented in Fig.~\ref{fig:abl} for a horse and deer: $(a)$ Original horse frames. $(b)$ Results obtained by the motion of the horse with the style of the deer. $(c)$ The equivariance loss is omitted, resulting in inconsistent motion, such as a separated leg. $(d)$ Without augmentations overfitting occur, resulting in significant artifacts. $(e)$ The domain-confusion loss is omitted, causing the keypoints to contain vital information about the shape, resulting in the horse shape and the deer's texture. $(f)$ Omitting the temporal regularization decreases temporal coherency and sometimes causes additional artifacts such as a missing head. $(g)$ Avoiding the two-step approach, that is the result is generated directly from the keypoint, without generating the shape first. As can be seen, in this case the appearance is inferior. 

We show additional ablations for temporal regularization in Fig.~\ref{fig:abl_temp}, focusing on the keypoint locations with and without regularization. Without regularization $(b)$,$(c)$ the keypoints movement from one frame to another is not proportional to the object's movement, resulting in inferior temporal consistency compared to using the regularization $(d)$,$(e)$. Video can be found at the webpage. We also measure the normalized distance between the same keypoints across adjacent frames. As expected, the distance without temporal regularization ($0.0055$) is substantially higher than using the regularization ($0.0023$). 

Lastly, we demonstrate the effect of the learned affine transformation in Fig~\ref{fig:abl_affine}. Given a video of a horizontal cat $(a)$, and rotated tiger $(b)$, the learned affine transformation results in rotating and scaling the object. Therefore, applying the transformation at inference $(e)$ resulting in a horizontal cat as the transformation fixes the rotation. Without applying it at inference $(d)$, the result is perfectly aligned to the source video depicting the exact rotation and scale. Omitting the learned affine transformation at training $(c)$ reduces the stability of the domain confusion loss, which sometimes results in artifacts.

\noindent\textbf{Limitations.}\quad
Looking forward, there is much room for further investigation. Currently, JOKR is agnostic to affine disproportions between the source and target videos. Beyond that, the retargeted videos should still bear some similarities, such as topology.  Second, while multiple objects can be handled separately, our method is unable to handle complicated scenes with multiple objects which may occlude each other. We also note that we ignore the background at the current scope of the paper. Requiring these segmentation maps implies supervision, however as we demonstrate, using off-the-shelf tools suffices in our case. Lastly, we note that training time is around 12 hours for a single NVIDIA GTX1080 GPU, and reducing training time could be helpful as future work.

\vspace{-0.1cm}
\section{Conclusion}
\vspace{-0.1cm}

We presented a method for unsupervised motion retargeting. Our key idea is to use a joint keypoint representation to capture motion that is common to both the source and target videos. JOKR demonstrates how imposing a bottleneck of geometric meaning, the aforementioned semantics are encouraged to adhere to geometric reasoning. Hence, poses that bear geometric similarities across the domains are represented by the same keypoints. Even more so, our editing experiments suggest that these semantics are intuitively interpretable. We demonstrate that such a representation can be used to retarget motion successfully across different domains, where the videos depict different shapes and styles, such as four-legged animals, flowers, and dancers. Through our GIFs example, we demonstrate that this representation can be jointly distilled even for short clips, where due to the low range of motion, the network reverts to simple synchronization successfully. Moving forward, challenges remain with videos depicting moving backgrounds, multiple objects, or different topology.

\clearpage

\bibliographystyle{plain}
\bibliography{gans}

\clearpage
\appendix
\appendixpage

\section{Ablation Study}

\begin{figure*}

\centering
\begin{tabular}{lllll}

$(a)$   &
\raisebox{-.5\totalheight}{\includegraphics[width=0.2\textwidth]{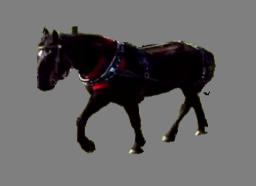}} &
\raisebox{-.5\totalheight}{\includegraphics[width=0.2\textwidth]{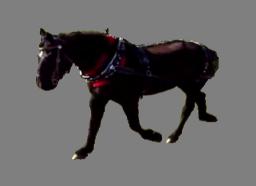}} &
\raisebox{-.5\totalheight}{\includegraphics[width=0.2\textwidth]{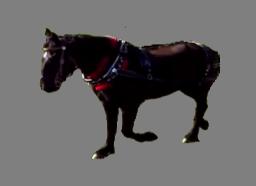}} &
\raisebox{-.5\totalheight}{\includegraphics[width=0.2\textwidth]{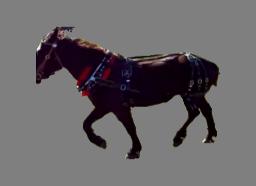}} \\
\noalign{\vskip 1mm} 

$(b)$   &
\raisebox{-.5\totalheight}{\includegraphics[width=0.2\textwidth]{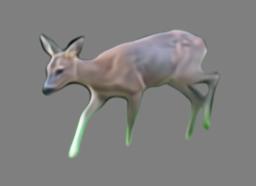}} &
\raisebox{-.5\totalheight}{\includegraphics[width=0.2\textwidth]{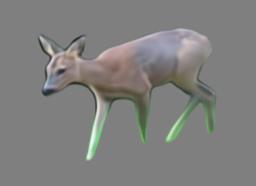}} &
\raisebox{-.5\totalheight}{\includegraphics[width=0.2\textwidth]{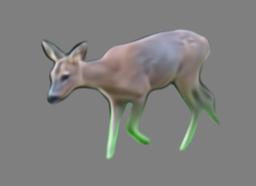}} &
\raisebox{-.5\totalheight}{\includegraphics[width=0.2\textwidth]{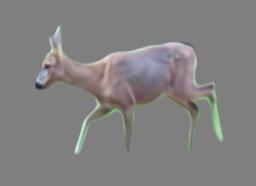}} \\
\noalign{\vskip 1mm} 

$(c)$   &
\raisebox{-.5\totalheight}{\includegraphics[width=0.2\textwidth]{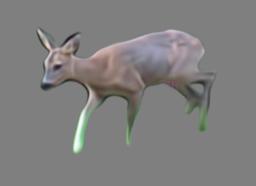}} &
\raisebox{-.5\totalheight}{\includegraphics[width=0.2\textwidth]{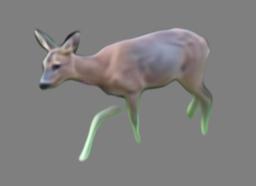}} &
\raisebox{-.5\totalheight}{\includegraphics[width=0.2\textwidth]{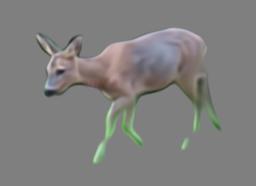}} &
\raisebox{-.5\totalheight}{\includegraphics[width=0.2\textwidth]{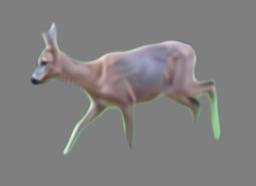}} \\
\noalign{\vskip 1mm}

$(d)$   &
\raisebox{-.5\totalheight}{\includegraphics[width=0.2\textwidth]{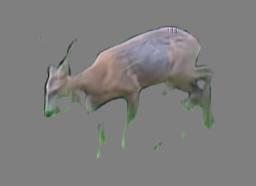}} &
\raisebox{-.5\totalheight}{\includegraphics[width=0.2\textwidth]{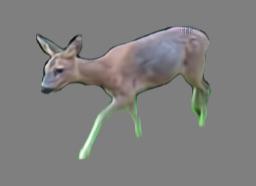}} &
\raisebox{-.5\totalheight}{\includegraphics[width=0.2\textwidth]{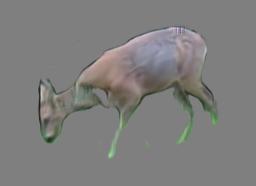}} &
\raisebox{-.5\totalheight}{\includegraphics[width=0.2\textwidth]{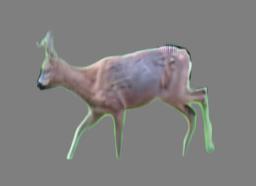}} \\
\noalign{\vskip 1mm} 

$(e)$   &
\raisebox{-.5\totalheight}{\includegraphics[width=0.2\textwidth]{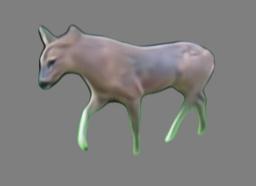}} &
\raisebox{-.5\totalheight}{\includegraphics[width=0.2\textwidth]{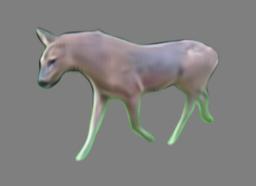}} &
\raisebox{-.5\totalheight}{\includegraphics[width=0.2\textwidth]{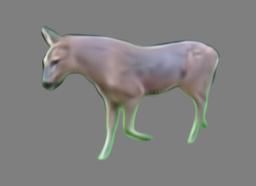}} &
\raisebox{-.5\totalheight}{\includegraphics[width=0.2\textwidth]{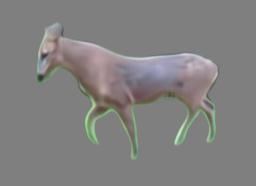}} \\
\noalign{\vskip 1mm} 

$(f)$   &
\raisebox{-.5\totalheight}{\includegraphics[width=0.2\textwidth]{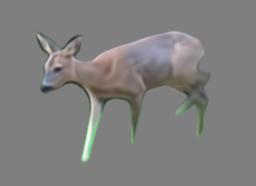}} &
\raisebox{-.5\totalheight}{\includegraphics[width=0.2\textwidth]{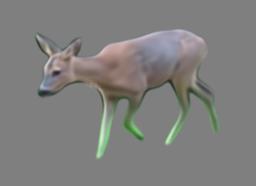}} &
\raisebox{-.5\totalheight}{\includegraphics[width=0.2\textwidth]{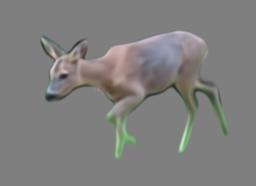}} &
\raisebox{-.5\totalheight}{\includegraphics[width=0.2\textwidth]{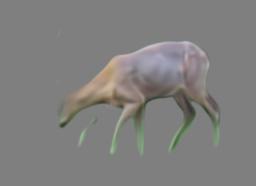}} \\
\noalign{\vskip 1mm}

$(g)$   &
\raisebox{-.5\totalheight}{\includegraphics[width=0.2\textwidth]{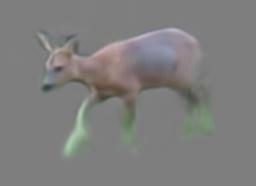}} &
\raisebox{-.5\totalheight}{\includegraphics[width=0.2\textwidth]{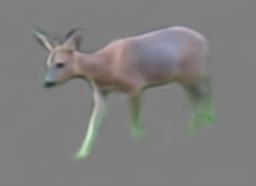}} &
\raisebox{-.5\totalheight}{\includegraphics[width=0.2\textwidth]{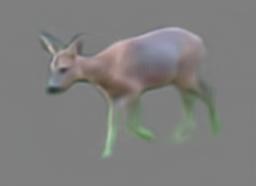}} &
\raisebox{-.5\totalheight}{\includegraphics[width=0.2\textwidth]{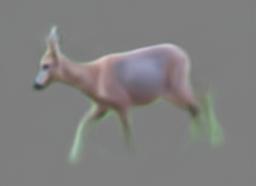}} \\

\end{tabular}

\caption{Ablation Study. $(a)$ original horse frames. $(b)$ our results. $(c)$ without equivariance loss. $(d)$ without augmentations. $(e)$ without domain-confusion loss. $(f)$ without temporal regularization. $(g)$ without two-step approach, i.e. generating the result directly from the keypoints.   }
\label{fig:abl}
\end{figure*}

\begin{figure*}[h]

\centering
\begin{tabular}{lll}

~ & ~~~~~~~~~~~~~ $t$ ~~~~ ~~~~~~~~~~~~~~~~~~~ $t+1$ &  ~~~~~~~~~ $t+10$  ~~~~~~~~~~~~~~~~~~~ $t+11$  \\

$(a)$   &
\raisebox{-.5\totalheight}{\includegraphics[width=0.4\textwidth]{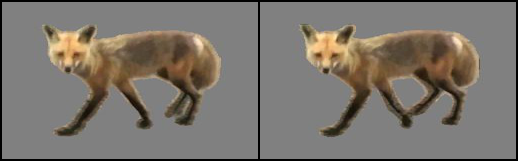}} & 
\raisebox{-.5\totalheight}{\includegraphics[width=0.4\textwidth]{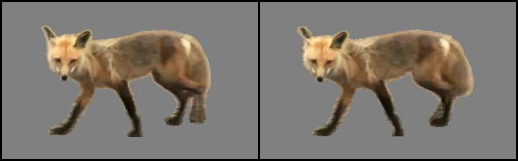}} \\
\noalign{\vskip 0.5mm} 
\hline
\noalign{\vskip 0.5mm} 
$(b)$   &
\raisebox{-.5\totalheight}{\includegraphics[width=0.4\textwidth]{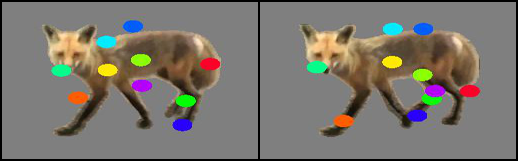}} & 
\raisebox{-.5\totalheight}{\includegraphics[width=0.4\textwidth]{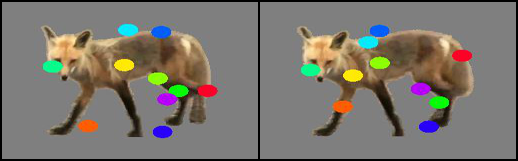}} \\
\noalign{\vskip 0.5mm} 
$(c)$   &
\raisebox{-.5\totalheight}{\includegraphics[width=0.4\textwidth]{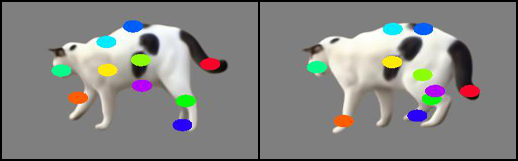}} &
\raisebox{-.5\totalheight}{\includegraphics[width=0.4\textwidth]{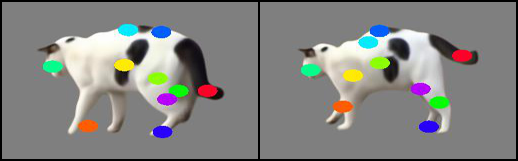}} \\
\noalign{\vskip 0.5mm} 
\hline
\noalign{\vskip 0.5mm} 
$(d)$   &
\raisebox{-.5\totalheight}{\includegraphics[width=0.4\textwidth]{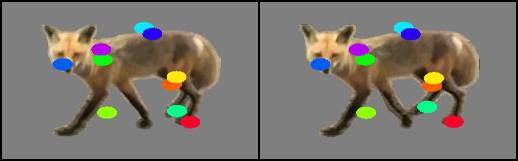}} & 
\raisebox{-.5\totalheight}{\includegraphics[width=0.4\textwidth]{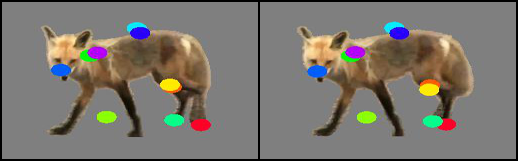}} \\
\noalign{\vskip 0.5mm} 
$(e)$   &
\raisebox{-.5\totalheight}{\includegraphics[width=0.4\textwidth]{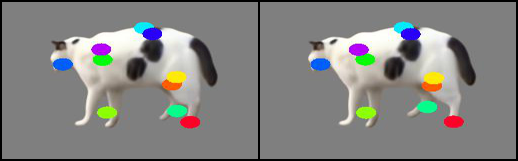}} &
\raisebox{-.5\totalheight}{\includegraphics[width=0.4\textwidth]{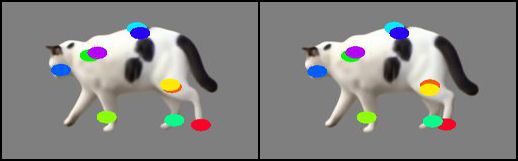}} \\

\end{tabular}

\caption{Ablation study for the temporal regularization. $(a)$ Input. $(b)$ Extracted keypoints without temporal regularization, presented on the top of the input frame. $(c)$ Result using the extracted keypoints without temporal regularization, presented with the keypoints. $(d)$ Extracted keypoints using temporal regularization, presented on the top of the input frame. $(e)$ Result using temporal regularization, presented with the extracted keypoints. As can be seen, without temporal consistency the keypoints move much further within adjacent frames, resulting in temporal inconsistency. }
\label{fig:abl_temp}
\end{figure*}

\begin{figure*}[h]

\centering
\begin{tabular}{ll}

$(a)$   &
\raisebox{-.5\totalheight}{\includegraphics[width=0.6\textwidth]{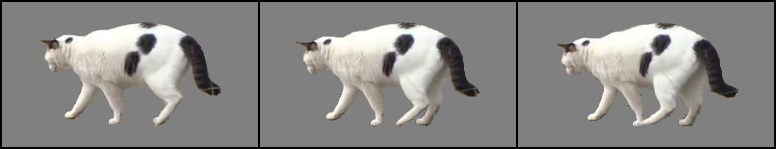}} \\
\noalign{\vskip 2mm} 

$(b)$   &
\raisebox{-.5\totalheight}{\includegraphics[width=0.6\textwidth]{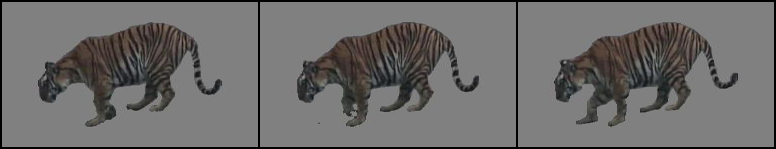}} \\
\noalign{\vskip 1mm} 
\hline
\noalign{\vskip 1mm} 

$(c)$   &
\raisebox{-.5\totalheight}{\includegraphics[width=0.6\textwidth]{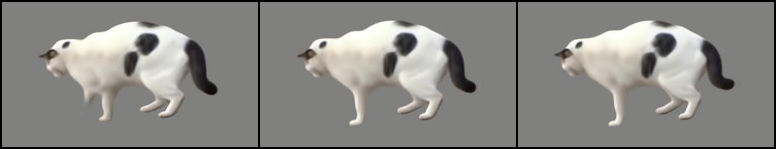}} \\
\noalign{\vskip 2mm} 

$(d)$   &
\raisebox{-.5\totalheight}{\includegraphics[width=0.6\textwidth]{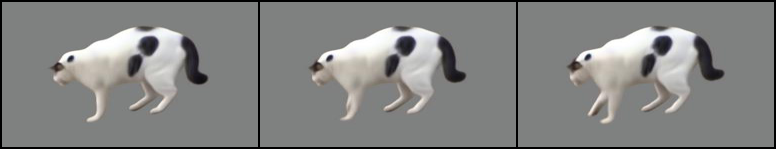}} \\
\noalign{\vskip 2mm} 
$(e)$   &
\raisebox{-.5\totalheight}{\includegraphics[width=0.6\textwidth]{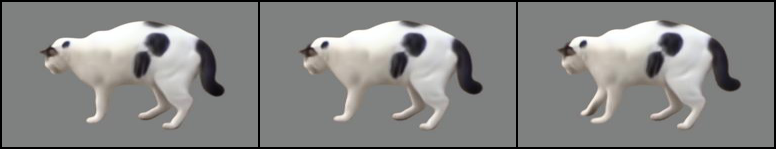}} \\
\end{tabular}

\caption{Ablation study for the affine invariant domain confusion component. $(a)$, $(b)$ depict the original cat and tiger. The $(c)$ row presents the artifacts caused after omitting the learned affine transformation. Bottom rows depict the results for avoiding the learned transformation at inference $(d)$ and applying the transformation at inference $(e)$, after using the learned affine transformation at training. }
\label{fig:abl_affine}
\end{figure*}

An ablation study is provided in the main text, we present the visual results in Fig.~\ref{fig:abl}, Fig.~\ref{fig:abl_temp}, and Fig~\ref{fig:abl_affine}.

\section{Implementation details}
\label{sec:impl}

\subsection{Keypoint extraction}

For the keypoint extractor $E$, similarly to previous work ~\cite{suwajanakorn2018discovery, siarohin2019animating, siarohin2019first}, we employ a U-Net~\cite{unet} architecture  which estimates K heatmaps from the input image and uses a spatial softmax layer to produce a distribution map over the image pixels, denoted $\{H^\ell\}^{K-1}_{\ell=0}$. We then compute the expected values of these distributions to recover a pixel coordinate for each keypoint:
\begin{align}
k^\ell = [u^\ell, v^\ell]^T = \sum_{u,v} [u \cdot H^\ell(u,v), v \cdot H^\ell(u,v)]^T
\end{align}
where we sum over all image pixels $\{u,v\}$. The keypoints are then projected to spatial confidence maps $h^\ell$ such that for any pixel coordinates in the image $p = (u,v)$, we have: 
\begin{align}
h^\ell(p) = \frac{1}{\alpha}\exp{ \frac{-\abs{p - k^\ell}}{\sigma^2} }
\end{align}
where $\alpha,\sigma$ are constant across all the experiments and $k^\ell$ are the extracted keypoints. We have used the constant values $\alpha=1$ and $\sigma=0.1$ for all experiments. 

\subsection{Network Architecture}

For $E$ we use the architecture based on U-Net ~\cite{unet}, as proposed by Siarohin et al.~\cite{siarohin2019first}. For $G_A$, $G_B$, $R_A$, $R_B$ we use the generator architecture proposed by Zhu et al.~\cite{CycleGAN2017}, which utilizes skip connections for better visual quality. 
The generator consists of $9$ residual blocks, each contains convolution, ReLU, and Instance Normalization layers. The discriminator consists of $3$ fully connected and Leaky ReLU layers, followed by a final sigmoid activation, similarly to Mokady et al.~\cite{mokady2019mask}. 

\subsection{Training details}
We use the Adam~\cite{adam} optimizer with a learning rate of $1e^{-4}$ for both generators and discriminators. Training time is approximately $12$ hours over a  single NVIDIA GTX1080 GPU. 

For the first step optimization:
\begin{align}
\mathcal{L}_{\text{G}} = \lambda_{\text{seg}}\mathcal{L_{\text{seg}}} + \lambda_{\text{DC}}\mathcal{L}^A_{\text{DC}} + \lambda_{\text{tmp}}\mathcal{L}_{\text{tmp}} + \lambda_{\text{eq}}\mathcal{L}_{\text{eq}} + \lambda_{\text{sep}} \mathcal{L}_{\text{sep}} + \lambda_{\text{sill}} \mathcal{L}_{\text{sill}}, 
\end{align} 
we use the following hyperparameters: $ \lambda_{\text{seg}} = 50$, $ \lambda_{\text{DC}} = 0.5$, $ \lambda_{\text{tmp}} = 1.0$, $ \lambda_{\text{eq}} = 1.0$, $ \lambda_{\text{sep}} = 1.0$, $ \lambda_{\text{sill}} = 0.5$, $\delta= 0.1$.

And for the second step:
\begin{align}
\mathcal{L}_{\text{G}} = \mathcal{L}_{L1} + \lambda_{\text{LPIPS}}\mathcal{L}_{\text{LPIPS}},
\end{align}

We use $\lambda_{\text{LPIPS}} = 2.0$.
For all experiments we use between $10$ to $14$ keypoints, and train each step for approximately $45,000$ iterations.

\subsection{Segmentation}

Even though our method requires a binary segmentation of the object (silhouette), various methods can be used to acquire it, allowing a wide variety of object types to be considered. For the Chihuahua in Fig.~\ref{fig:comp4} and Fig.~\ref{fig:add7}, we used an off-the-shelf pretrained saliency segmentation network of Qin et al.~\cite{qin2020u2}. For YouTube-VOS dataset~\cite{xu2018youtube}, we used manually annotated segmentations. GIFs and flowers have no background, and so the silhouette can be extracted using a simple threshold. For dancing videos, we used a pretrained network of Guler et al.~\cite{guler2018densepose} for human segmentation.

\section{Additional visual results}

\begin{figure*}[h]
\begin{tabular}{ll}
~ & ~~~~~~~~~~~~~ $t$ ~~~ ~~~~~~~~~~~~~~$t+5$  ~~~~~~~~~~~~~~ $t+10$  ~~~~~~~~~~~~~ $t+15$  ~~~~~~~~~~~~~ $t+20$ \\

Input   &
\raisebox{-.5\totalheight}{\includegraphics[width=0.8\textwidth]{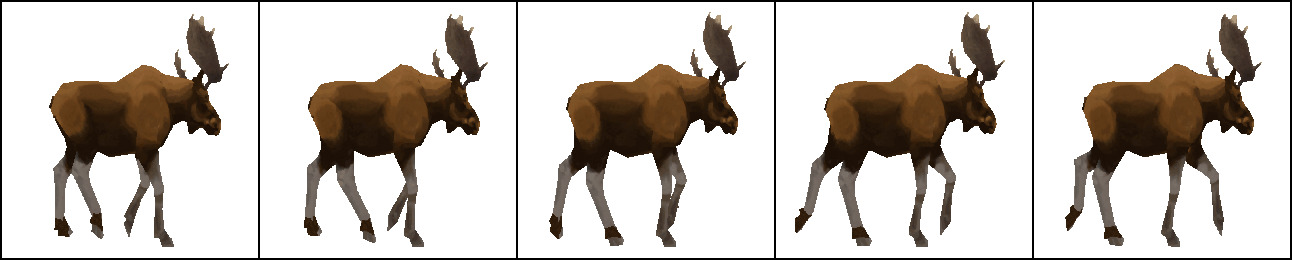}} \\
Ours   &
\raisebox{-.5\totalheight}{\includegraphics[width=0.8\textwidth]{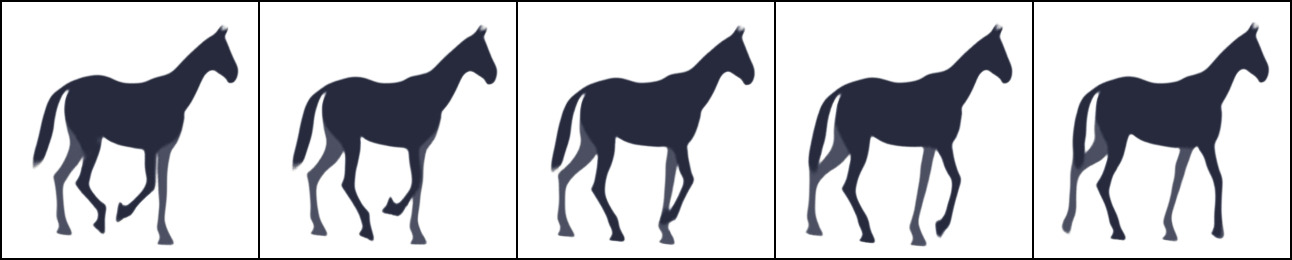}} \\
Segmentation & \raisebox{-.5\totalheight}{\includegraphics[width=0.8\textwidth]{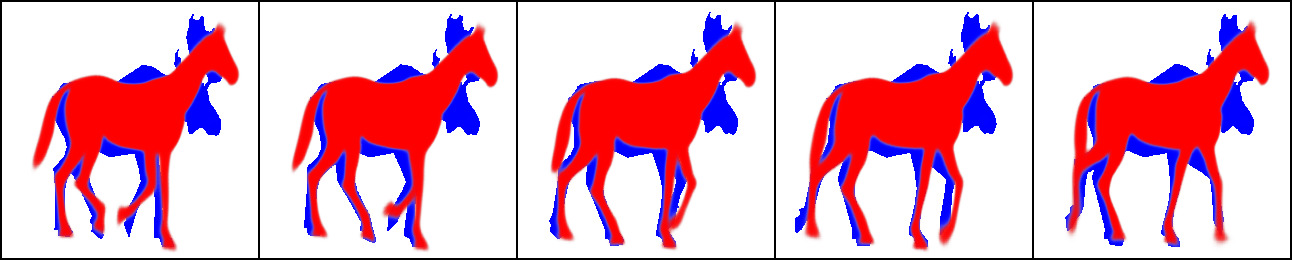}} \\
\noalign{\vskip 1mm} 
Input   &
\raisebox{-.5\totalheight}{\includegraphics[width=0.8\textwidth]{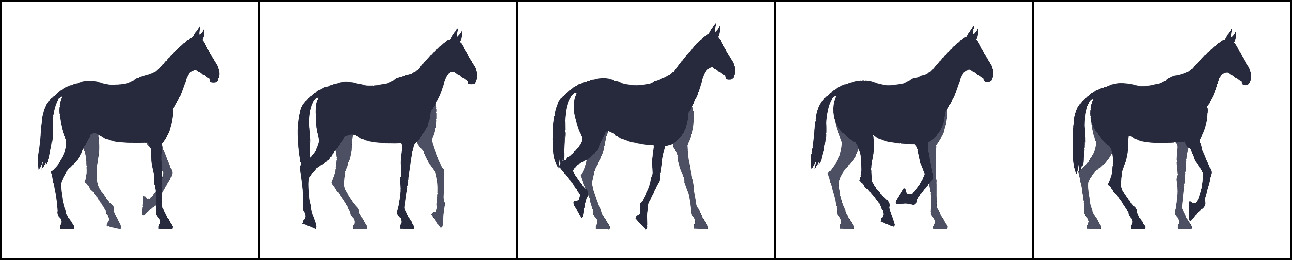}} \\
Ours   &
\raisebox{-.5\totalheight}{\includegraphics[width=0.8\textwidth]{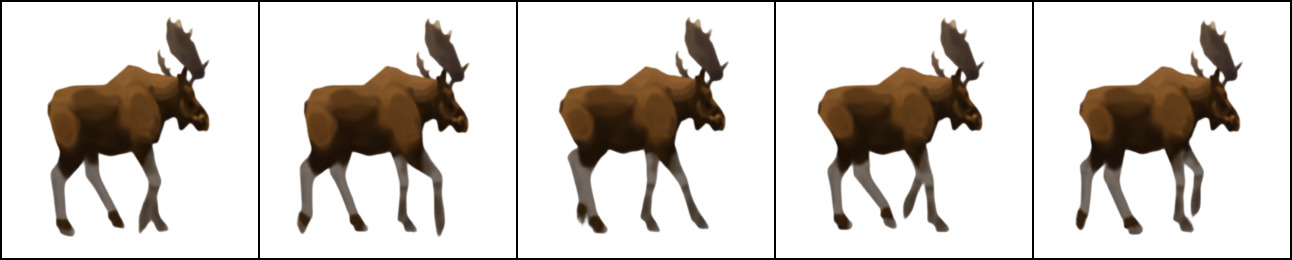}} \\
Segmentation & \raisebox{-.5\totalheight}{\includegraphics[width=0.8\textwidth]{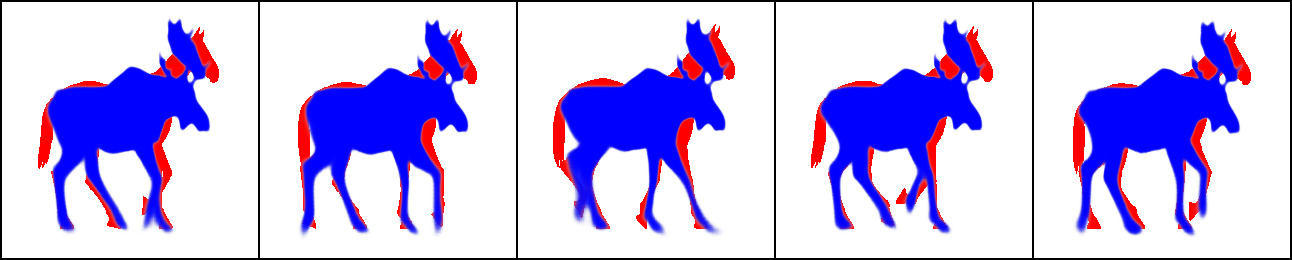}} \\
\end{tabular}
\caption{GIF Synchronization. To demonstrate alignment we show the segmentation of the elk (blue) and the horse (red) on top of each other.}
\label{fig:elk}
\end{figure*}

\begin{figure*}[h]

\begin{tabular}{c}

 ~~~~~~~~~~~~~ ~~~~~~~~~~~~~ ~~~~~~~~~~~ $t$  ~~~~ ~~~~~~~~~~~~~ $t+5$ ~ ~~~~~~~~~~~~~ $t+10$  ~~~~~~~~~~~~~ $t+15$  ~~~~~~~~~~~~~ $t+20$ ~~~~~~~~ \\

\begin{tabular}{l} \noalign{\vskip -6mm} Input \\ \noalign{\vskip 13mm}  Ours\\ \noalign{\vskip 13mm} FOMM~\cite{siarohin2019first}\\ \noalign{\vskip 13mm} Cycle~\cite{CycleGAN2017}\\  \noalign{\vskip 13mm}ReCycle~\cite{bansal2018recycle} \\ \end{tabular}
\raisebox{-.45\totalheight}{\includegraphics[width=0.8\textwidth]{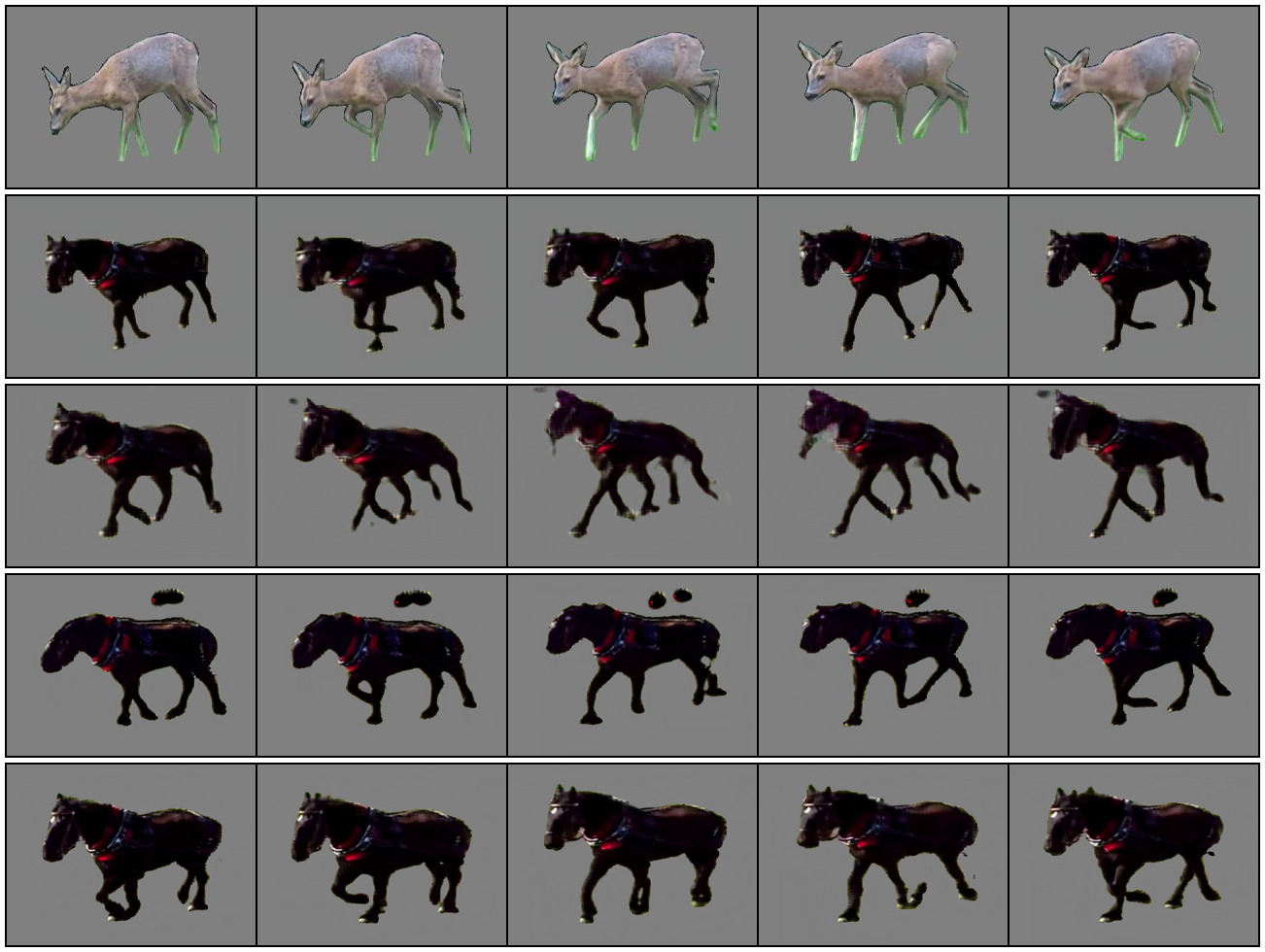}} \\
\noalign{\vskip 3mm} 
\begin{tabular}{l} \noalign{\vskip -6mm} Input \\ \noalign{\vskip 13mm}  Ours\\ \noalign{\vskip 13mm} FOMM~\cite{siarohin2019first}\\ \noalign{\vskip 13mm} Cycle~\cite{CycleGAN2017}\\  \noalign{\vskip 13mm}ReCycle~\cite{bansal2018recycle} \\ \end{tabular} 
\raisebox{-.45\totalheight}{\includegraphics[width=0.8\textwidth]{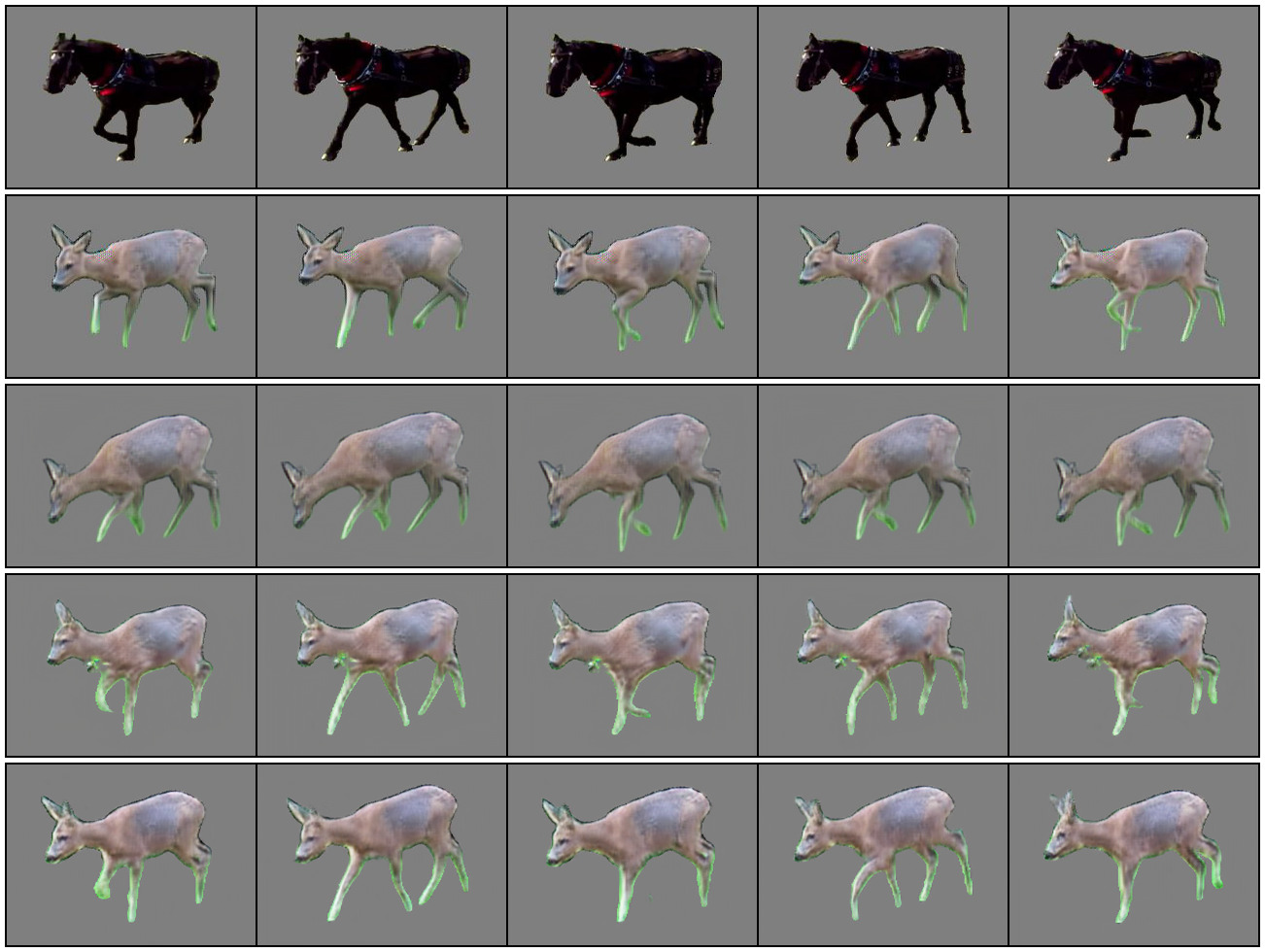}} \\

\end{tabular}

\caption{Comparison for horse/deer pair. As can be seen, our method successfully transfers motion while preserving the original style and appearance.}
\label{fig:comp1}
\end{figure*}

\begin{figure*}[h]

\centering
\begin{tabular}{c}

 ~~~~~~~~~~~~~ ~~~~~~~~~~~~~ ~~~~~~~~~~~ $t$  ~~~~ ~~~~~~~~~~~~~ $t+5$ ~ ~~~~~~~~~~~~~ $t+10$  ~~~~~~~~~~~~~ $t+15$  ~~~~~~~~~~~~~ $t+20$ ~~~~~~~~ \\

\begin{tabular}{l} \noalign{\vskip -6mm} Input \\ \noalign{\vskip 11mm}  Ours\\ \noalign{\vskip 11mm} FOMM~\cite{siarohin2019first}\\ \noalign{\vskip 11mm} Cycle~\cite{CycleGAN2017}\\  \noalign{\vskip 11mm}ReCycle~\cite{bansal2018recycle} \\ \end{tabular}   
\raisebox{-.45\totalheight}{\includegraphics[width=0.8\textwidth]{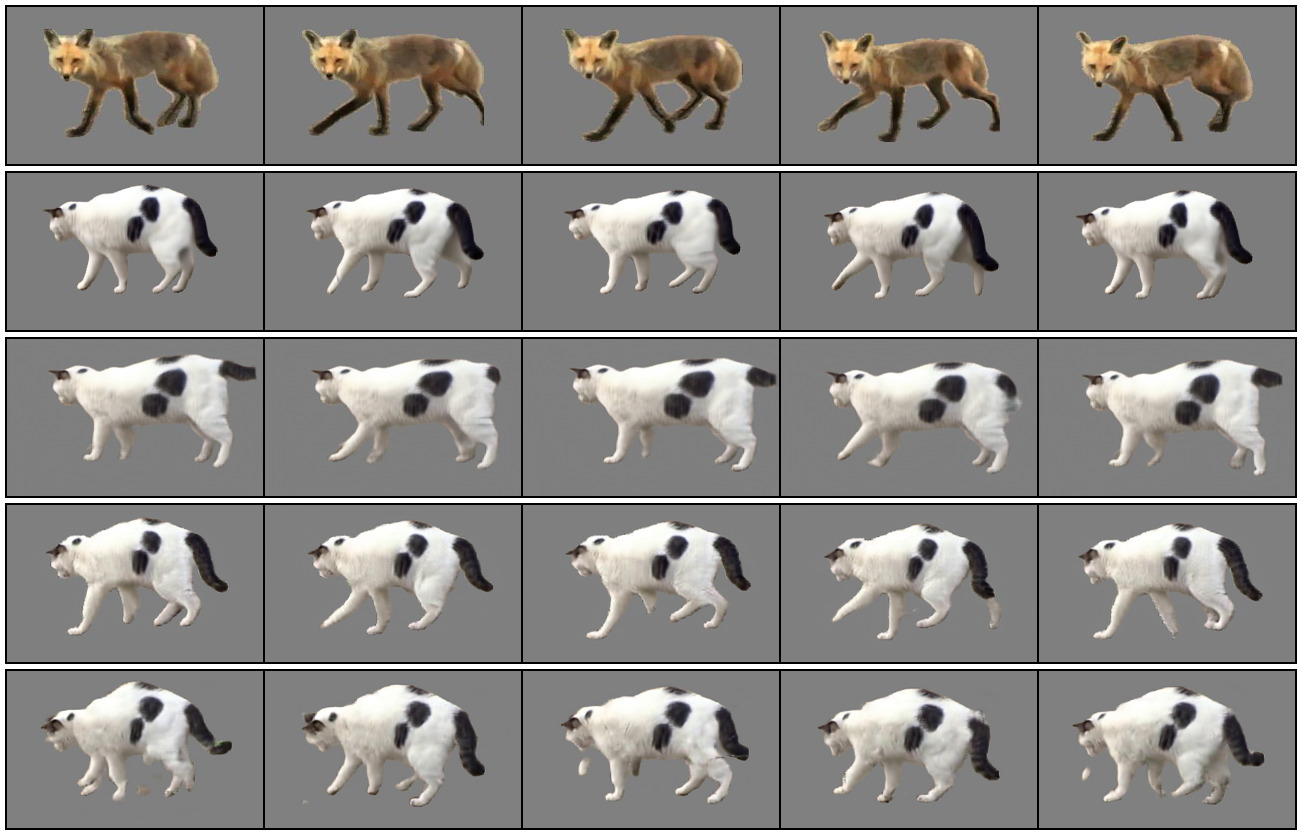}} \\
\noalign{\vskip 3mm} 
\begin{tabular}{l} \noalign{\vskip -6mm} Input \\ \noalign{\vskip 11mm}  Ours\\ \noalign{\vskip 11mm} FOMM~\cite{siarohin2019first}\\ \noalign{\vskip 11mm} Cycle~\cite{CycleGAN2017}\\  \noalign{\vskip 11mm}ReCycle~\cite{bansal2018recycle} \\ \end{tabular}   
\raisebox{-.45\totalheight}{\includegraphics[width=0.8\textwidth]{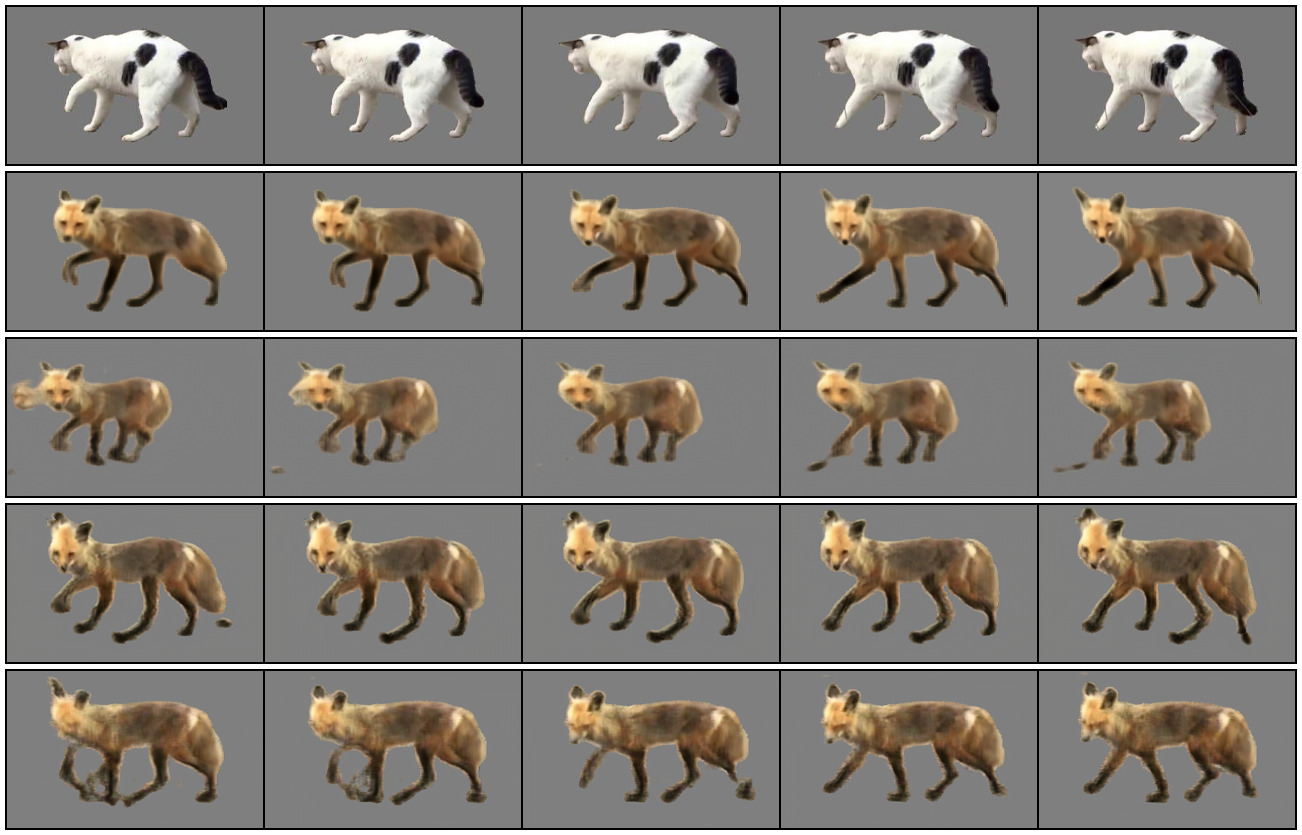}} \\

\end{tabular}

\caption{Comparison for cat/fox. As can be seen, our method successfully transfers motion while preserving the original style and appearance.}
\label{fig:comp2}
\end{figure*}

\begin{figure*}[h]

\centering
\begin{tabular}{c}

 ~~~~~~~~~~~~~ ~~~~~~~~~~~~~ ~~~~~~~~~~~ $t$  ~~~~ ~~~~~~~~~~~~~ $t+5$ ~ ~~~~~~~~~~~~~ $t+10$  ~~~~~~~~~~~~~ $t+15$  ~~~~~~~~~~~~~ $t+20$ ~~~~~~~~ \\

\begin{tabular}{l} \noalign{\vskip -6mm} Input \\ \noalign{\vskip 13mm}  Ours\\ \noalign{\vskip 13mm} FOMM~\cite{siarohin2019first}\\ \noalign{\vskip 13mm} Cycle~\cite{CycleGAN2017}\\  \noalign{\vskip 13mm}ReCycle~\cite{bansal2018recycle} \\ \end{tabular}  
\raisebox{-.45\totalheight}{\includegraphics[width=0.8\textwidth]{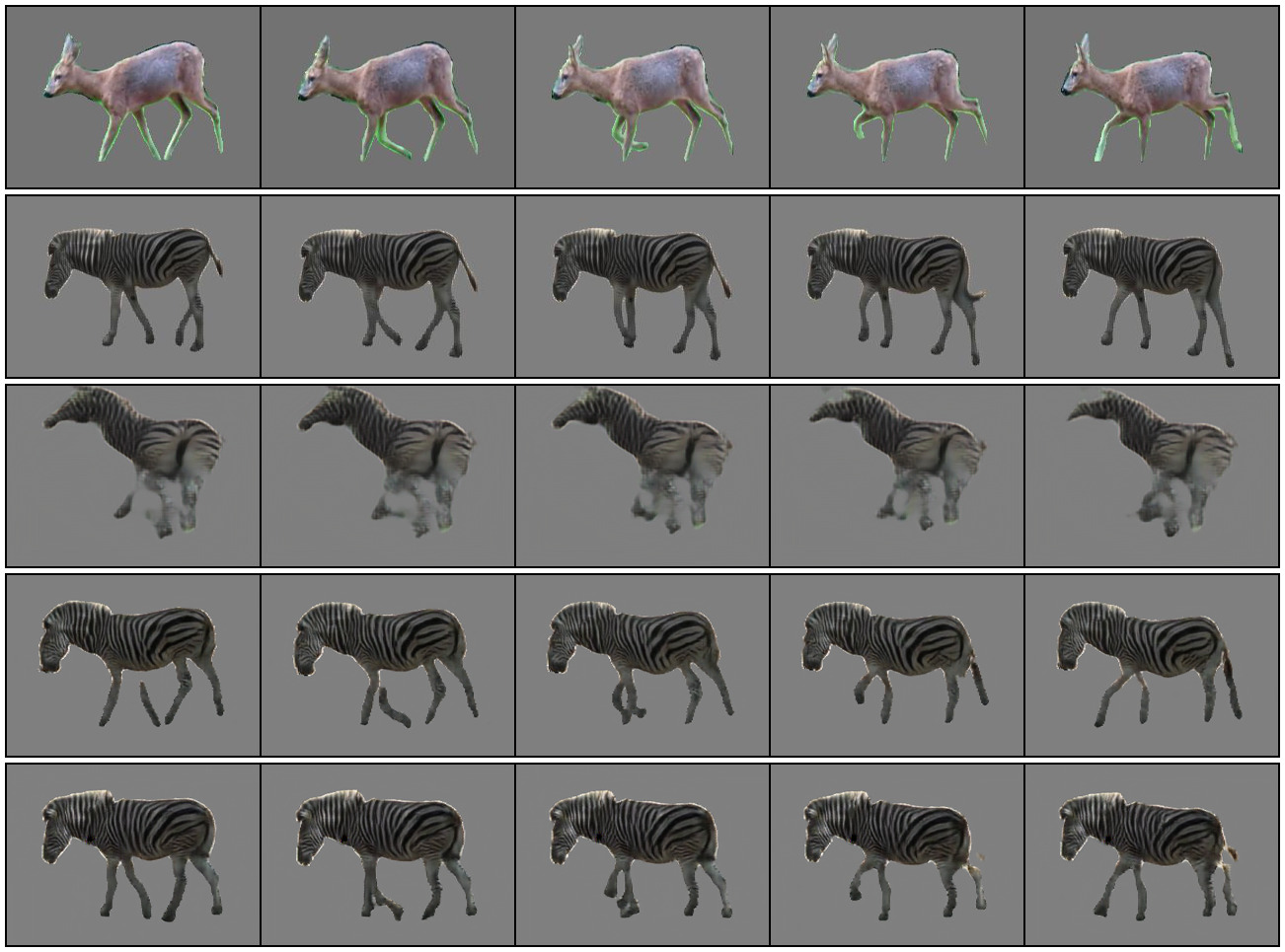}} \\
\noalign{\vskip 3mm} 
\begin{tabular}{l} \noalign{\vskip -6mm} Input \\ \noalign{\vskip 13mm}  Ours\\ \noalign{\vskip 13mm} FOMM~\cite{siarohin2019first}\\ \noalign{\vskip 13mm} Cycle~\cite{CycleGAN2017}\\  \noalign{\vskip 13mm}ReCycle~\cite{bansal2018recycle} \\ \end{tabular}  
\raisebox{-.45\totalheight}{\includegraphics[width=0.8\textwidth]{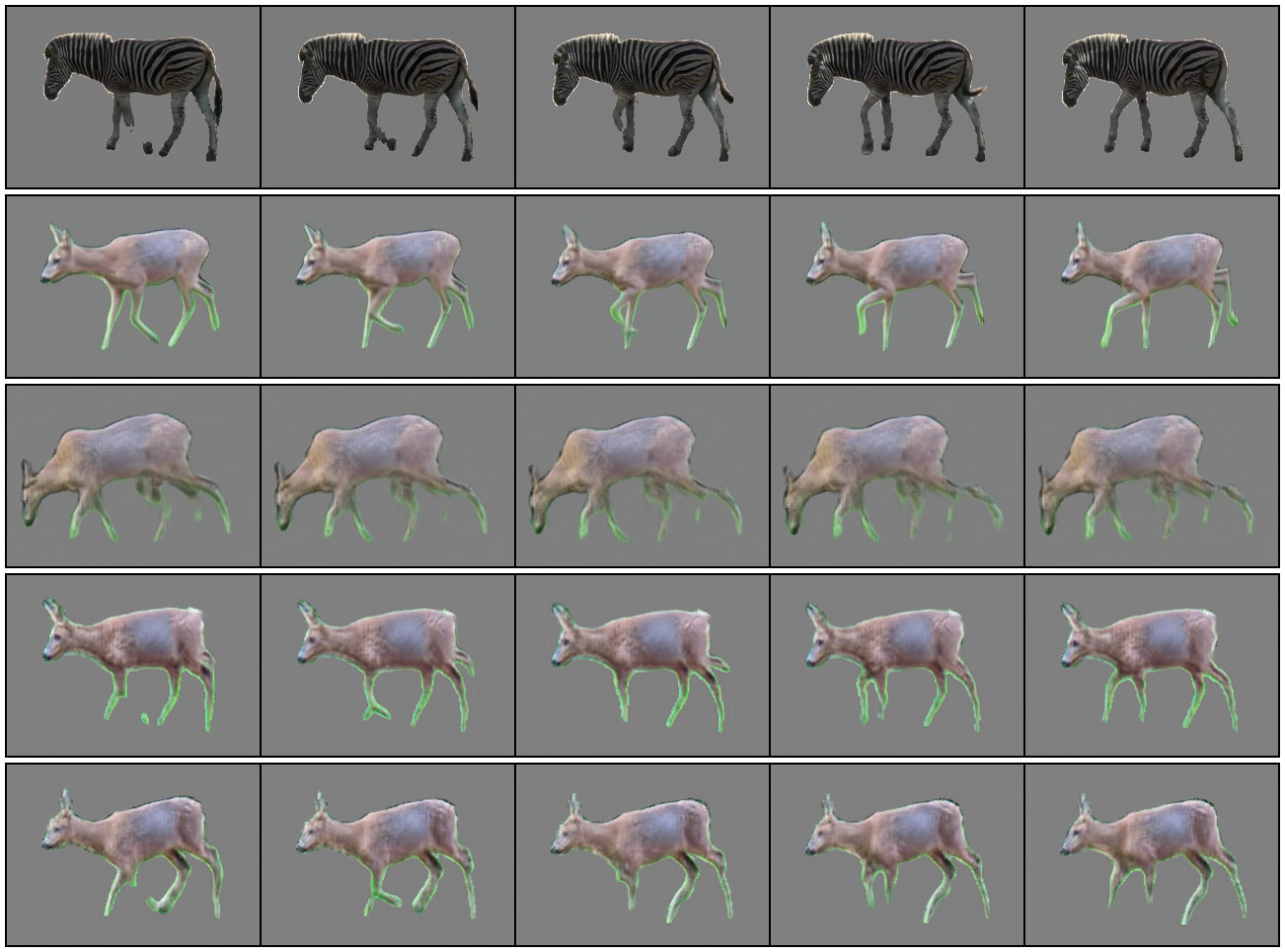}} \\

\end{tabular}

\caption{Comparison for zebra/deer. As can be seen, our method successfully transfers motion while preserving the original style and appearance.}
\label{fig:comp3}
\end{figure*}

\begin{figure*}[h]

\centering
\begin{tabular}{c}

 ~~~~~~~~~~~~~ ~~~~~~~~~~~~~ ~~~~~~~~~~~ $t$  ~~~~ ~~~~~~~~~~~~~ $t+5$ ~ ~~~~~~~~~~~~~ $t+10$  ~~~~~~~~~~~~~ $t+15$  ~~~~~~~~~~~~~ $t+20$ ~~~~~~~~ \\

\begin{tabular}{l} \noalign{\vskip -6mm} Input \\ \noalign{\vskip 10mm}  Ours\\ \noalign{\vskip 10mm} FOMM~\cite{siarohin2019first}\\ \noalign{\vskip 10mm} Cycle~\cite{CycleGAN2017}\\  \noalign{\vskip 10mm}ReCycle~\cite{bansal2018recycle} \\ \end{tabular}  
\raisebox{-.45\totalheight}{\includegraphics[width=0.8\textwidth]{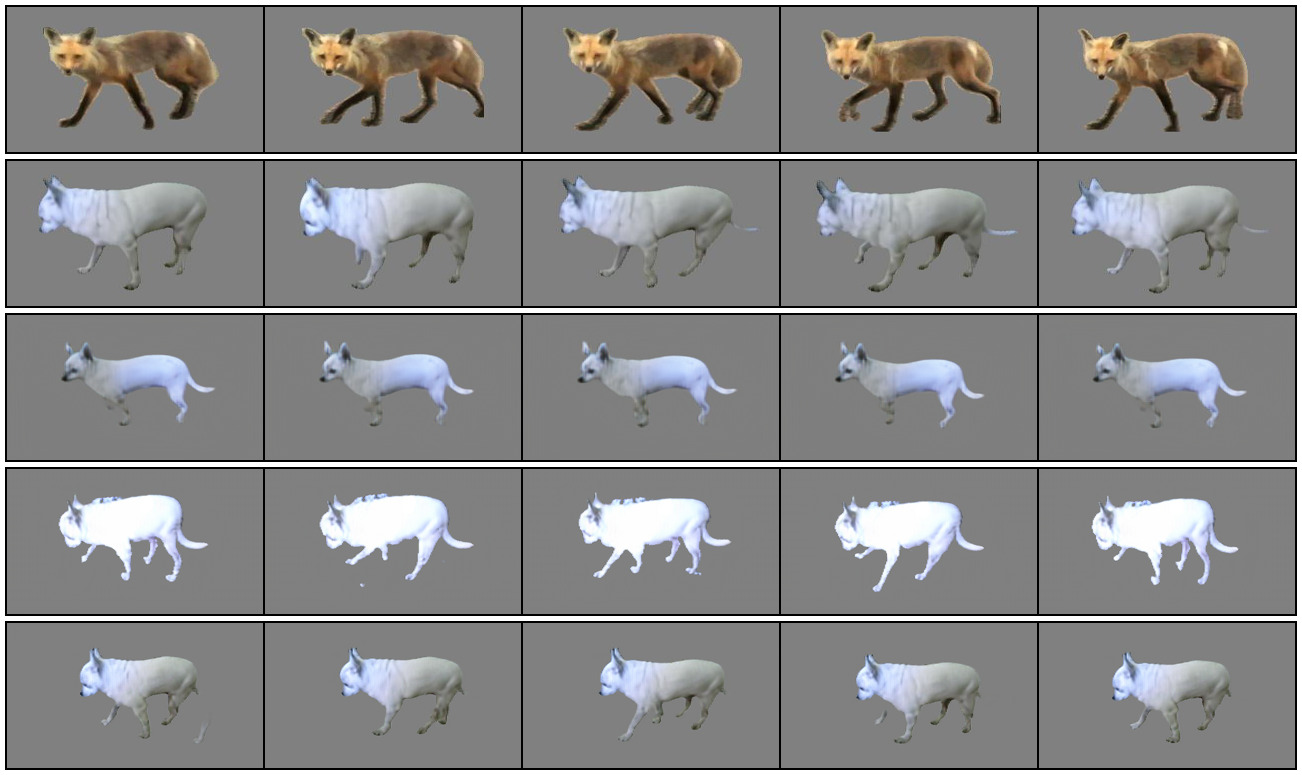}} \\
\noalign{\vskip 3mm} 
\begin{tabular}{l} \noalign{\vskip -6mm} Input \\ \noalign{\vskip 10mm}  Ours\\ \noalign{\vskip 10mm} FOMM~\cite{siarohin2019first}\\ \noalign{\vskip 10mm} Cycle~\cite{CycleGAN2017}\\  \noalign{\vskip 10mm}ReCycle~\cite{bansal2018recycle} \\ \end{tabular}  
\raisebox{-.45\totalheight}{\includegraphics[width=0.8\textwidth]{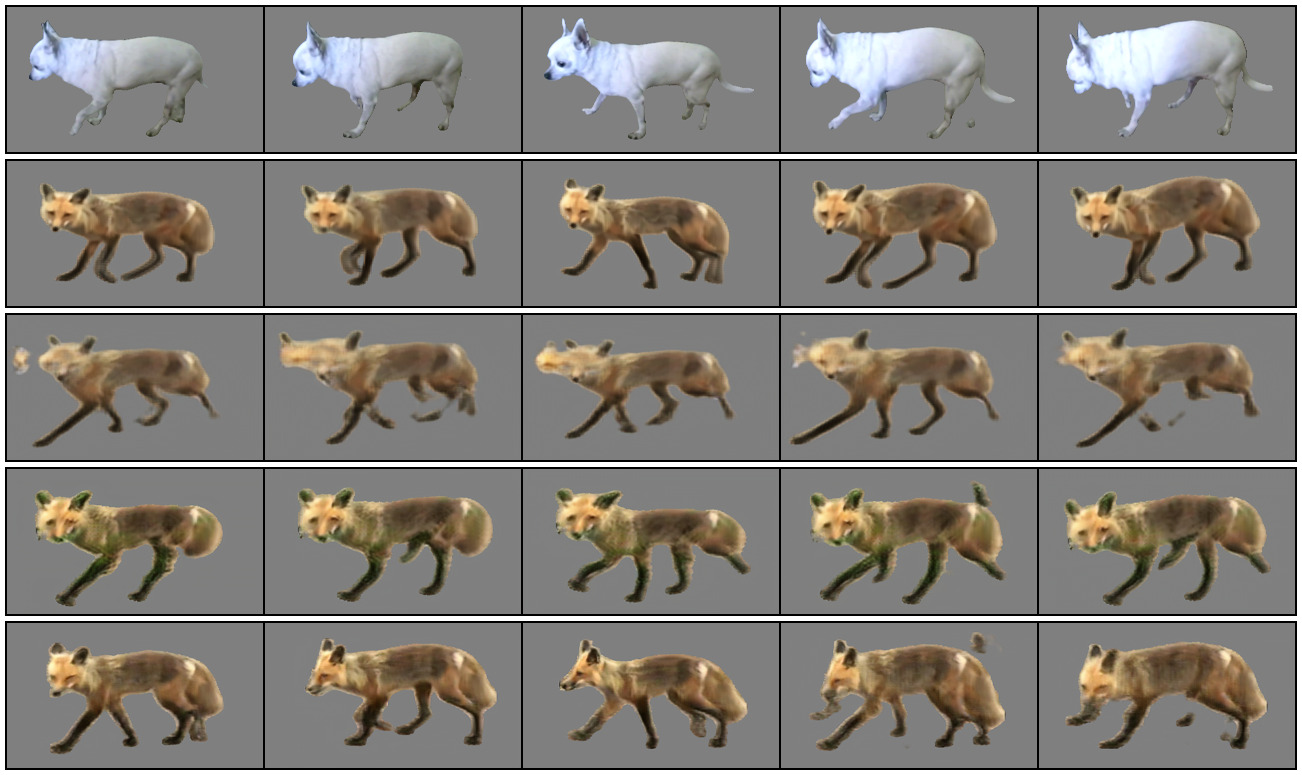}} \\

\end{tabular}

\caption{Comparison for chihuahua/fox. As can be seen, our method successfully transfers motion while preserving the original style and appearance.}
\label{fig:comp4}
\end{figure*}

\begin{figure*}[h]

\centering
\begin{tabular}{c}

 ~~~~~~~~~~~~~ ~~~~~~~~~~~~~ ~~~~~~~~~~~ $t$  ~~~~ ~~~~~~~~~~~~~ $t+5$ ~ ~~~~~~~~~~~~~ $t+10$  ~~~~~~~~~~~~~ $t+15$  ~~~~~~~~~~~~~ $t+20$ ~~~~~~~~ \\

\begin{tabular}{l} \noalign{\vskip -6mm} Input \\ \noalign{\vskip 10mm}  Ours\\ \noalign{\vskip 10mm} FOMM~\cite{siarohin2019first}\\ \noalign{\vskip 10mm} Cycle~\cite{CycleGAN2017}\\  \noalign{\vskip 10mm}ReCycle~\cite{bansal2018recycle} \\ \end{tabular}  
\raisebox{-.45\totalheight}{\includegraphics[width=0.8\textwidth]{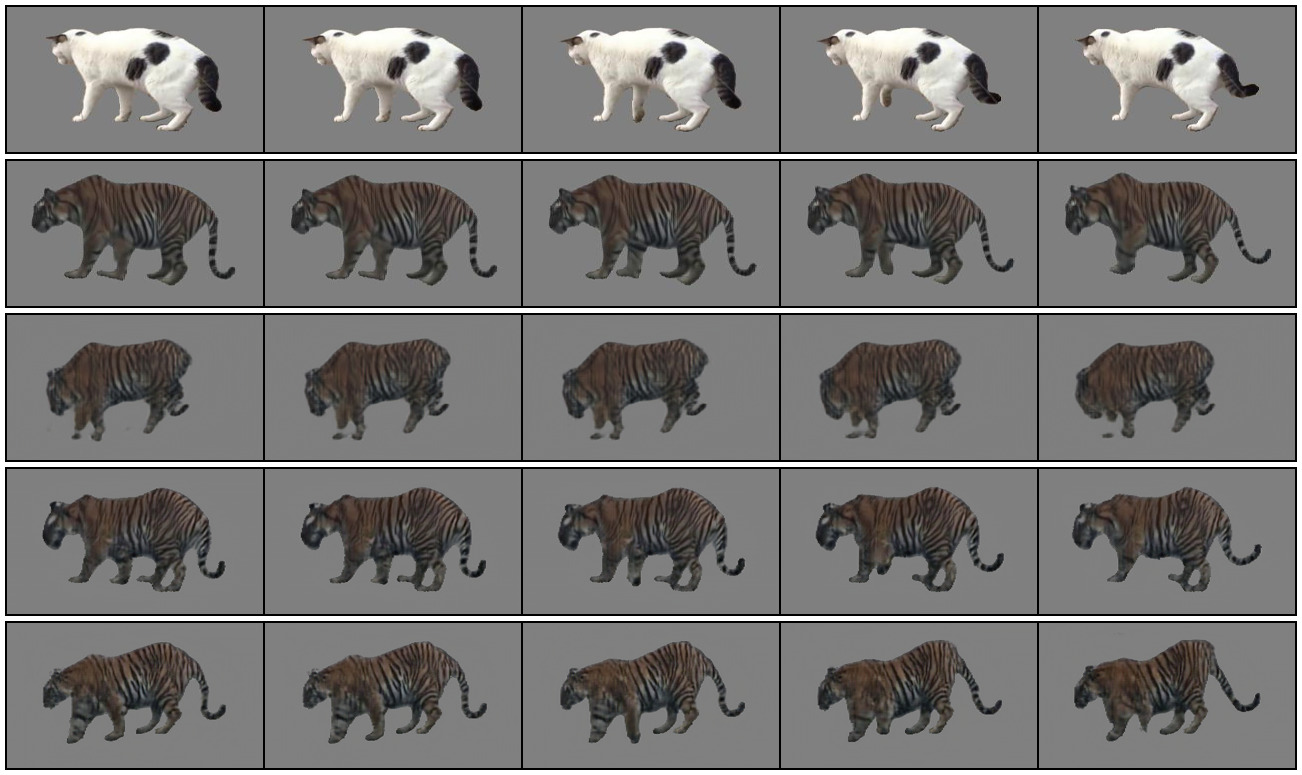}} \\
\noalign{\vskip 3mm} 
\begin{tabular}{l} \noalign{\vskip -6mm} Input \\ \noalign{\vskip 10mm}  Ours\\ \noalign{\vskip 10mm} FOMM~\cite{siarohin2019first}\\ \noalign{\vskip 10mm} Cycle~\cite{CycleGAN2017}\\  \noalign{\vskip 10mm}ReCycle~\cite{bansal2018recycle} \\ \end{tabular}  
\raisebox{-.45\totalheight}{\includegraphics[width=0.8\textwidth]{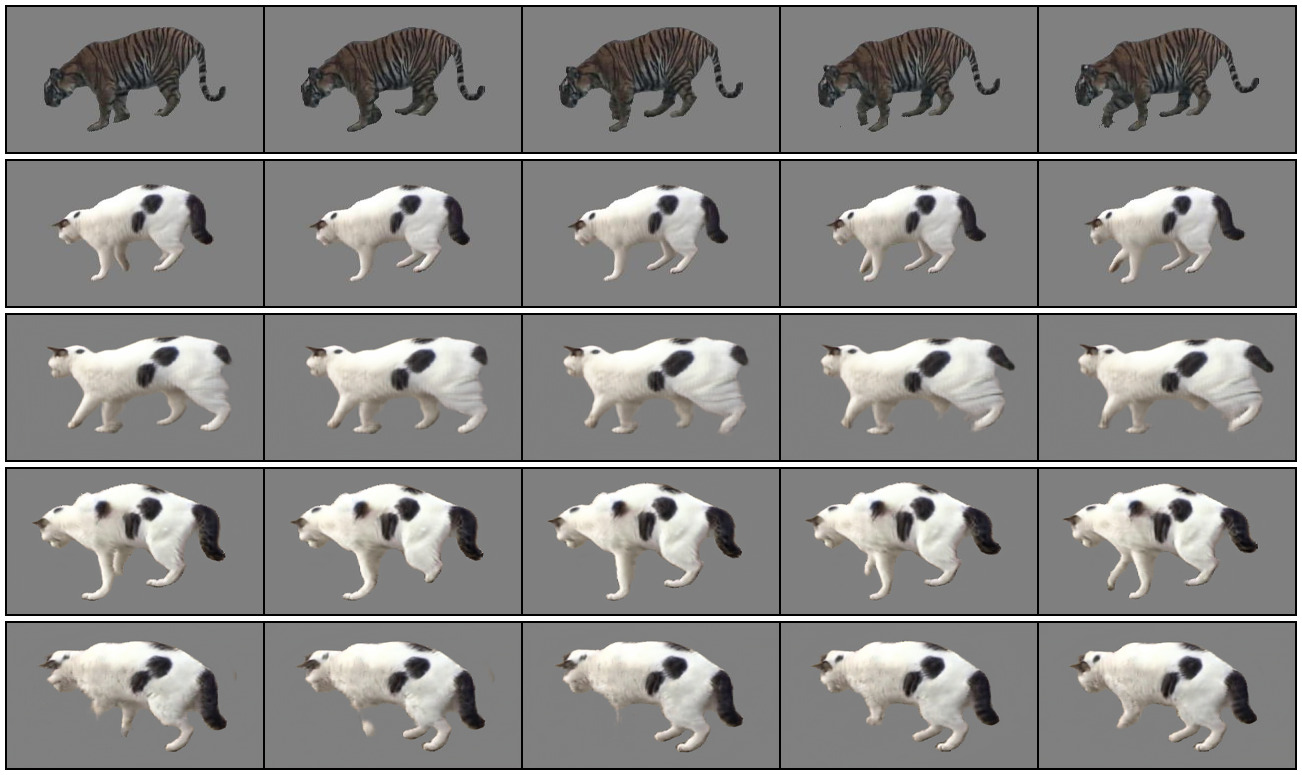}} \\

\end{tabular}

\caption{Comparison for cat/tiger. As can be seen, our method successfully transfers motion while preserving the original style and appearance.}
\label{fig:comp5}
\end{figure*}

\begin{figure*}[h]

\centering
\begin{tabular}{c}

 ~~~~~~~~~~~~~ ~~~~~~~~~~~~~ ~~~~~~~~~~~ $t$  ~~~~ ~~~~~~~~~~~~~ $t+5$ ~ ~~~~~~~~~~~~~ $t+10$  ~~~~~~~~~~~~~ $t+15$  ~~~~~~~~~~~~~ $t+20$ ~~~~~~~~ \\

\begin{tabular}{l} \noalign{\vskip -6mm} Input \\ \noalign{\vskip 10mm}  Ours\\ \noalign{\vskip 10mm} FOMM~\cite{siarohin2019first}\\ \noalign{\vskip 10mm} Cycle~\cite{CycleGAN2017}\\  \noalign{\vskip 10mm}ReCycle~\cite{bansal2018recycle} \\ \end{tabular}  
\raisebox{-.45\totalheight}{\includegraphics[width=0.8\textwidth]{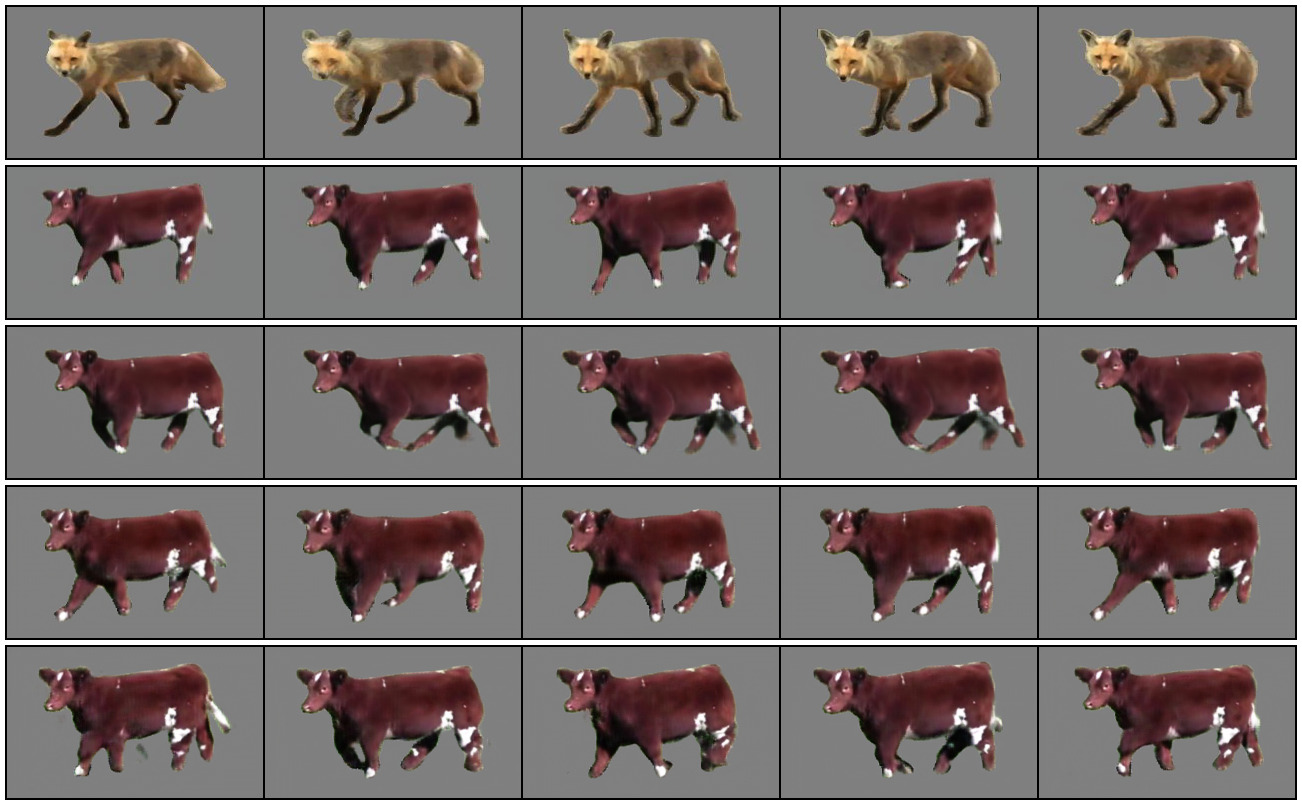}} \\
\noalign{\vskip 3mm} 
\begin{tabular}{l} \noalign{\vskip -6mm} Input \\ \noalign{\vskip 10mm}  Ours\\ \noalign{\vskip 10mm} FOMM~\cite{siarohin2019first}\\ \noalign{\vskip 10mm} Cycle~\cite{CycleGAN2017}\\  \noalign{\vskip 10mm}ReCycle~\cite{bansal2018recycle} \\ \end{tabular}  
\raisebox{-.45\totalheight}{\includegraphics[width=0.8\textwidth]{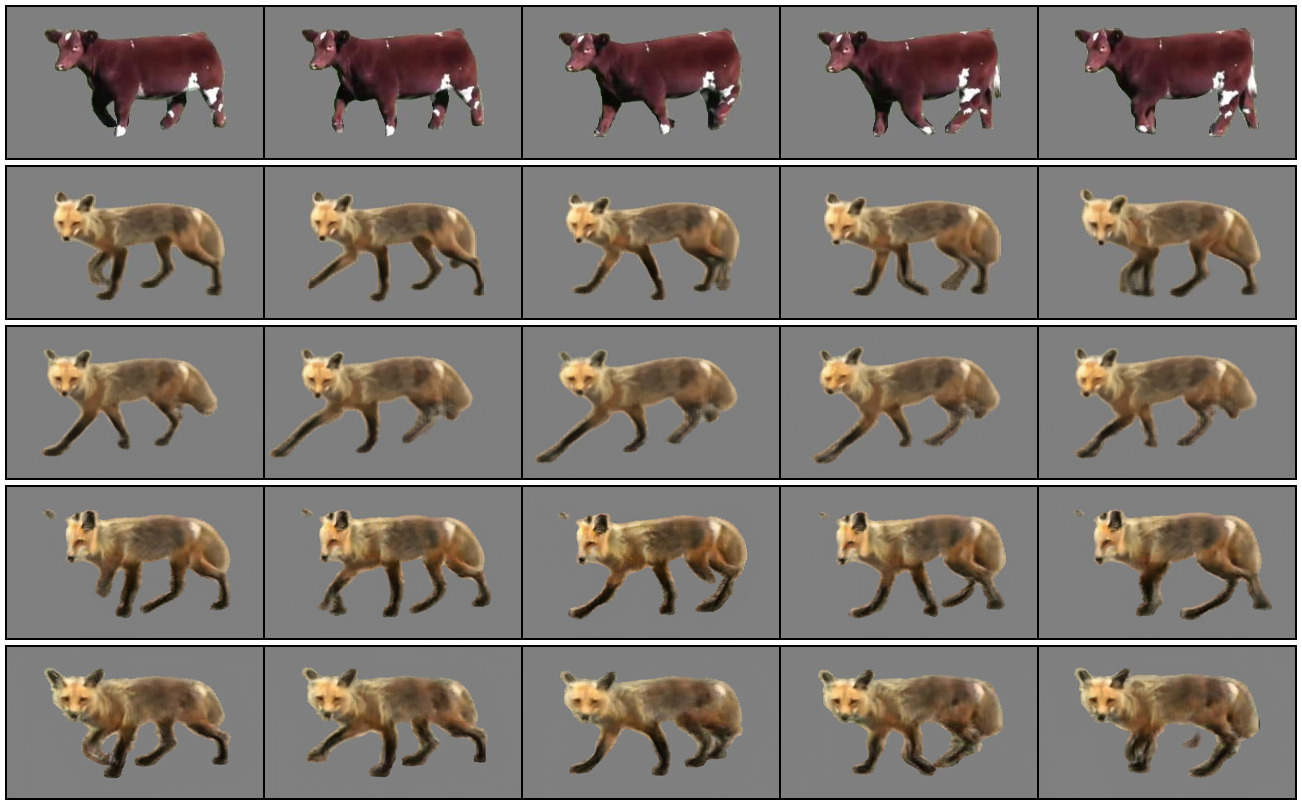}} \\

\end{tabular}

\caption{Comparison for cow/fox. As can be seen, our method successfully transfers motion while preserving the original style and appearance.}
\label{fig:comp6}
\end{figure*}

\begin{figure*}[h]

\centering
\begin{tabular}{ll}

~~~~~~ & ~~~~~~~~~~~ $t$ ~~~~~~~~~~~~~~~~~~ $t+1$ ~ ~~~~~~~~~~~~~~ $t+2$ ~ ~~~~~~~~~~~~ $t+3$ ~ ~~~~~~~~~~~~~ $t+4$  \\

\begin{tabular}{l} Input \\ ~\\ ~\\ ~\\ Ours \\ \end{tabular}   &
\raisebox{-.5\totalheight}{\includegraphics[width=0.8\textwidth]{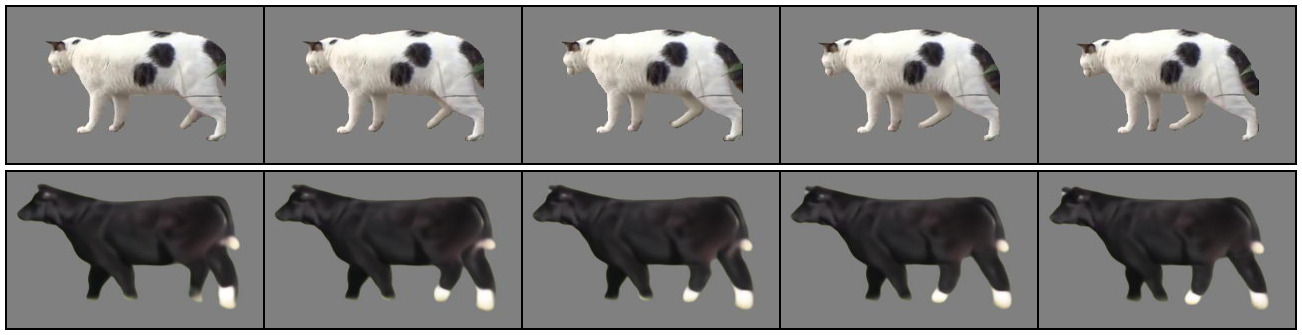}} \\
\noalign{\vskip 2mm} 
\begin{tabular}{l} Input \\ ~\\ ~\\ ~\\ Ours \\ \end{tabular}   &
\raisebox{-.5\totalheight}{\includegraphics[width=0.8\textwidth]{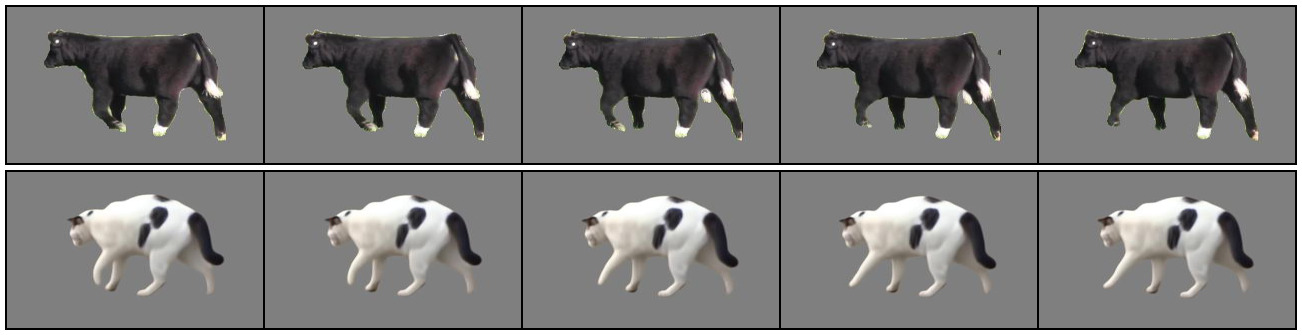}} \\
\noalign{\vskip 2mm} 
~ & ~~~~~~~ $t+5$ ~~~~~~~~~~~~~~~~~~ $t+6$ ~ ~~~~~~~~~~~~~~ $t+7$ ~ ~~~~~~~~~~~~ $t+8$ ~ ~~~~~~~~~~~~~ $t+9$  \\
\begin{tabular}{l} Input \\ ~\\ ~\\ ~\\ Ours \\ \end{tabular}   &
\raisebox{-.5\totalheight}{\includegraphics[width=0.8\textwidth]{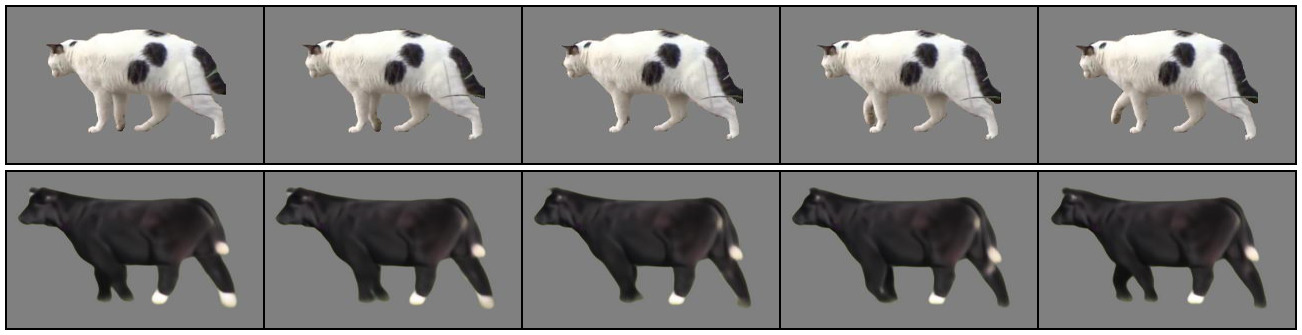}} \\
\noalign{\vskip 2mm} 
\begin{tabular}{l} Input \\ ~\\ ~\\ ~\\ Ours \\ \end{tabular}   &
\raisebox{-.5\totalheight}{\includegraphics[width=0.8\textwidth]{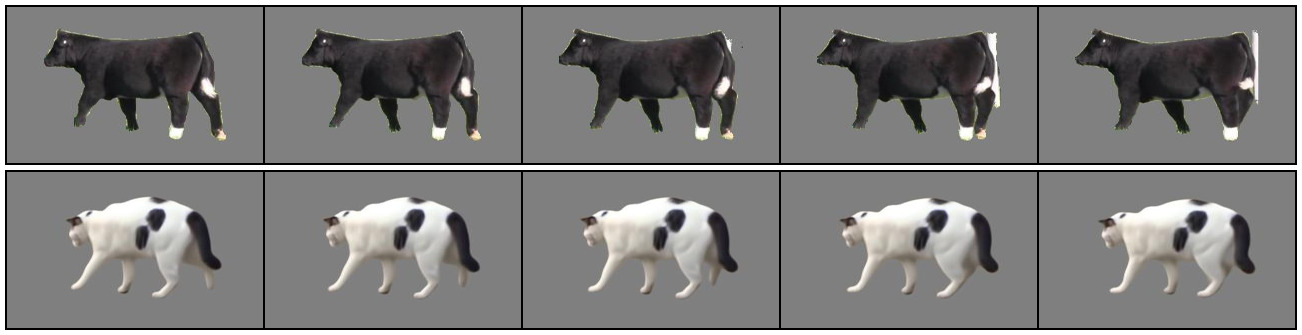}} \\
\end{tabular}

\caption{Additional results for our method.}
\label{fig:add1}
\end{figure*}

\begin{figure*}[h]

\centering
\begin{tabular}{ll}
~~~~~~ & ~~~~~~~~~~~ $t$ ~~~~~~~~~~~~~~~~~~ $t+5$ ~ ~~~~~~~~~~~~~~ $t+10$ ~ ~~~~~~~~ $t+15$ ~ ~~~~~~~~~~~~~ $t+20$  \\
\begin{tabular}{l} Input \\ ~\\ ~\\ ~\\ Ours \\ \end{tabular}   &
\raisebox{-.5\totalheight}{\includegraphics[width=0.8\textwidth]{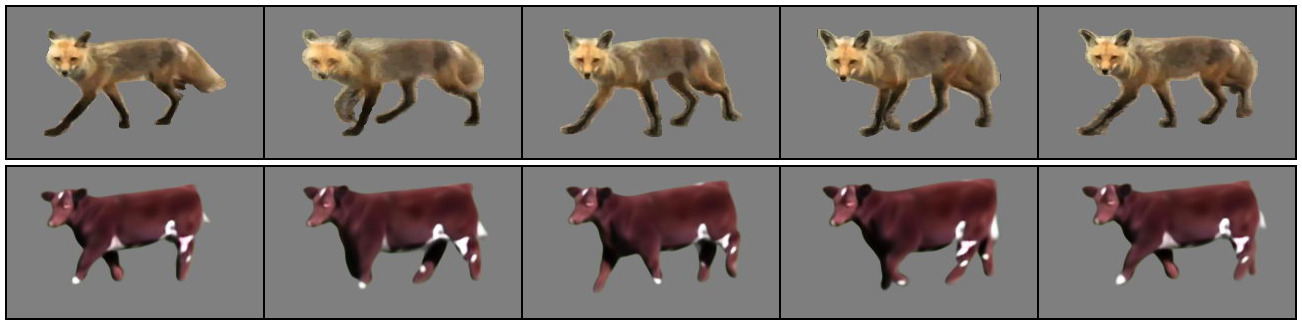}} \\
\noalign{\vskip 2mm} 
\begin{tabular}{l} Input \\ ~\\ ~\\ ~\\ Ours \\ \end{tabular}   &
\raisebox{-.5\totalheight}{\includegraphics[width=0.8\textwidth]{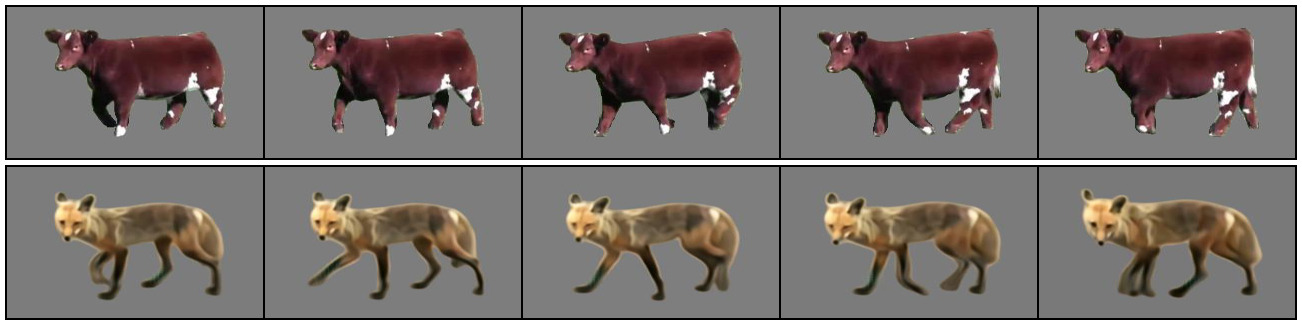}} \\
\noalign{\vskip 2mm} 
~ & ~~~~~~~ $t+25$ ~~~~~~~~~~~~~ $t+30$ ~ ~~~~~~~~~~~ $t+35$ ~ ~~~~~~~~~~~~ $t+40$ ~ ~~~~~~~~~~ $t+45$  \\
\begin{tabular}{l} Input \\ ~\\ ~\\ ~\\ Ours \\ \end{tabular}   &
\raisebox{-.5\totalheight}{\includegraphics[width=0.8\textwidth]{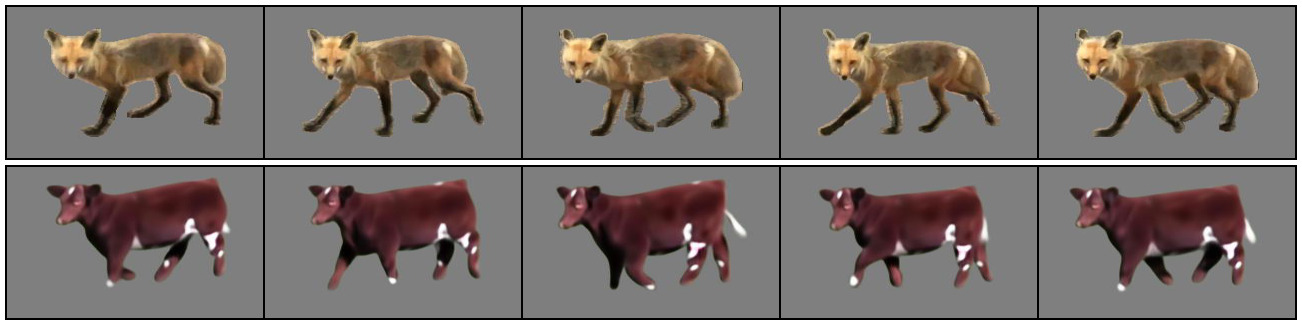}} \\
\noalign{\vskip 2mm} 
\begin{tabular}{l} Input \\ ~\\ ~\\ ~\\ Ours \\ \end{tabular}   &
\raisebox{-.5\totalheight}{\includegraphics[width=0.8\textwidth]{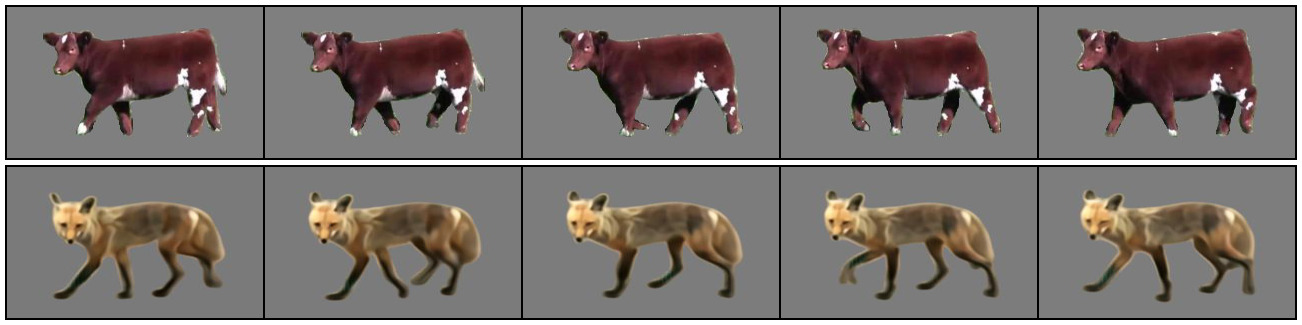}} \\
\end{tabular}

\caption{Additional results for our method.}
\label{fig:add2}
\end{figure*}

\begin{figure*}[h]

\centering
\begin{tabular}{ll}
~~~~~~ & ~~~~~~~~~~~ $t$ ~~~~~~~~~~~~~~~~~~ $t+2$ ~ ~~~~~~~~~~~~~~ $t+4$ ~ ~~~~~~~~~~~~ $t+6$ ~ ~~~~~~~~~~~~~ $t+8$  \\
\begin{tabular}{l} Input \\ ~\\ ~\\ ~\\ Ours \\ \end{tabular}   &
\raisebox{-.5\totalheight}{\includegraphics[width=0.8\textwidth]{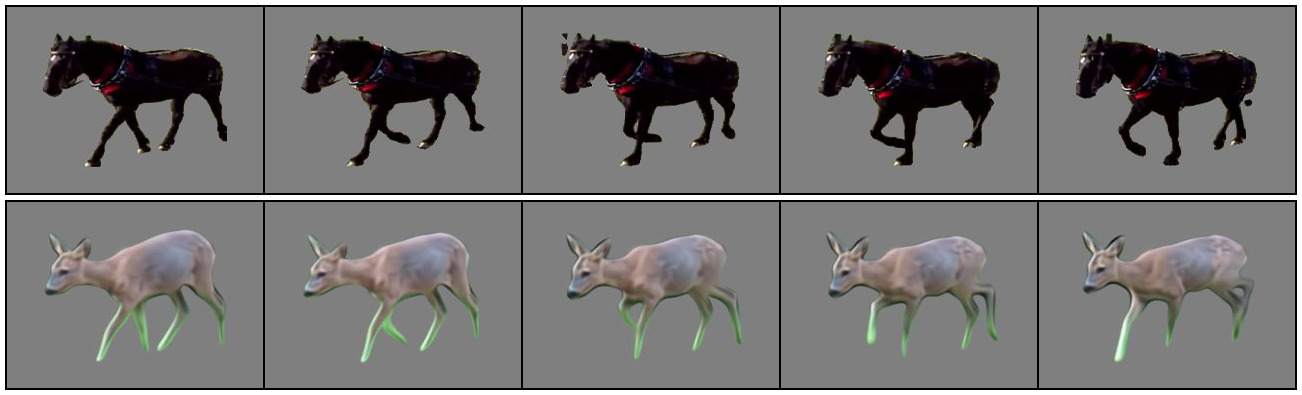}} \\
\noalign{\vskip 2mm} 
\begin{tabular}{l} Input \\ ~\\ ~\\ ~\\ Ours \\ \end{tabular}   &
\raisebox{-.5\totalheight}{\includegraphics[width=0.8\textwidth]{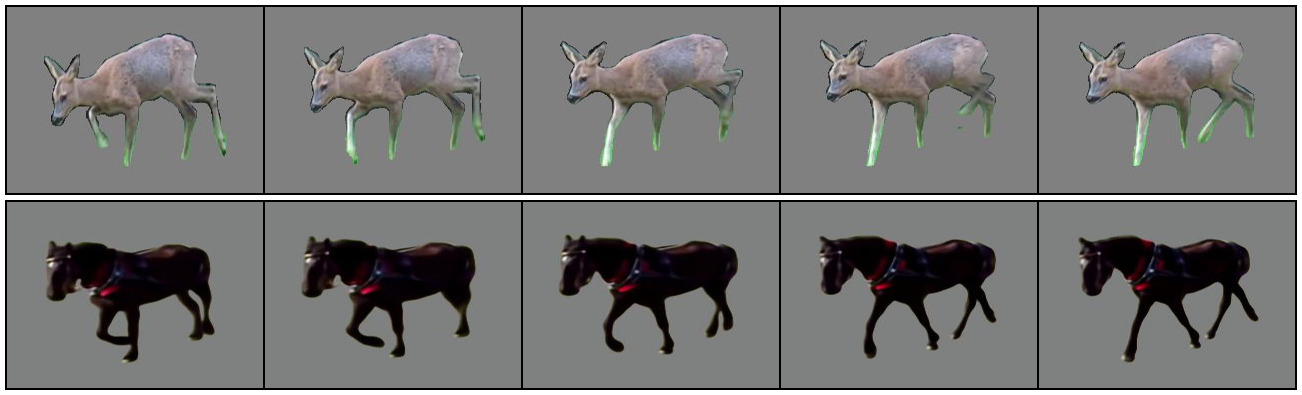}} \\
\noalign{\vskip 2mm} 
~ & ~~~~~~~ $t+10$ ~~~~~~~~~~~~~ $t+12$ ~ ~~~~~~~~~~~ $t+14$ ~ ~~~~~~~~~~~~ $t+16$ ~ ~~~~~~~~~~ $t+18$  \\
\begin{tabular}{l} Input \\ ~\\ ~\\ ~\\ Ours \\ \end{tabular}   &
\raisebox{-.5\totalheight}{\includegraphics[width=0.8\textwidth]{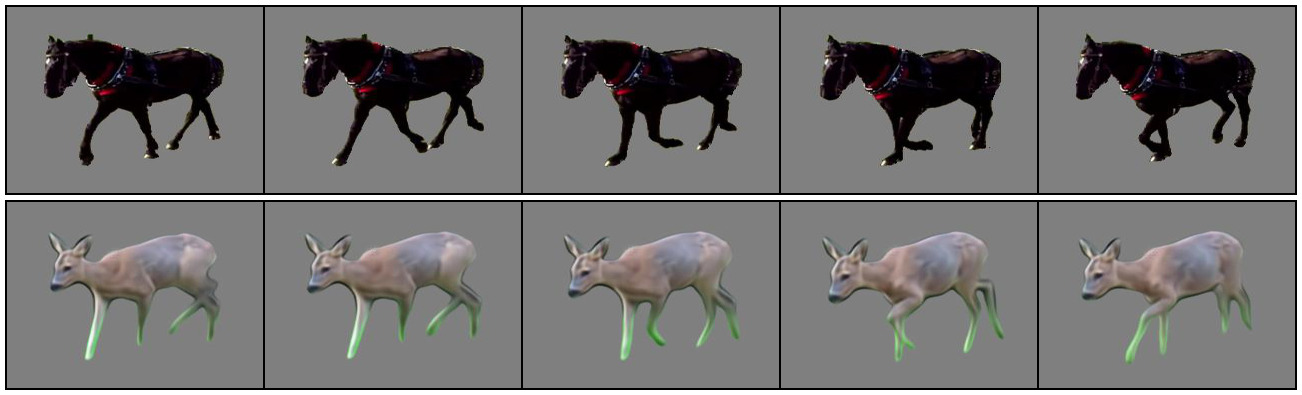}} \\
\noalign{\vskip 2mm} 
\begin{tabular}{l} Input \\ ~\\ ~\\ ~\\ Ours \\ \end{tabular}   &
\raisebox{-.5\totalheight}{\includegraphics[width=0.8\textwidth]{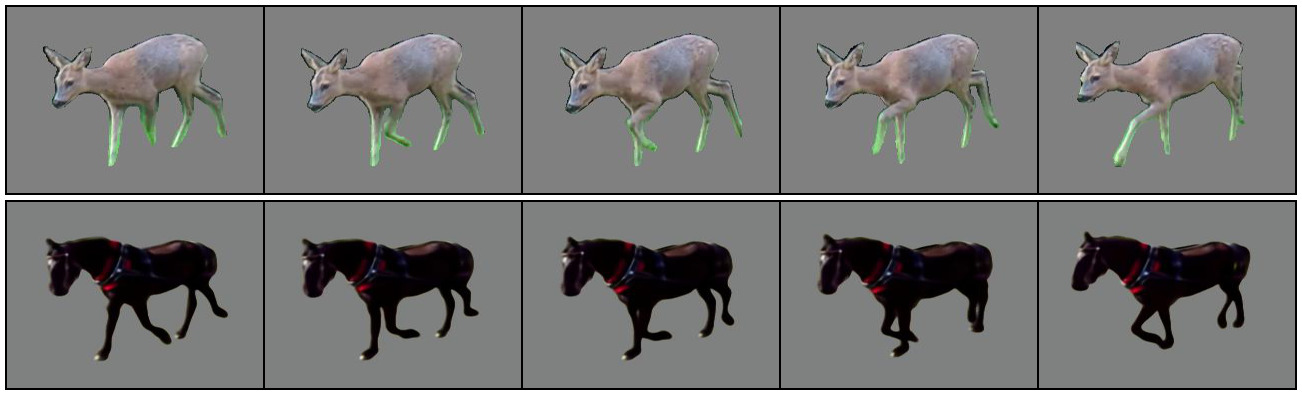}} \\
\end{tabular}

\caption{Additional results for our method.}
\label{fig:add3}
\end{figure*}

\begin{figure*}[h]

\centering
\begin{tabular}{ll}

~~~~~~ & ~~~~~~~~~~~ $t$ ~~~~~~~~~~~~~~~~~~ $t+5$ ~ ~~~~~~~~~~~~~~ $t+10$ ~ ~~~~~~~~ $t+15$ ~ ~~~~~~~~~~~~~ $t+20$  \\

\begin{tabular}{l} Input \\ ~\\ ~\\ ~\\ Ours \\ \end{tabular}   &
\raisebox{-.5\totalheight}{\includegraphics[width=0.8\textwidth]{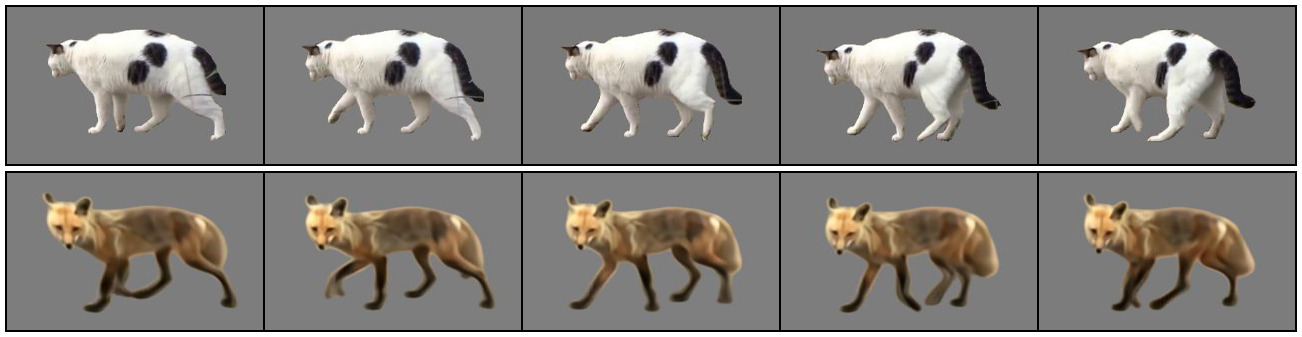}} \\
\noalign{\vskip 2mm} 
\begin{tabular}{l} Input \\ ~\\ ~\\ ~\\ Ours \\ \end{tabular}   &
\raisebox{-.5\totalheight}{\includegraphics[width=0.8\textwidth]{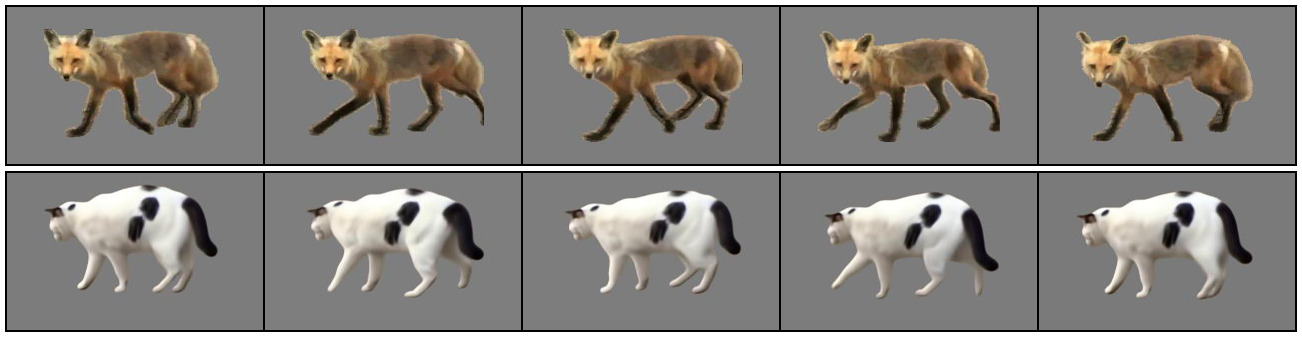}} \\
\noalign{\vskip 2mm} 
~ & ~~~~~~~ $t+25$ ~~~~~~~~~~~~~ $t+30$ ~ ~~~~~~~~~~~ $t+35$ ~ ~~~~~~~~~~~~ $t+40$ ~ ~~~~~~~~~~ $t+45$  \\
\begin{tabular}{l} Input \\ ~\\ ~\\ ~\\ Ours \\ \end{tabular}   &
\raisebox{-.5\totalheight}{\includegraphics[width=0.8\textwidth]{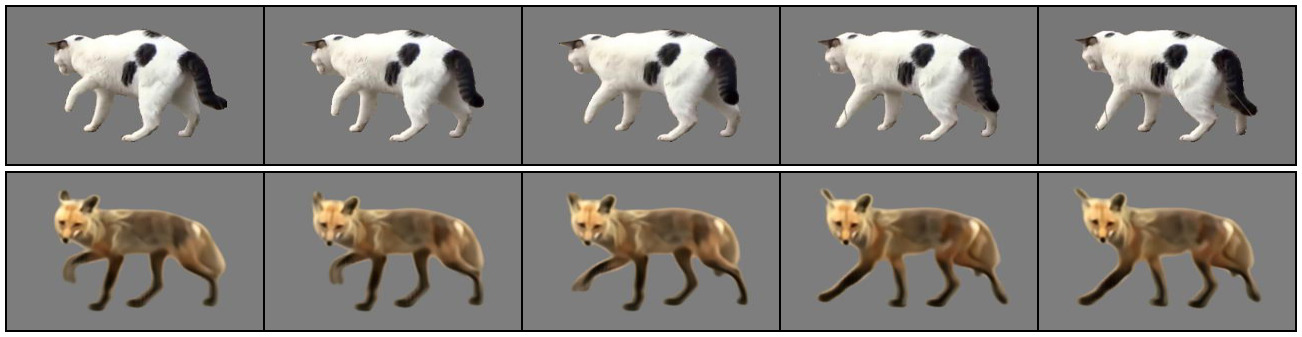}} \\
\noalign{\vskip 2mm} 
\begin{tabular}{l} Input \\ ~\\ ~\\ ~\\ Ours \\ \end{tabular}   &
\raisebox{-.5\totalheight}{\includegraphics[width=0.8\textwidth]{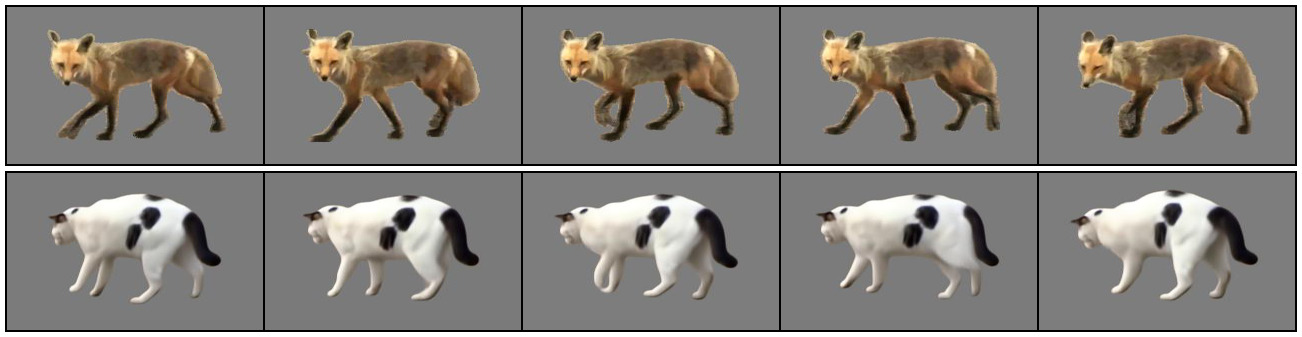}} \\
\end{tabular}

\caption{Additional results for our method.}
\label{fig:add4}
\end{figure*}

\begin{figure*}[h]

\centering
\begin{tabular}{ll}

~~~~~~ & ~~~~~~~~~~~ $t$ ~~~~~~~~~~~~~~~~~~ $t+2$ ~ ~~~~~~~~~~~~~~ $t+4$ ~ ~~~~~~~~~~~~ $t+6$ ~ ~~~~~~~~~~~~~ $t+8$  \\

\begin{tabular}{l} Input \\ ~\\ ~\\ ~\\ Ours \\ \end{tabular}   &
\raisebox{-.5\totalheight}{\includegraphics[width=0.8\textwidth]{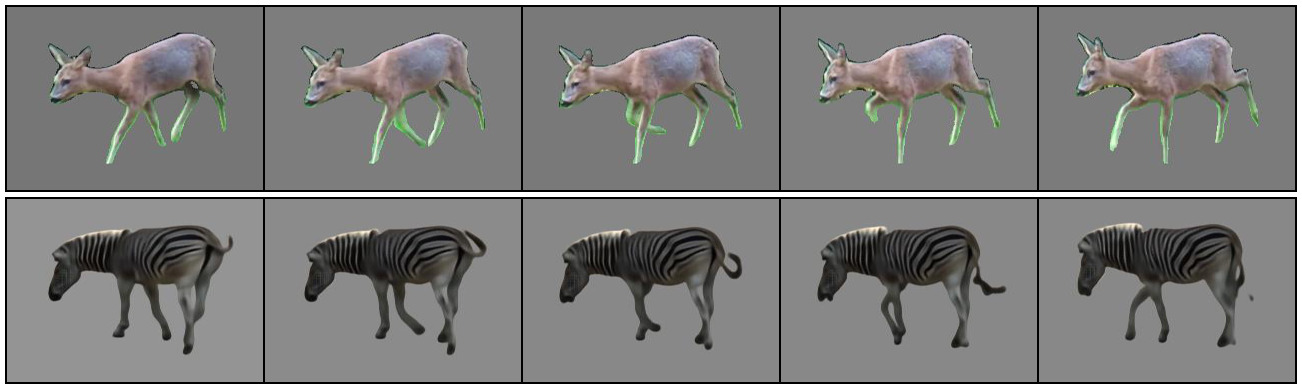}} \\
\noalign{\vskip 2mm} 
\begin{tabular}{l} Input \\ ~\\ ~\\ ~\\ Ours \\ \end{tabular}   &
\raisebox{-.5\totalheight}{\includegraphics[width=0.8\textwidth]{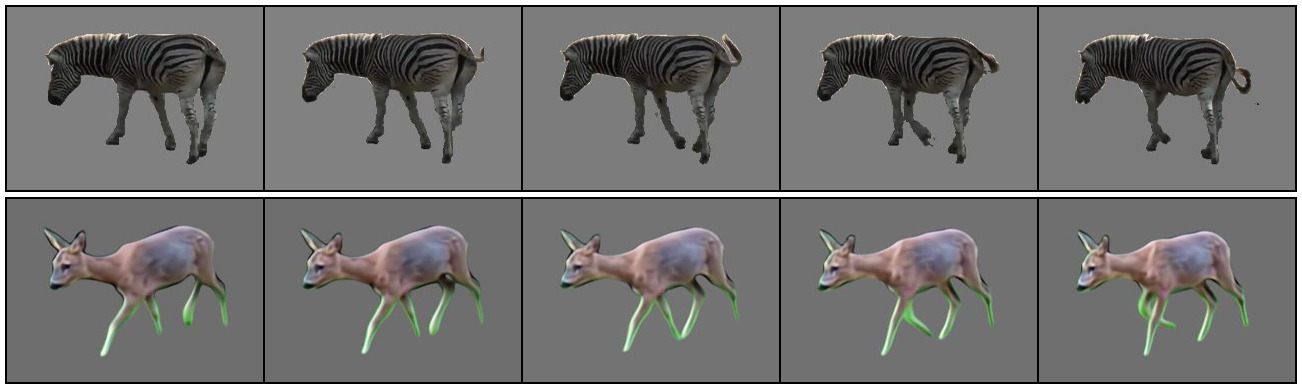}} \\
\noalign{\vskip 2mm} 
~ & ~~~~~~~ $t+10$ ~~~~~~~~~~~~~ $t+12$ ~ ~~~~~~~~~~~ $t+14$ ~ ~~~~~~~~~~~~ $t+16$ ~ ~~~~~~~~~~ $t+18$  \\
\begin{tabular}{l} Input \\ ~\\ ~\\ ~\\ Ours \\ \end{tabular}   &
\raisebox{-.5\totalheight}{\includegraphics[width=0.8\textwidth]{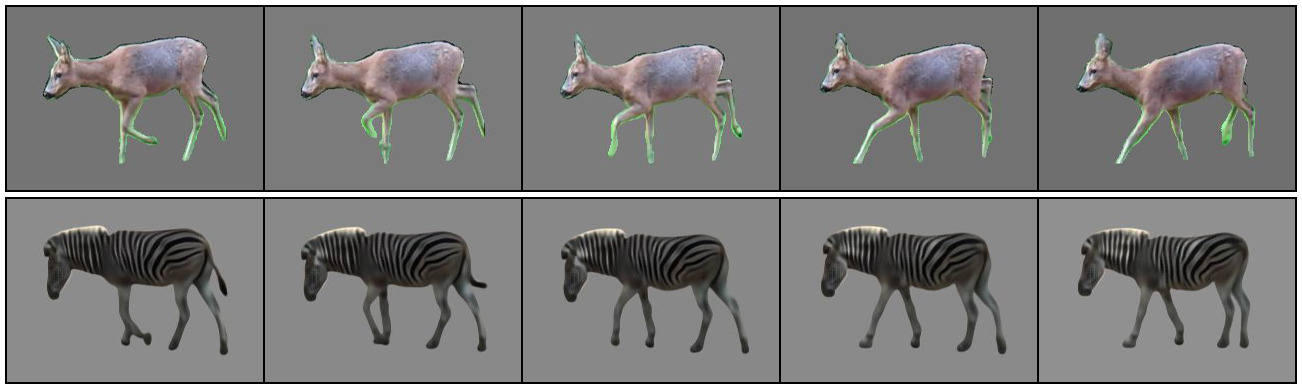}} \\
\noalign{\vskip 2mm} 
\begin{tabular}{l} Input \\ ~\\ ~\\ ~\\ Ours \\ \end{tabular}   &
\raisebox{-.5\totalheight}{\includegraphics[width=0.8\textwidth]{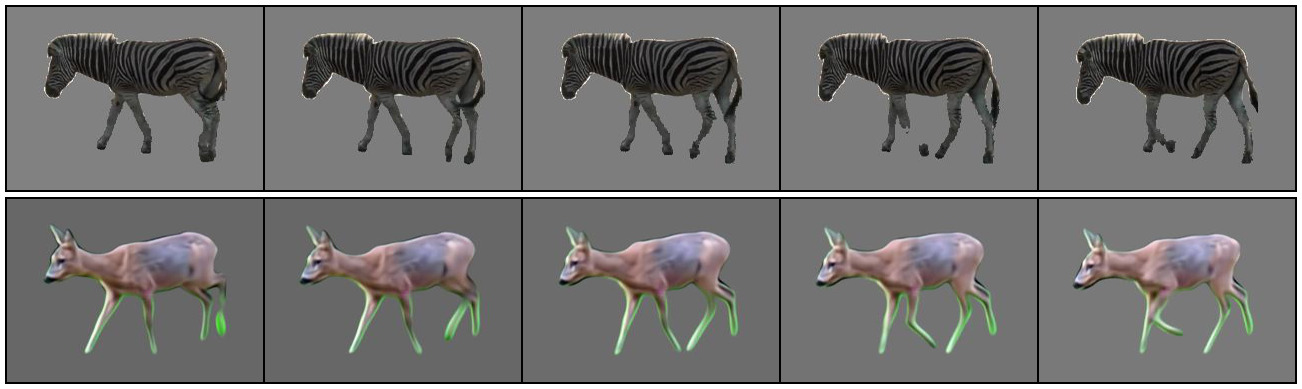}} \\
\end{tabular}

\caption{Additional results for our method.}
\label{fig:add5}
\end{figure*}

\begin{figure*}[h]

\centering
\begin{tabular}{ll}

~~~~~~ & ~~~~~~~~~~~ $t$ ~~~~~~~~~~~~~~~~~~ $t+2$ ~ ~~~~~~~~~~~~~~ $t+4$ ~ ~~~~~~~~~~~~ $t+6$ ~ ~~~~~~~~~~~~~ $t+8$  \\

\begin{tabular}{l} Input \\ ~\\ ~\\ ~\\ Ours \\ \end{tabular}   &
\raisebox{-.5\totalheight}{\includegraphics[width=0.8\textwidth]{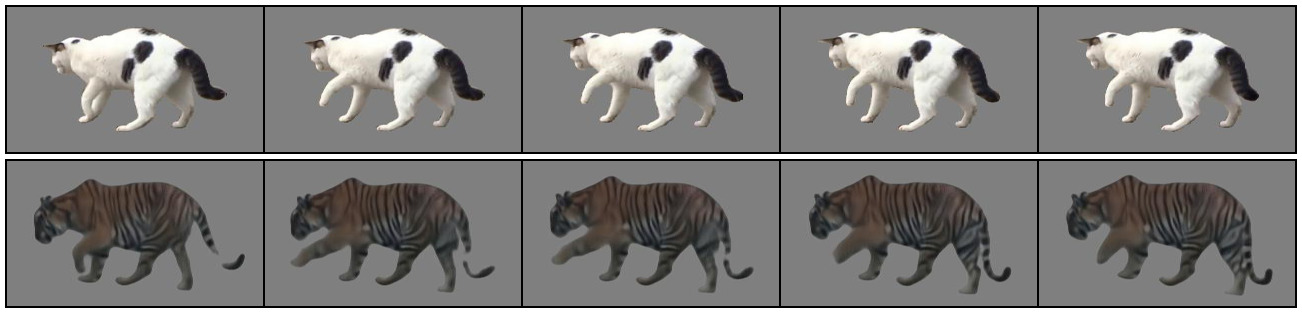}} \\
\noalign{\vskip 2mm} 
\begin{tabular}{l} Input \\ ~\\ ~\\ ~\\ Ours \\ \end{tabular}   &
\raisebox{-.5\totalheight}{\includegraphics[width=0.8\textwidth]{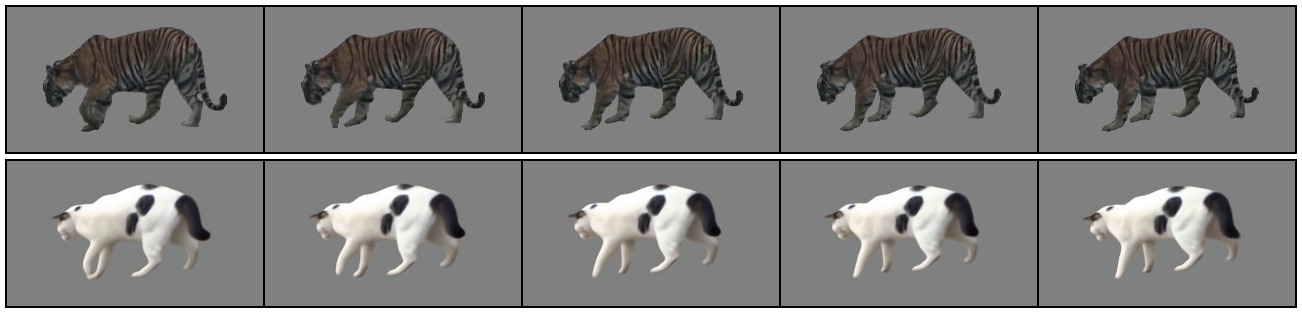}} \\
\noalign{\vskip 2mm} 
~ & ~~~~~~~ $t+10$ ~~~~~~~~~~~~~ $t+12$ ~ ~~~~~~~~~~~ $t+14$ ~ ~~~~~~~~~~~~ $t+16$ ~ ~~~~~~~~~~ $t+18$  \\
\begin{tabular}{l} Input \\ ~\\ ~\\ ~\\ Ours \\ \end{tabular}   &
\raisebox{-.5\totalheight}{\includegraphics[width=0.8\textwidth]{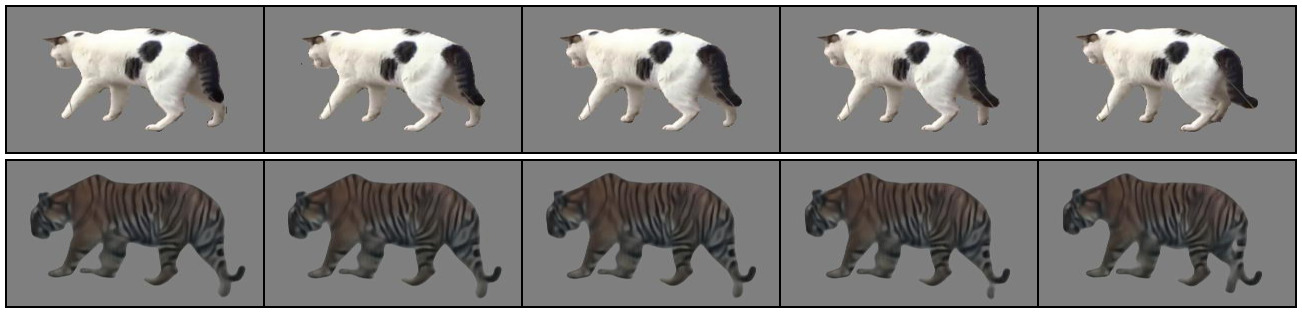}} \\
\noalign{\vskip 2mm} 
\begin{tabular}{l} Input \\ ~\\ ~\\ ~\\ Ours \\ \end{tabular}   &
\raisebox{-.5\totalheight}{\includegraphics[width=0.8\textwidth]{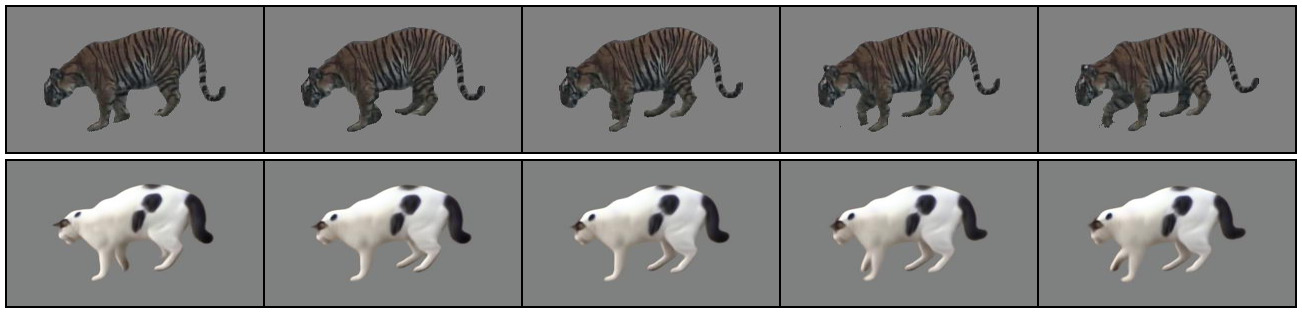}} \\
\end{tabular}

\caption{Additional results for our method.}
\label{fig:add6}
\end{figure*}

\begin{figure*}[h]

\centering
\begin{tabular}{ll}
~ & ~~~~~~~~~~~ $t$ ~~~~~~~~~~~~~~~~~ $t+20$ ~ ~~~~~~~~~~~ $t+40$ ~ ~~~~~~~~~~~~ $t+60$ ~ ~~~~~~~~~~ $t+80$  \\


\begin{tabular}{l} Input \\ ~\\ ~\\ ~\\ Ours \\ \end{tabular}   &
\raisebox{-.5\totalheight}{\includegraphics[width=0.8\textwidth]{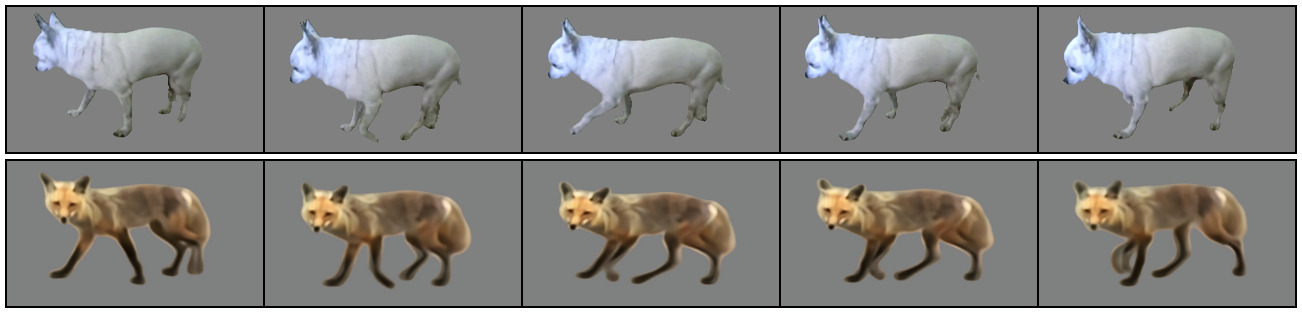}} \\
\noalign{\vskip 2mm} 
~ & ~~~~~~~~~~~ $t$ ~~~~~~~~~~~~~~~~~~~ $t+5$ ~ ~~~~~~~~~~~ $t+10$ ~ ~~~~~~~~~~~~ $t+15$ ~ ~~~~~~~~~~ $t+20$  \\

\begin{tabular}{l} Input \\ ~\\ ~\\ ~\\ Ours \\ \end{tabular}   &
\raisebox{-.5\totalheight}{\includegraphics[width=0.8\textwidth]{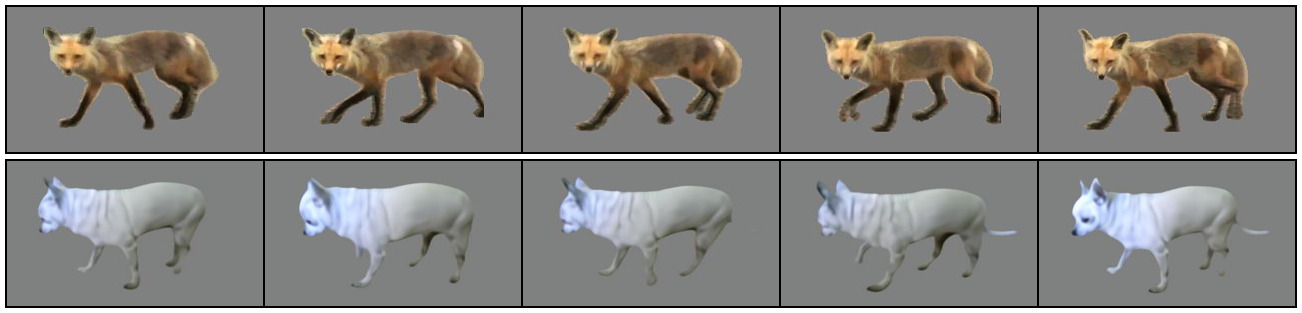}} \\
\noalign{\vskip 2mm} 
~ & ~~~~~~~ $t+100$ ~~~~~~~~~~ $t+120$ ~~~~~~~~~~~ $t+140$ ~~~~~~~~~~~ $t+160$ ~~~~~~~~~~~ $t+180$  \\
\begin{tabular}{l} Input \\ ~\\ ~\\ ~\\ Ours \\ \end{tabular}   &
\raisebox{-.5\totalheight}{\includegraphics[width=0.8\textwidth]{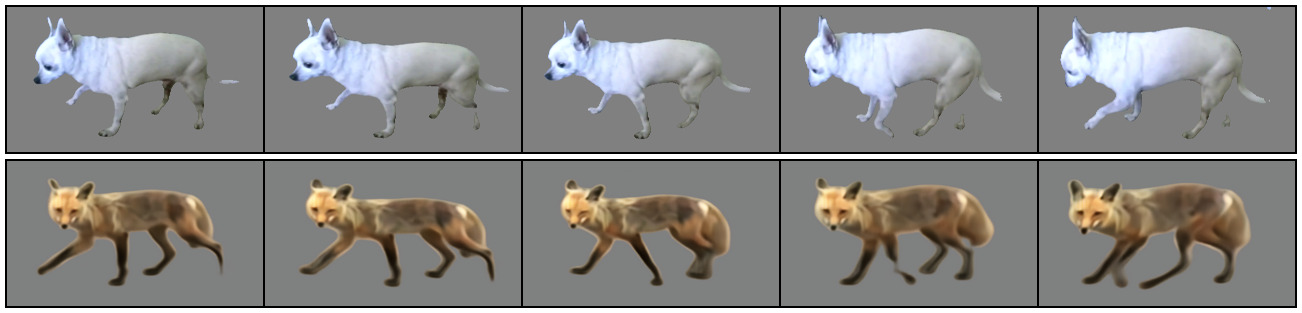}} \\
\noalign{\vskip 2mm} 

~ & ~~~~~~~ $t+25$ ~~~~~~~~~~~~~ $t+30$ ~ ~~~~~~~~~~~ $t+35$ ~ ~~~~~~~~~~~~ $t+40$ ~ ~~~~~~~~~~ $t+45$  \\

\begin{tabular}{l} Input \\ ~\\ ~\\ ~\\ Ours \\ \end{tabular}   &
\raisebox{-.5\totalheight}{\includegraphics[width=0.8\textwidth]{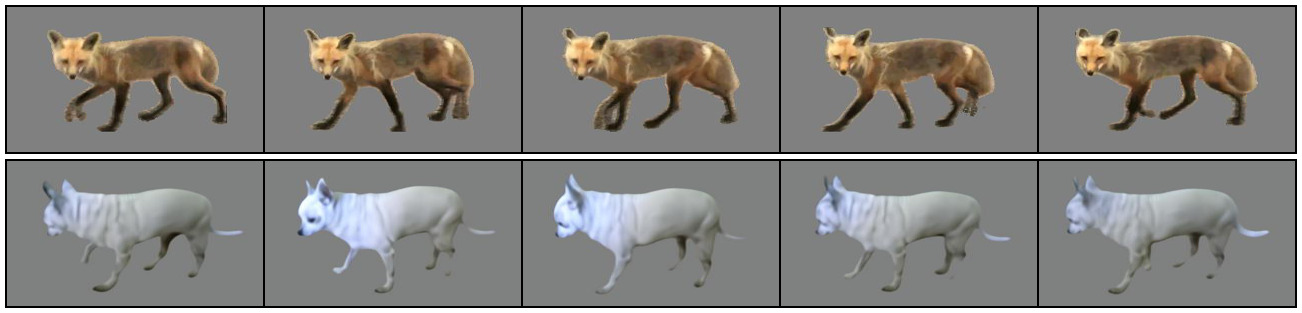}} \\
\end{tabular}

\caption{Additional results for our method. We show different time offsets, as the input videos has significantly different frame rates.}
\label{fig:add7}
\end{figure*}

\begin{figure*}[h]
\begin{tabular}{l}

~~~~~~~~ $\leftarrow$ ~~~~~~~~~~~~~~~~~~~~~~~~~~~~~~~~~~~~~~~~~~~~~~~~~~~~~~~~~~ Original  ~~~~~~~~~~~~~~~~~~~~~~~~~~~~~~~~~~~~~~~~~~~~~~~~~~~~~~~~~ $\rightarrow$ \\
\raisebox{-.5\totalheight}{\includegraphics[width=1\textwidth]{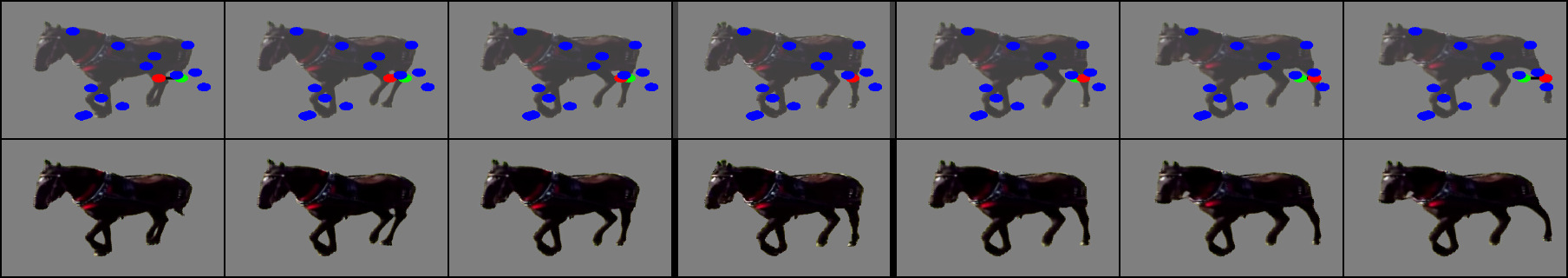}} 
\vspace{0.5cm}
\\

~~~~~~~~ $\leftarrow$ ~~~~~~~~~~~~~~~~~~~~~~~~~~~~~~~~~~~~~~~~~~~~~~~~~~~~~~~~~~ Original  ~~~~~~~~~~~~~~~~~~~~~~~~~~~~~~~~~~~~~~~~~~~~~~~~~~~~~~~~~ $\rightarrow$ \\
\raisebox{-.5\totalheight}{\includegraphics[width=1\textwidth]{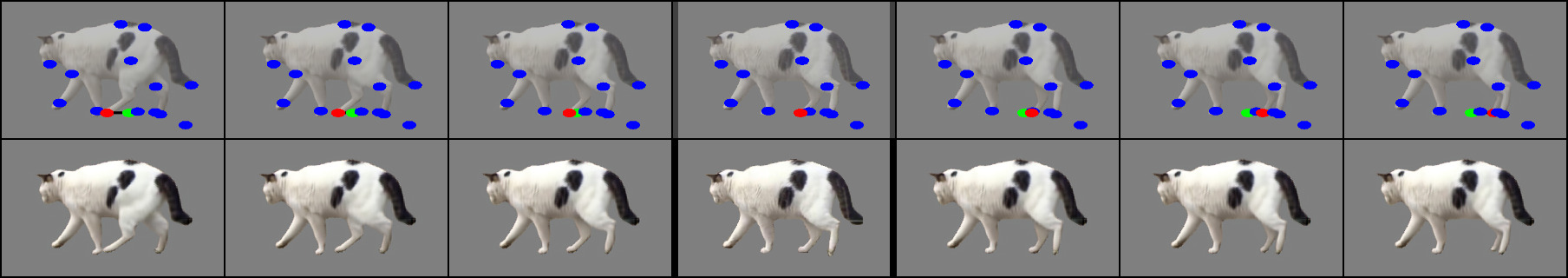}} 
\vspace{0.5cm}
\\

~~~~~~~~ $\downarrow$ ~~~~~~~~~~~~~~~~~~~~~~~~~~~~~~~~~~~~~~~~~~~~~~~~~~~~~~~~~~ Original  ~~~~~~~~~~~~~~~~~~~~~~~~~~~~~~~~~~~~~~~~~~~~~~~~~~~~~~~~~~~~ $\uparrow$ \\
\raisebox{-.5\totalheight}{\includegraphics[width=1\textwidth]{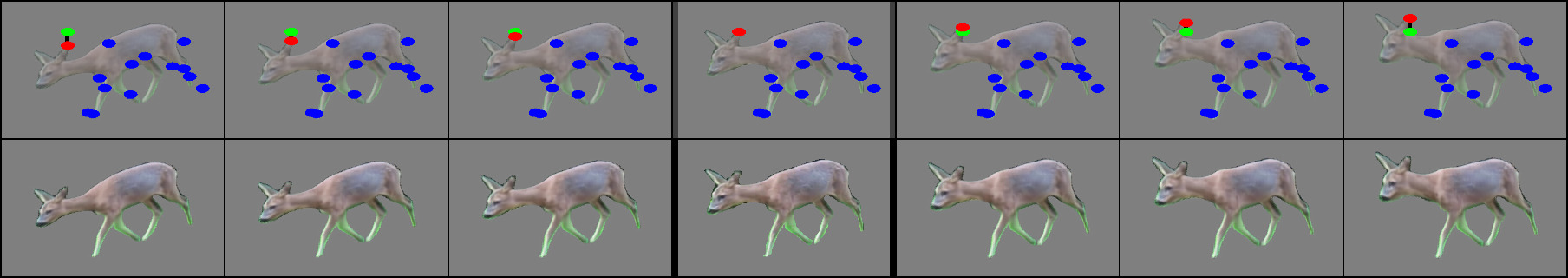}} 
\vspace{0.5cm}
\\

~~~~~~~~ $\leftarrow$ ~~~~~~~~~~~~~~~~~~~~~~~~~~~~~~~~~~~~~~~~~~~~~~~~~~~~~~~~~~ Original  ~~~~~~~~~~~~~~~~~~~~~~~~~~~~~~~~~~~~~~~~~~~~~~~~~~~~~~~~~ $\rightarrow$ \\
\raisebox{-.5\totalheight}{\includegraphics[width=1\textwidth]{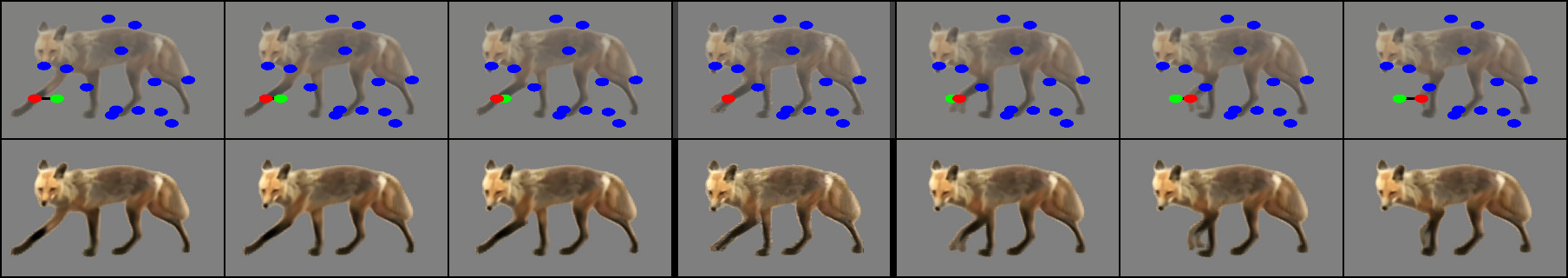}} 

\end{tabular}
\caption{Additional editing results. Top to bottom: Moving the horse leg left/right, Moving the cat leg left/right, Moving the deer head up/down, Moving the fox leg left/right. As can be seen, moving the red keypoint induces a meaningful and realistic editing operation. }
\label{fig:editing_supp}
\end{figure*}

\begin{figure*}[ht]
\centering
\begin{tabular}{ll}
& ~~~~~~~~~~~ $t$ ~~~~~~~~~~~~~~~~~~~ $t+5$  ~~~~~~~~~~~~~~ $t+10$  ~~~~~~~~~~~~~ $t+15$ ~~~~~~~~~~~~~ $t+20$ \\

\rotatebox[origin=t]{90}{~~~~~~~~~ EDN ~~~~~~~~~~~~~~~~~ Ours ~~~~~~~~~~~~~~~~~ Input}   &
\raisebox{-.425\totalheight}{\includegraphics[width=0.8\textwidth]{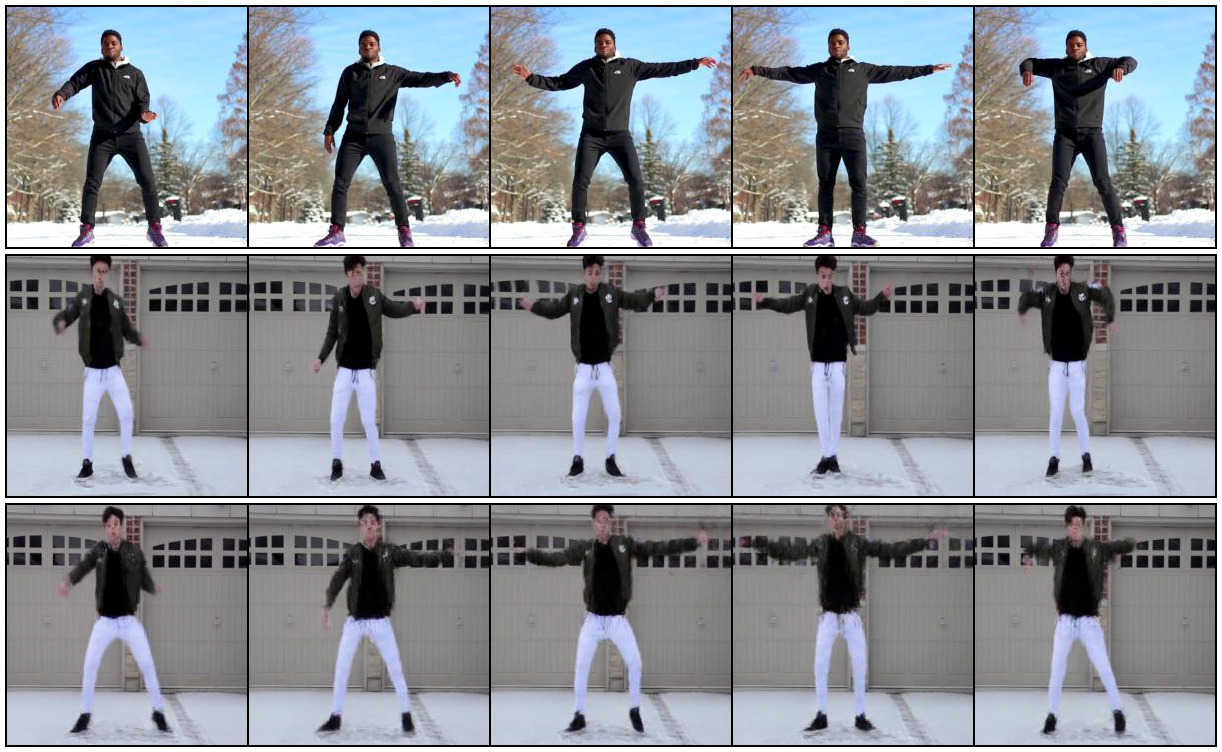}} \\ 
\noalign{\vskip 2.0mm} 
\rotatebox[origin=t]{90}{~~~~~~~~~ EDN ~~~~~~~~~~~~~~~~~ Ours ~~~~~~~~~~~~~~~~~ Input}   &
\raisebox{-.425\totalheight}{\includegraphics[width=0.8\textwidth]{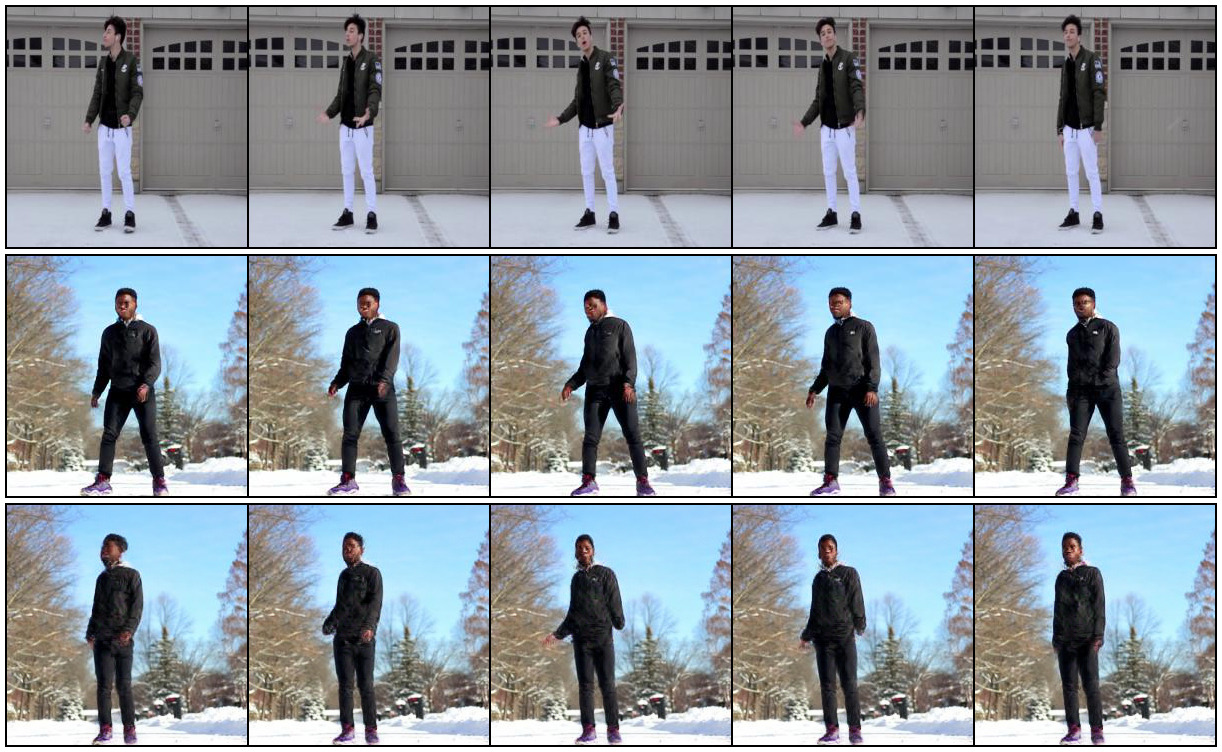}} \\ 

\end{tabular}
\caption{Dancing persons videos with comparison to EDN~\cite{chan2019everybody}. As can be seen, our results are comparable to EDN~\cite{chan2019everybody} which assumes a stronger supervision.
}
\vspace{-0.5cm}
\label{fig:edn}
\end{figure*}

\begin{figure*}[ht]
\centering
\begin{tabular}{ll}
& ~~~~~~~~~~~ $t$ ~~~~~~~~~~~~~~~~~~~ $t+5$  ~~~~~~~~~~~~~~ $t+10$  ~~~~~~~~~~~~~ $t+15$ ~~~~~~~~~~~~~ $t+20$ \\

\rotatebox[origin=t]{90}{~~~~~~~~~ EDN ~~~~~~~~~~~~~~~~~ Ours ~~~~~~~~~~~~~~~~~ Input}   &
\raisebox{-.425\totalheight}{\includegraphics[width=0.8\textwidth]{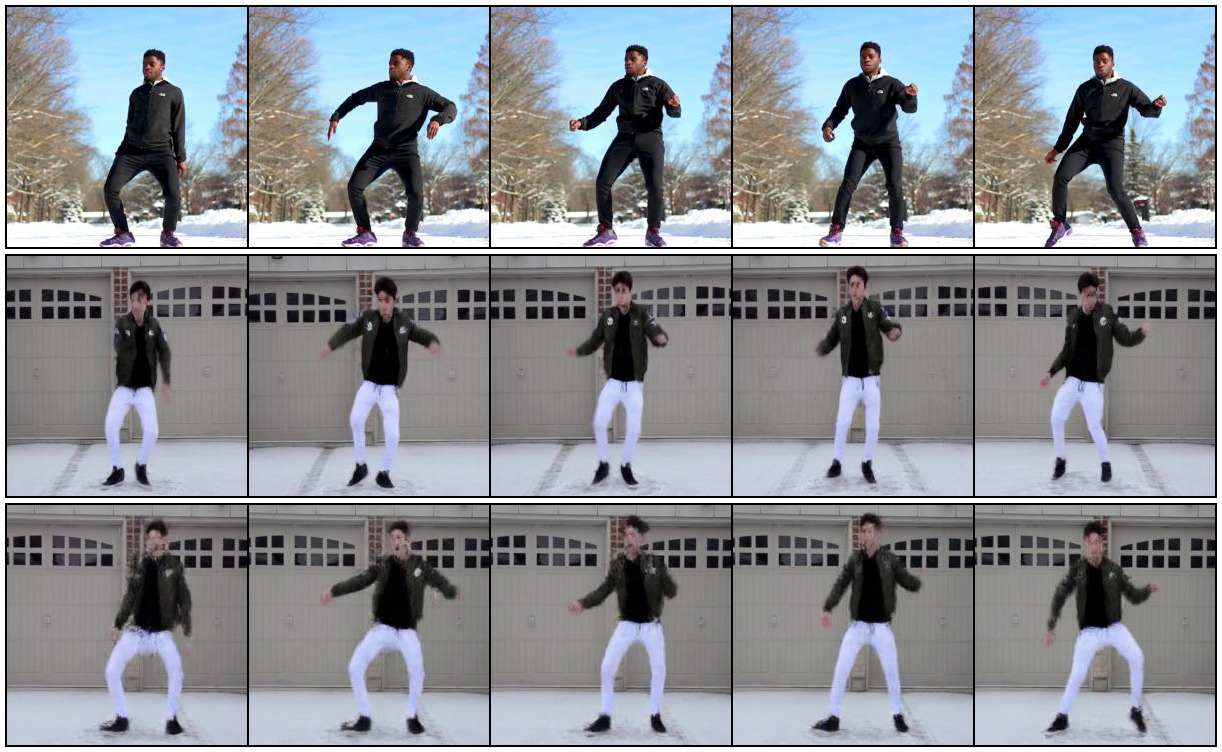}} \\ 
\noalign{\vskip 2.0mm} 
\rotatebox[origin=t]{90}{~~~~~~~~~ EDN ~~~~~~~~~~~~~~~~~ Ours ~~~~~~~~~~~~~~~~~ Input}   &
\raisebox{-.425\totalheight}{\includegraphics[width=0.8\textwidth]{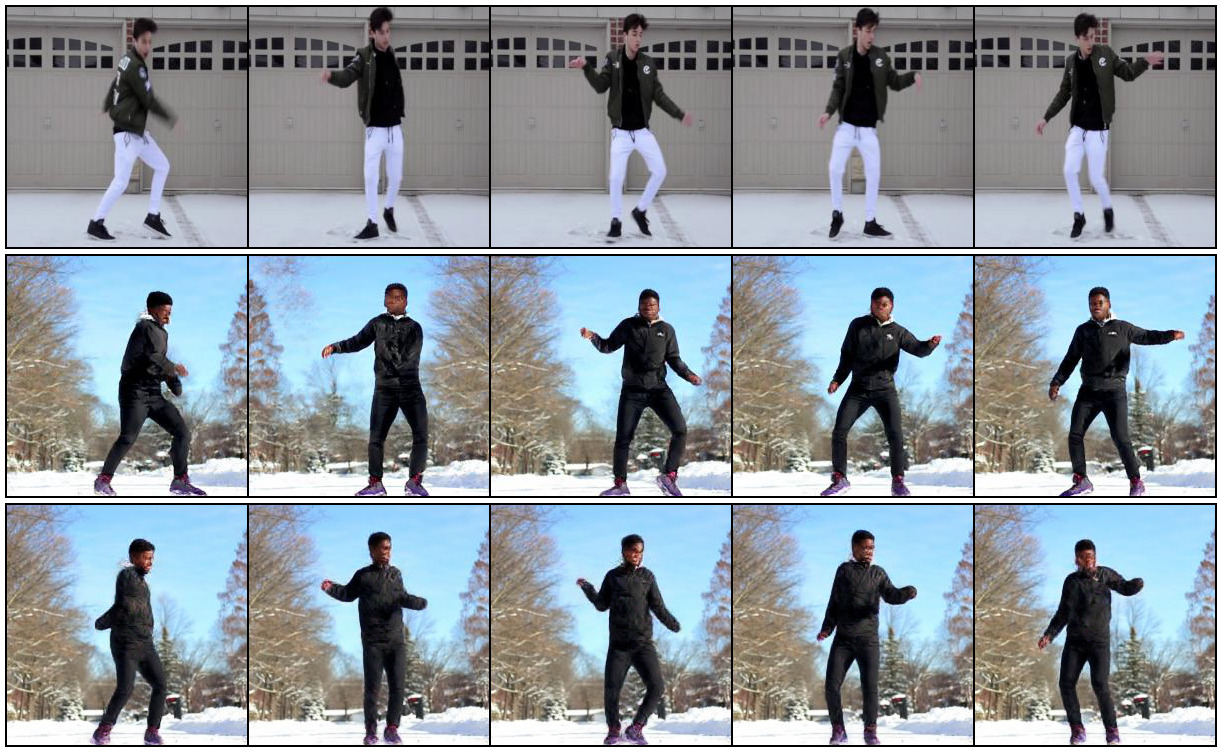}} \\ 

\end{tabular}
\caption{Dancing person videos with comparison to EDN~\cite{chan2019everybody}. As can be seen, our results are comparable to EDN~\cite{chan2019everybody} which assumes stronger supervision.
}
\vspace{-0.5cm}
\label{fig:edn2}
\end{figure*}

\begin{figure*}[ht]
\centering
\begin{tabular}{ll}

\rotatebox[origin=t]{90}{Input}   &
\raisebox{-.3\totalheight}{\includegraphics[width=0.9\textwidth]{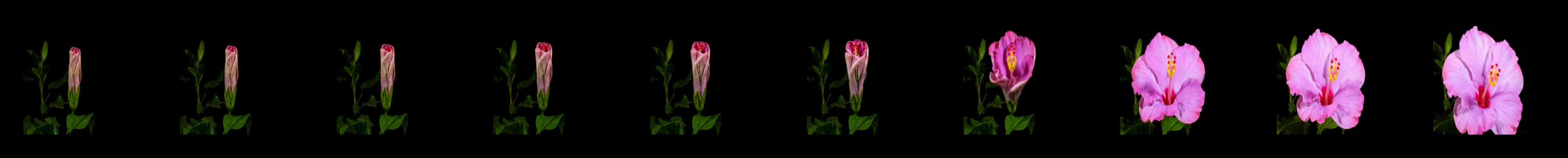}} \\
\rotatebox[origin=t]{90}{Ours}   &
\raisebox{-.3\totalheight}{\includegraphics[width=0.9\textwidth]{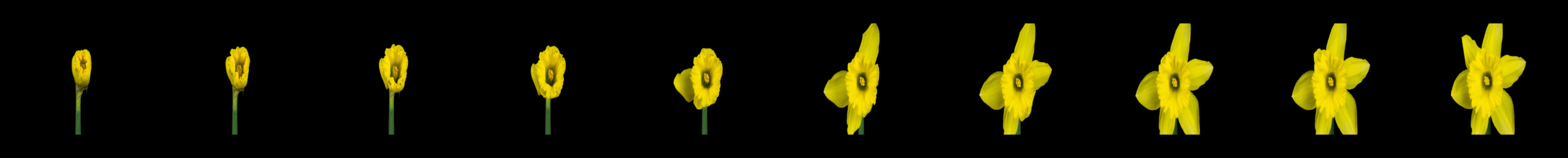}}  \\
\noalign{\vskip 1.5mm} 
\rotatebox[origin=t]{90}{Input}   &
\raisebox{-.3\totalheight}{\includegraphics[width=0.9\textwidth]{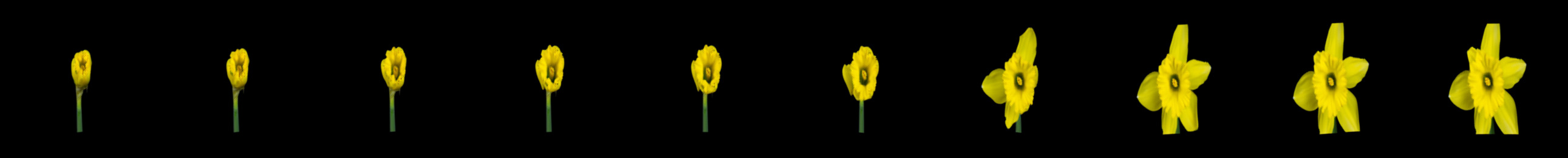}} \\
\rotatebox[origin=t]{90}{Ours}   &
\raisebox{-.3\totalheight}{\includegraphics[width=0.9\textwidth]{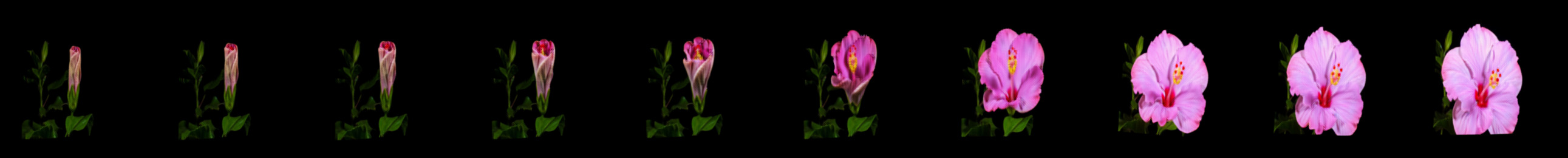}} \\

\end{tabular}
\caption{Blooming flowers results.
}
\label{fig:fl}
\end{figure*}

\begin{figure*}[ht]
\centering
\begin{tabular}{lll}
& ~~~~~~~~ Input ~~~~~~~~~~~ Shape ~~~~~~~~~~~~ Output   &
  ~~~~~~~~ Input ~~~~~~~~~~~ Shape ~~~~~~~~~~~~ Output  \\

\rotatebox[origin=t]{90}{Seed A}   &
\raisebox{-.3\totalheight}{\includegraphics[width=0.45\textwidth]{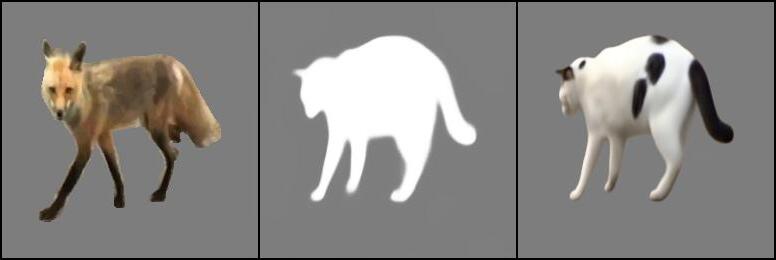}} &
\raisebox{-.3\totalheight}{\includegraphics[width=0.45\textwidth]{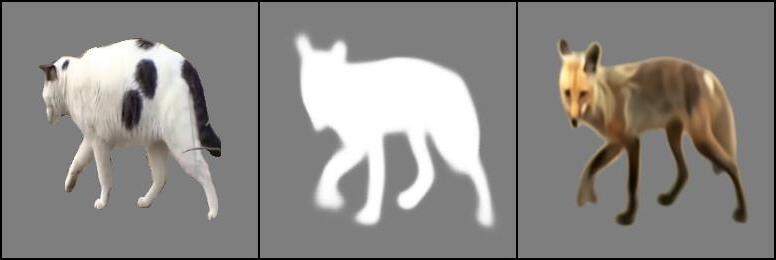}} \\

\rotatebox[origin=t]{90}{Seed B}   &
\raisebox{-.3\totalheight}{\includegraphics[width=0.45\textwidth]{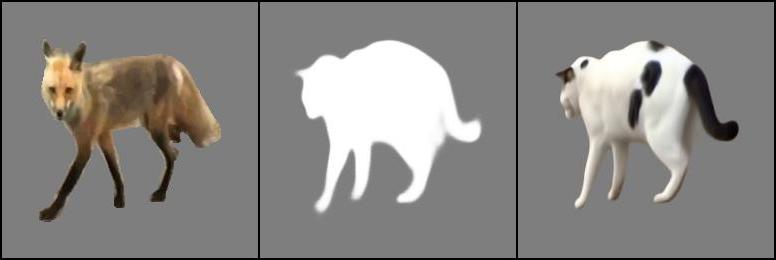}} &
\raisebox{-.3\totalheight}{\includegraphics[width=0.45\textwidth]{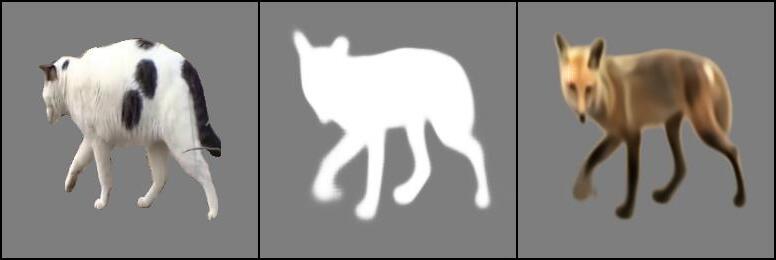}} \\

\rotatebox[origin=t]{90}{Seed C}   &
\raisebox{-.3\totalheight}{\includegraphics[width=0.45\textwidth]{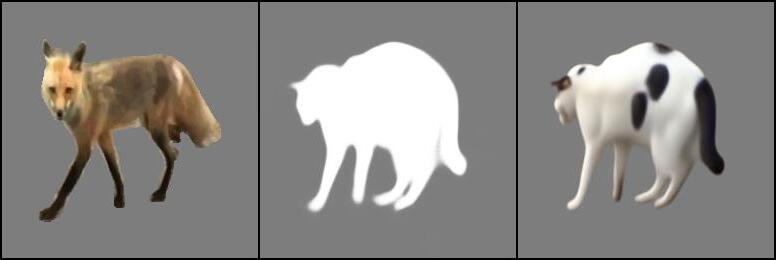}} &
\raisebox{-.3\totalheight}{\includegraphics[width=0.45\textwidth]{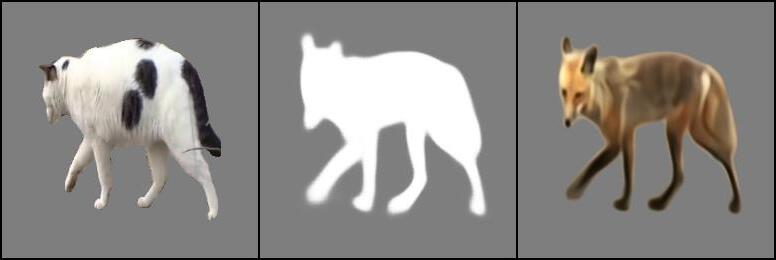}} \\

\end{tabular}
\caption{Results for different random seeds, as can be seen, our method generates similar results for the random initializations. Left to right: Input image, intermediate shape, and final result.
}
\label{fig:seed}
\end{figure*}

 Additional GIF synchronization result is in Fig.~\ref{fig:elk}. Comparisons to baselines over four-legged animals is given in Fig.~\ref{fig:comp1} to Fig.~\ref{fig:comp6}. Additional results for our method is given in Fig.~\ref{fig:add1} to Fig.~\ref{fig:add7}. Editing results are in Fig.~\ref{fig:editing_supp}. Both Fig~.\ref{fig:edn} and Fig~.\ref{fig:edn2} present additional dancing results, while Fig~.\ref{fig:fl} presents additional flower results. Fig.~\ref{fig:seed} demonstrates that our method generates similar results for different random initializations.


\clearpage
\end{document}